\documentclass{article}

\pdfoutput=1

\usepackage[numbers]{natbib}
\usepackage{fullpage}

\usepackage[utf8]{inputenc} %
\usepackage[T1]{fontenc}    %
\usepackage{hyperref}       %
\usepackage{url}            %
\usepackage{booktabs}       %
\usepackage{amsfonts}       %
\usepackage{nicefrac}       %
\usepackage{microtype}      %

\usepackage{graphicx} %
\usepackage{subfigure} 
\usepackage{amssymb}
\usepackage{amsmath}
\usepackage{epstopdf}
\usepackage{multirow}
\usepackage{pdflscape}
\usepackage{commath}

\newcommand{\mc}[1]{\mathcal{#1}}

\DeclareMathOperator*{\E}{E}
\newcommand{\expec}[2]{\E_{#1}\left[ #2 \right]}

\newcommand{\pdfnorm}[3]{\mc{N}(#1\vert #2, #3)}

\newcommand{\algfull}{Monte Carlo Structured SVI}
\newcommand{\alg}{MC-SSVI}
\newcommand{\lvm}{2L-LVM}
\newcommand{\lvmone}{1L-LVM}
\newcommand{\dsvi}{S-DSVI}
\newcommand{\hsvi}{H-MC-SSVI}
\newcommand{\hsvifull}{Hybrid MC-SSVI}
\newcommand{\minibatch}{\mathcal{M}}
\newcommand{\cov}{S}

\newcommand{\citeAY}[1]{\cite{#1}}

\title{Monte Carlo Structured SVI for\\Two-Level Non-Conjugate Models}

\author{
Rishit Sheth and Roni Khardon \\
{\tt rishit.sheth@tufts.edu, roni@cs.tufts.edu} \\
Department of Computer Science \\
Tufts University \\
Medford, MA, USA
}

\begin{document}

\maketitle

\begin{abstract}
The stochastic variational inference (SVI) paradigm, which combines variational inference, natural gradients, and stochastic updates, was recently proposed for large-scale data analysis in conjugate Bayesian models and demonstrated to be effective in several problems. 
This paper studies a family of Bayesian latent variable models with two levels of hidden variables but without any conjugacy requirements, making several contributions in this context. 
The first is observing that SVI, with an improved structured variational approximation, 
is applicable under more general conditions than previously thought with the only requirement being that the approximating variational distribution be in the same family as the prior. 
The resulting approach, \algfull{} (\alg), significantly extends the scope of SVI, enabling large-scale learning in non-conjugate models. 
For models with latent Gaussian variables we propose a hybrid algorithm, using both standard and natural gradients, which is shown to improve stability and convergence.
Applications in mixed effects models, sparse Gaussian processes, probabilistic matrix factorization and correlated topic models demonstrate the generality of the approach and the advantages of the proposed algorithms.
\end{abstract}

\graphicspath{{figs/}}
\DeclareGraphicsExtensions{.pdf}
\newcommand{\mpw}{0.48\linewidth}
\newcommand{\mpwsc}{0.249\linewidth}

\newcommand{\putSGPPlotRef}{
\begin{figure*}[t]
\begin{center}
\includegraphics[width=0.5\linewidth]{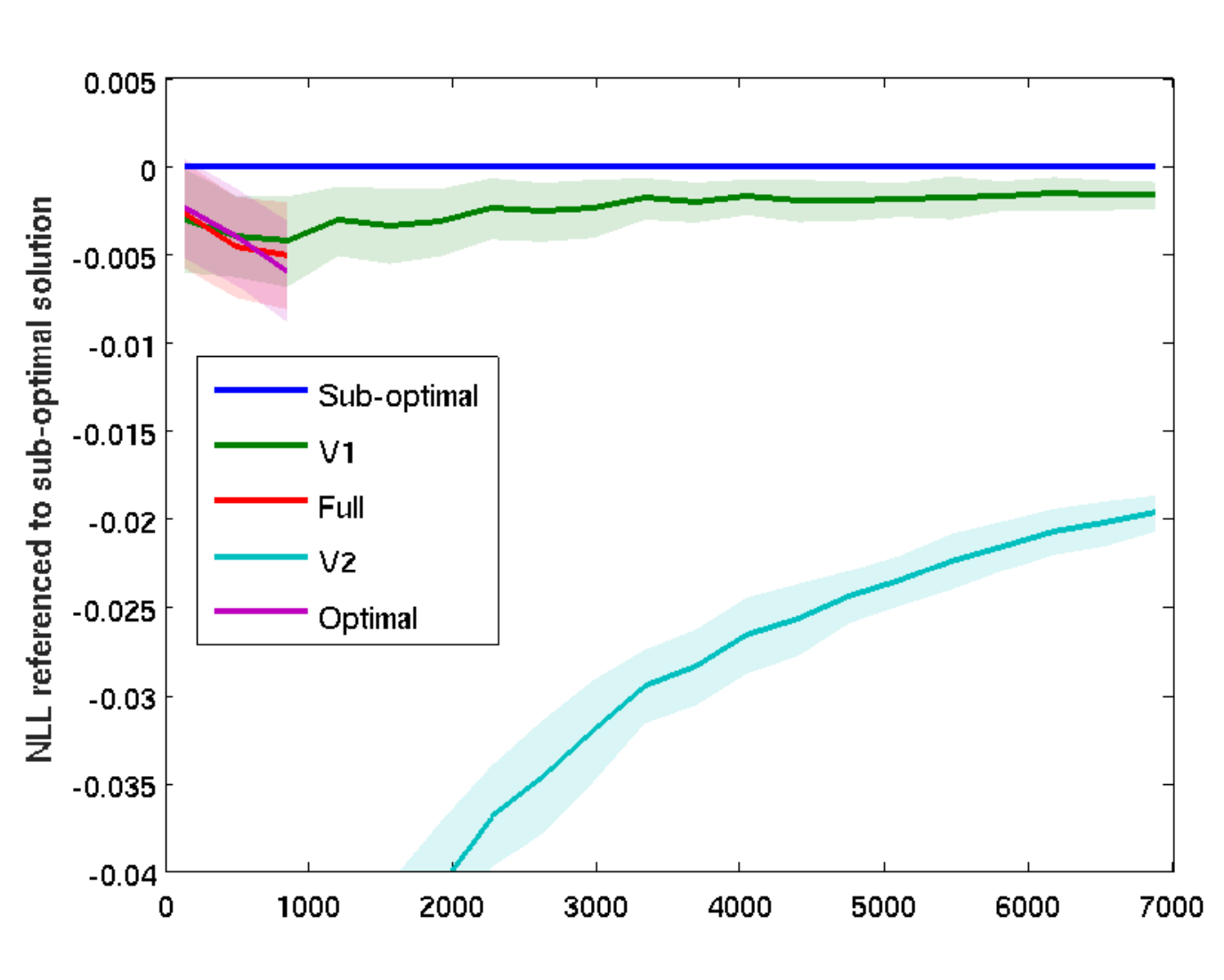}
\end{center}

\caption{
    Test NLL referenced to suboptimal solution vs. training set size.
}
\label{fig:sgpplotref}
\end{figure*}
}

\newcommand{\putGMEPlot}{
\begin{figure*}[t]
\begin{center}
    \includegraphics[width=0.5\linewidth]{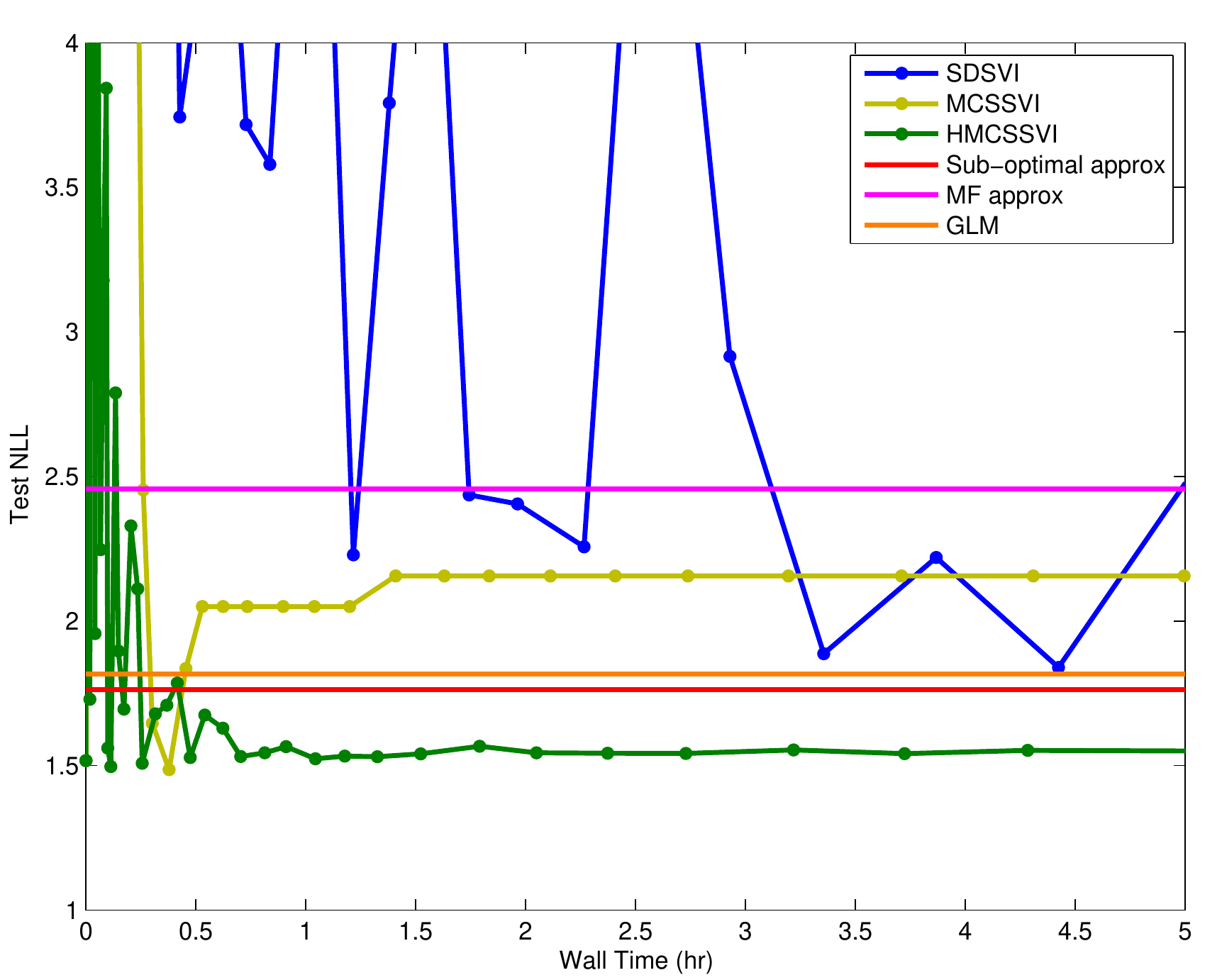}
\end{center}

\caption{
    Mixed effects GLM with Poisson likelihood experiment comparing performance of optimal, suboptimal, and mean-field approximations as well as different algorithms optimizing each criterion.
}
\label{fig:gme}
\end{figure*}
}

\newcommand{\putPMFPlotVLBMCSSVI}{
\begin{figure*}[t]
\begin{center}
\includegraphics[width=0.5\linewidth]{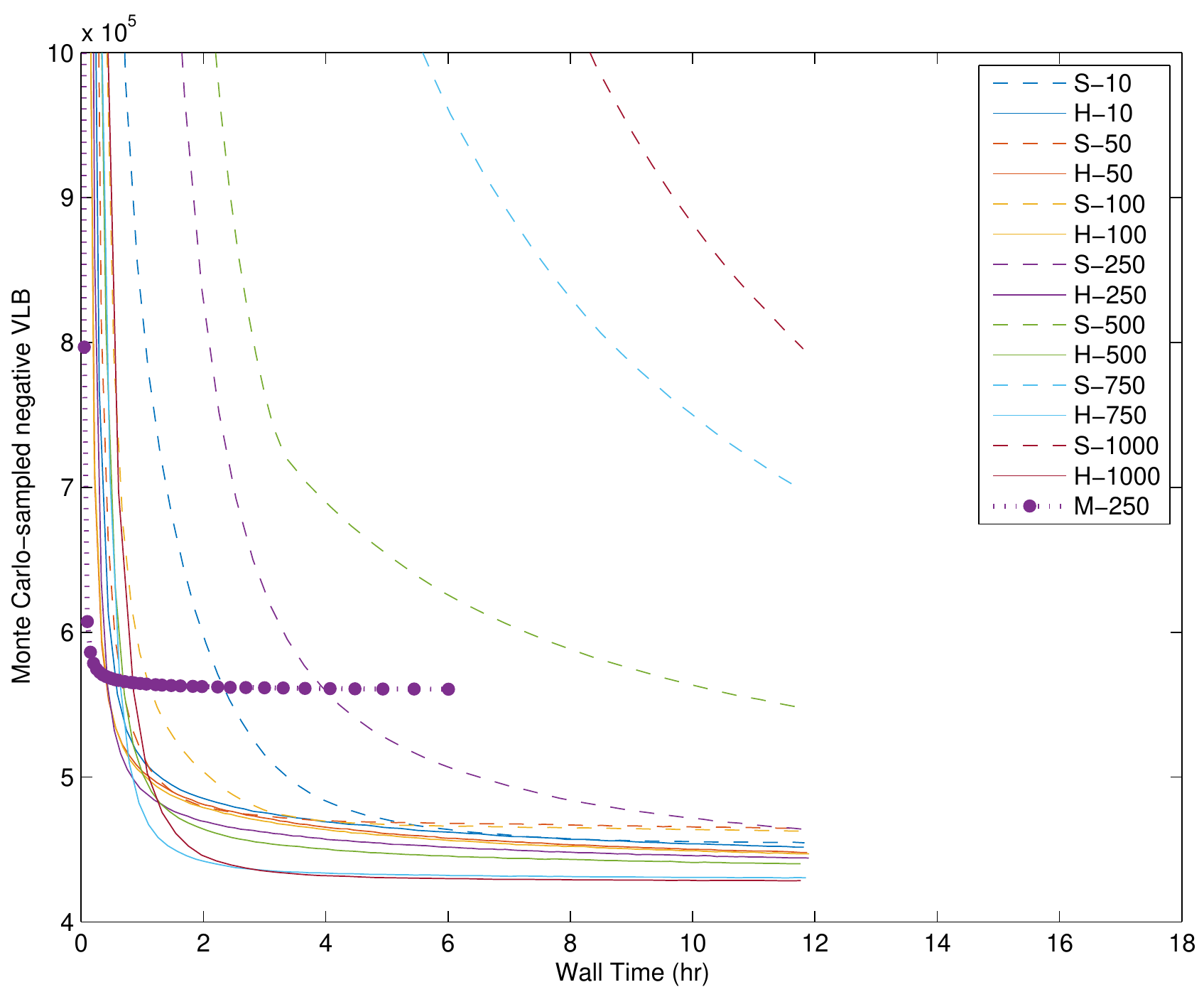}
\end{center}

\caption{
    Training criterion (negative Monte Carlo VLB) for logistic PMF experiment with output of MC-SSVI at 250 sub-samples overlaid on S-DSVI and H-MC-SSVI results.
}
\label{fig:pmfplotvlbmcssvi}
\end{figure*}
}

\newcommand{\putSGPPlotNonRef}{
\begin{figure*}[t]
\begin{center}
\includegraphics[width=0.5\linewidth]{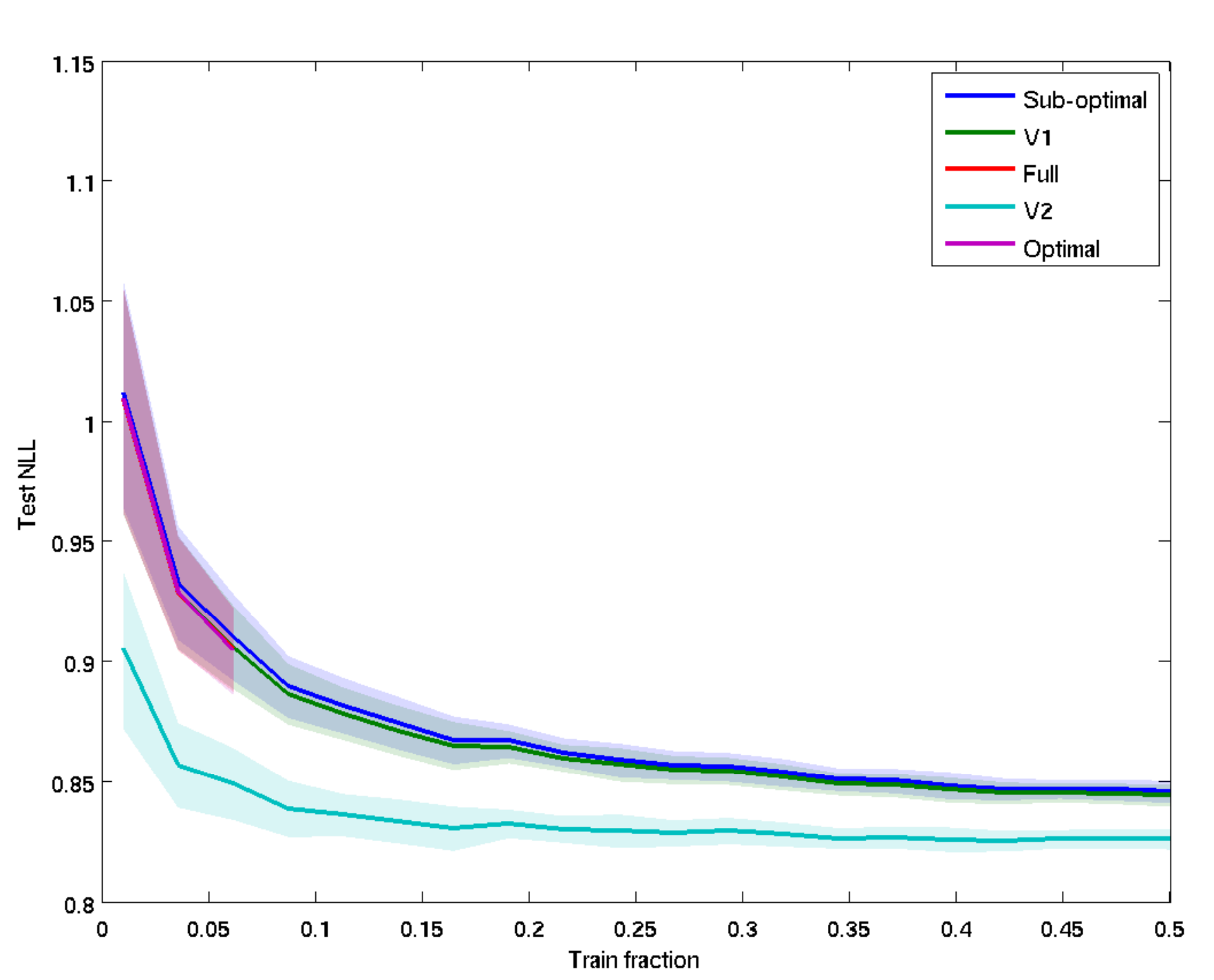}
\end{center}

\caption{
    ``Non-referenced'' version of sparse GP figure in main paper.
}
\label{fig:sgpplotnonref}
\end{figure*}
}

\newcommand{\putWangBleiEval}{
\begin{figure*}[h]

\begin{minipage}{0.475\linewidth}
\begin{center}
    \includegraphics[width=0.99\linewidth]{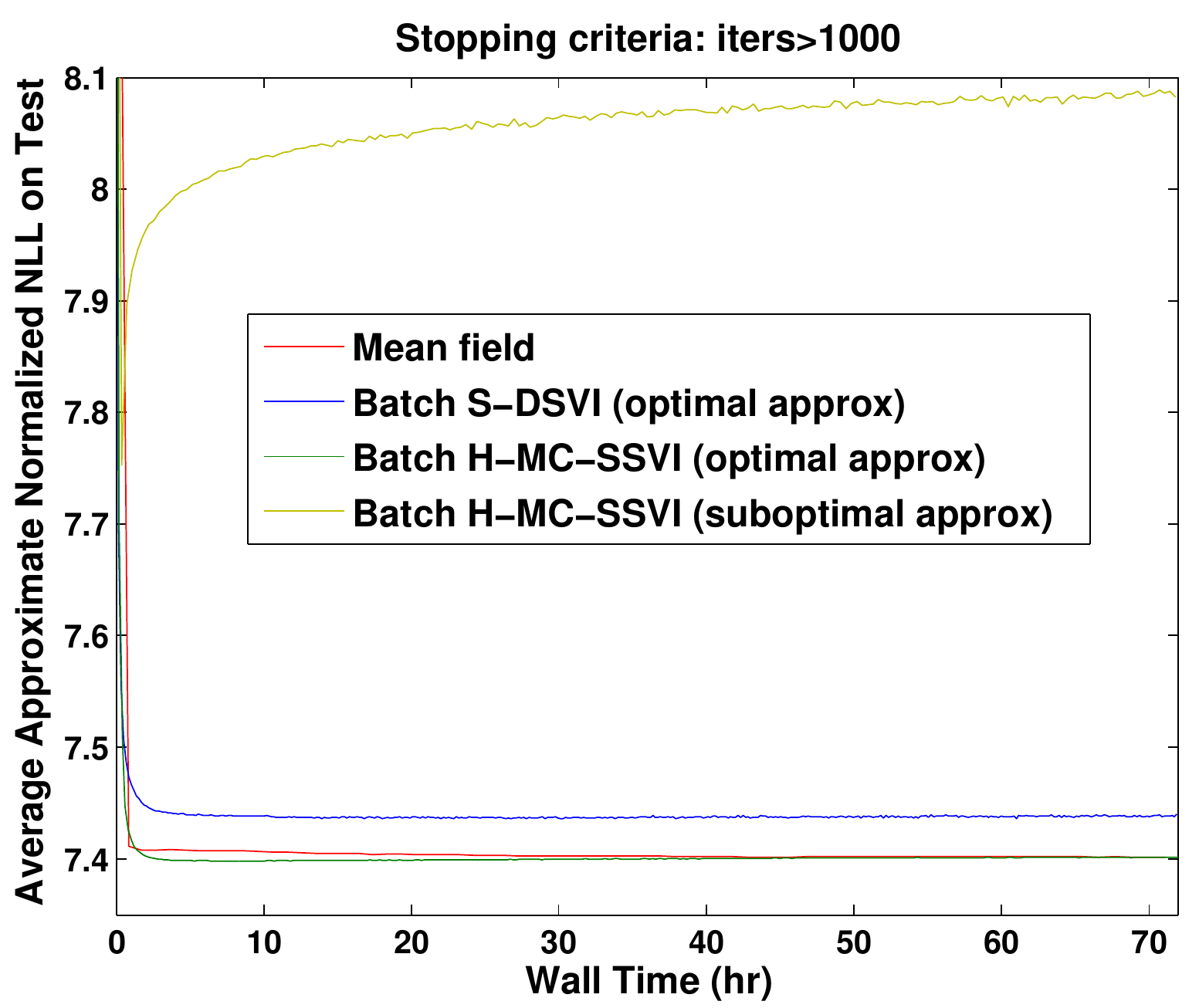}
\end{center}
\end{minipage}\quad
\begin{minipage}{0.475\linewidth}
\begin{center}
\includegraphics[width=0.99\linewidth]{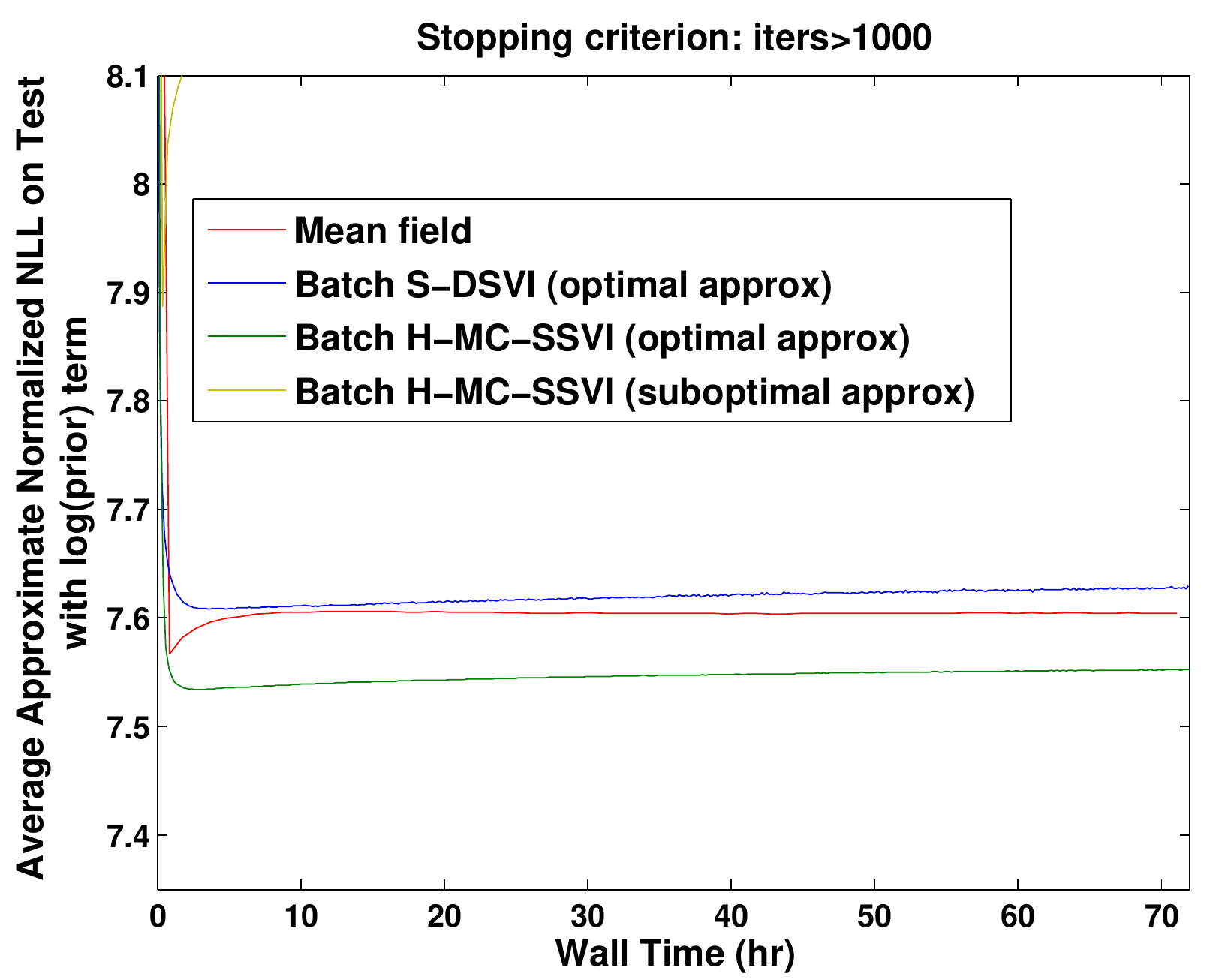}
\end{center}
\end{minipage}

\caption{
Approximations of CTM normalized NLL values on \emph{nips} test set with $K=50$ using (left) (\ref{eq:wangblei}) and (right) (\ref{eq:wangblei}) + log prior term.
}

\label{fig:wangbleieval}
\end{figure*}

}

\newcommand{\putStaircasePlots}{
\begin{landscape}
\begin{figure*}[h]

\begin{minipage}{\mpwsc}
\begin{center}
\includegraphics[width=0.99\linewidth]{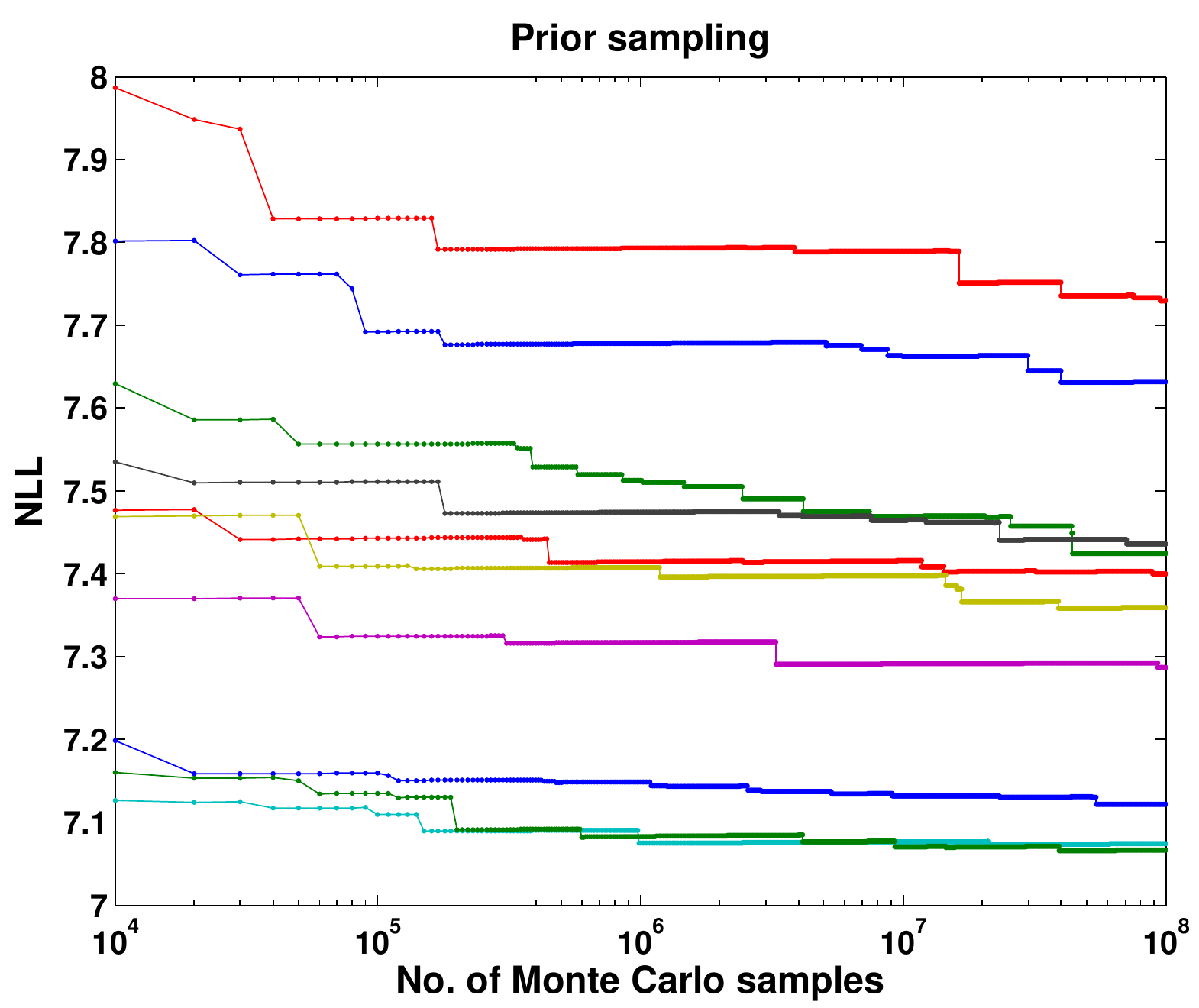}
\end{center}
\end{minipage}
\begin{minipage}{\mpwsc}
\begin{center}
\includegraphics[width=0.99\linewidth]{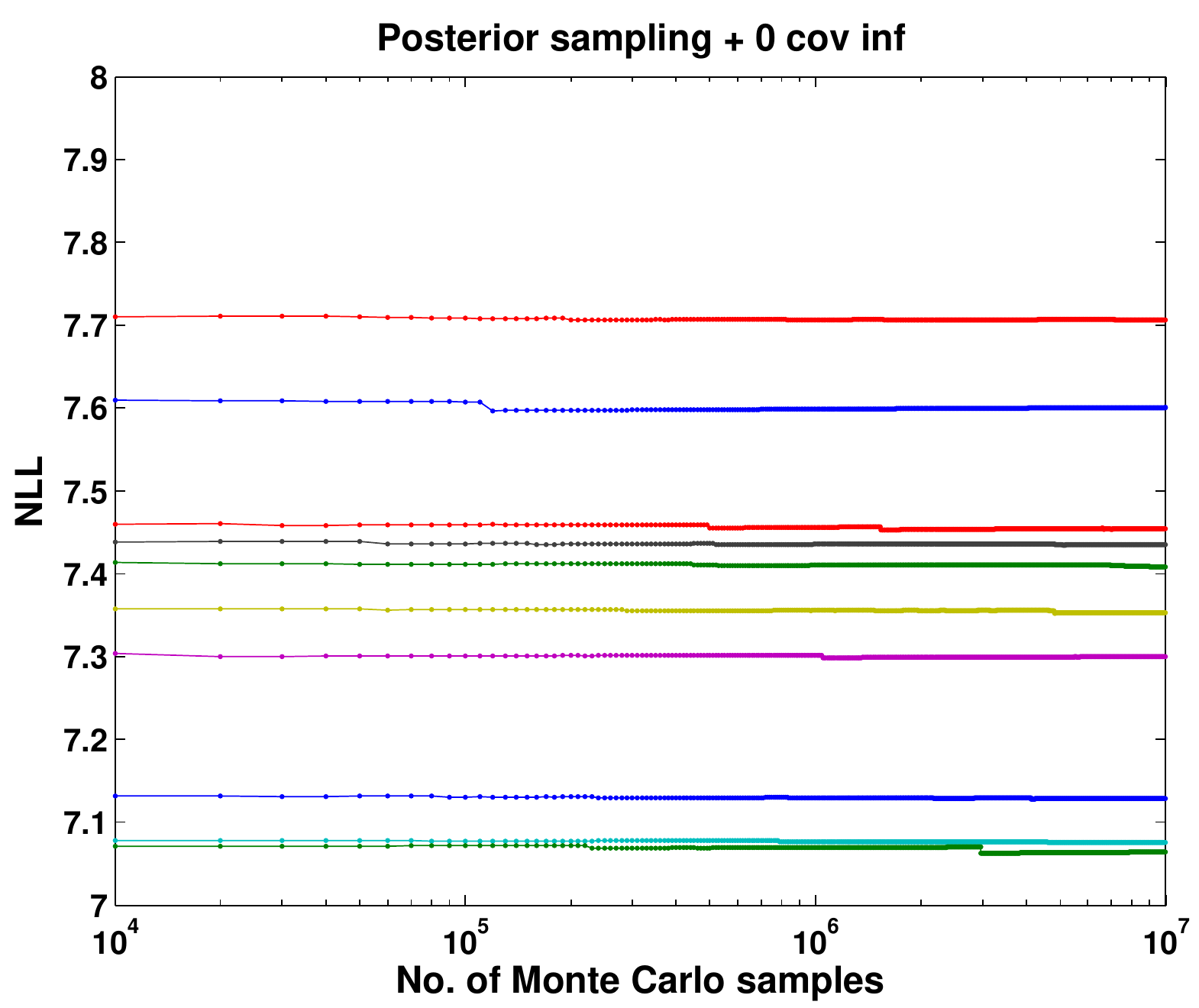}
\end{center}
\end{minipage}
\begin{minipage}{\mpwsc}
\begin{center}
\includegraphics[width=0.99\linewidth]{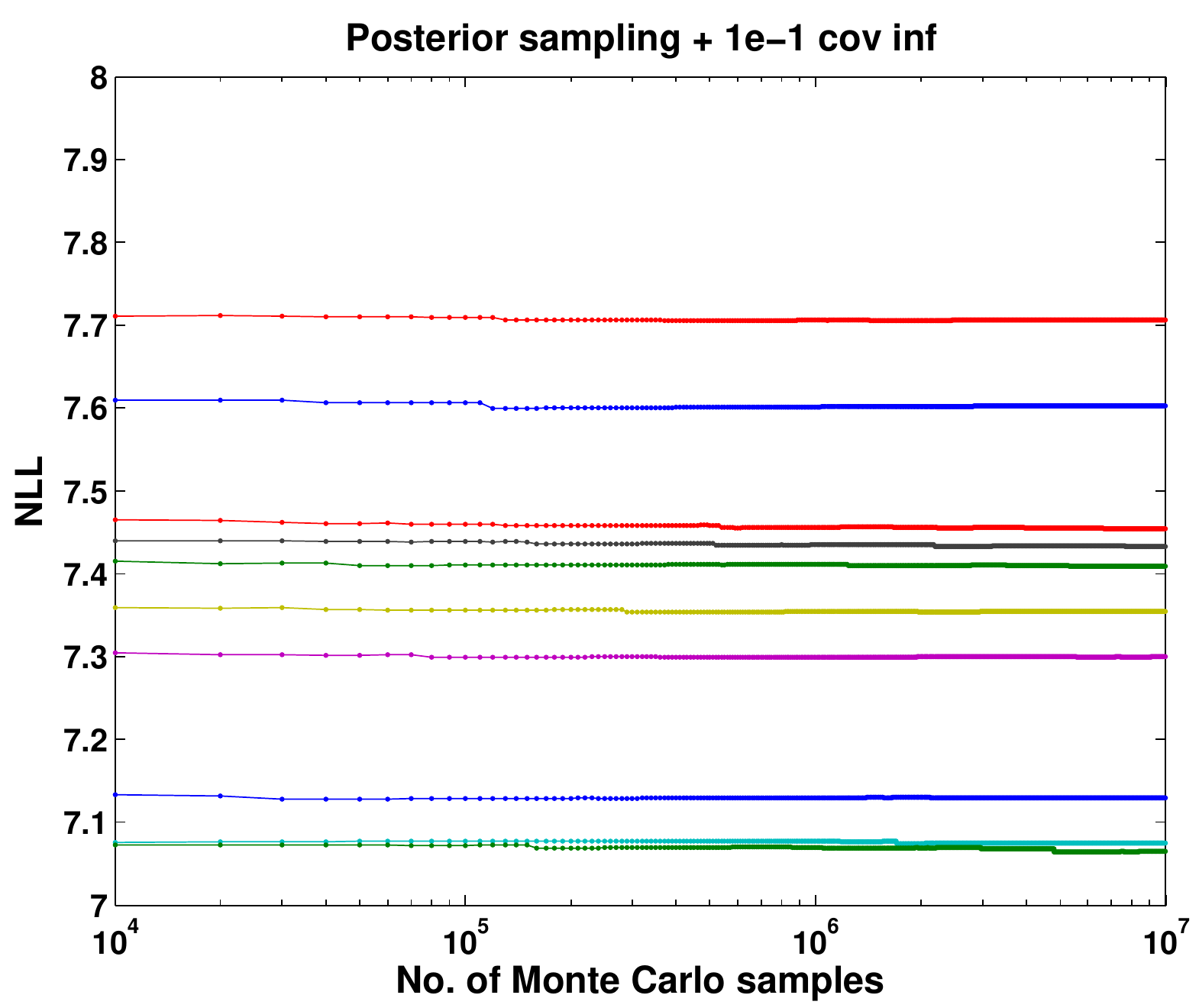}
\end{center}
\end{minipage}
\begin{minipage}{\mpwsc}
\begin{center}
\includegraphics[width=0.99\linewidth]{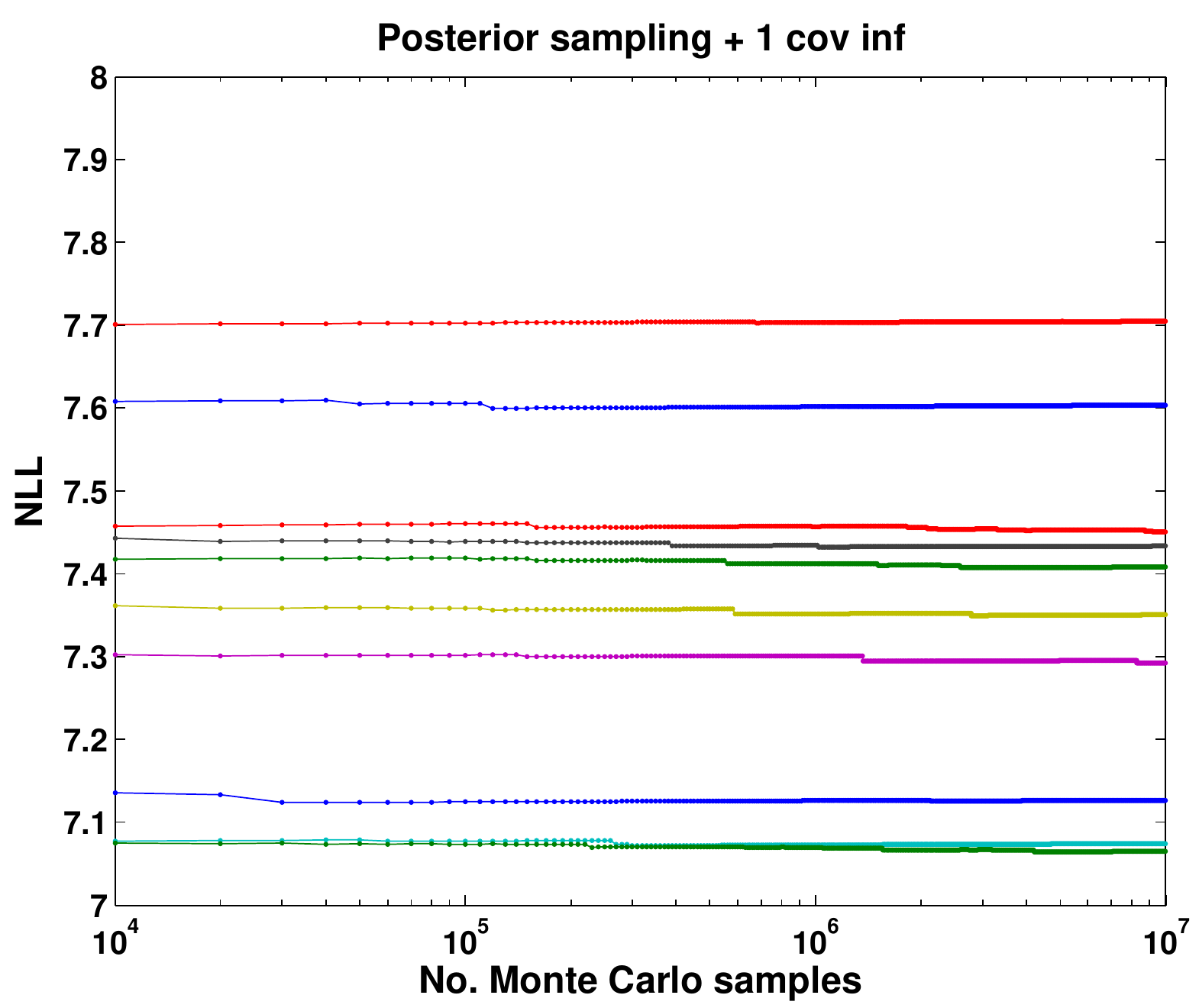}
\end{center}
\end{minipage}

\begin{minipage}{\mpwsc}
\begin{center}
\includegraphics[width=0.99\linewidth]{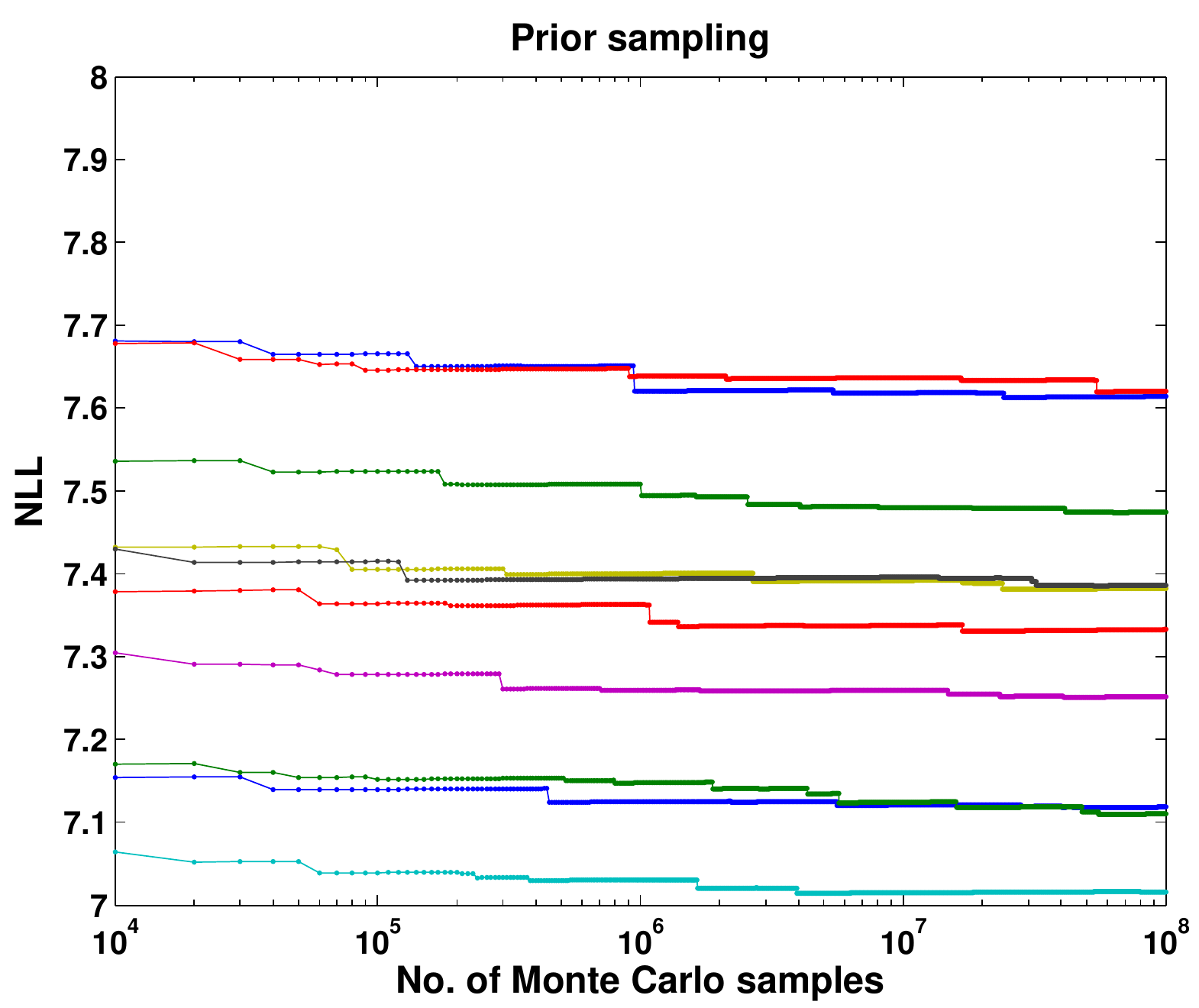}
\end{center}
\end{minipage}
\begin{minipage}{\mpwsc}
\begin{center}
\includegraphics[width=0.99\linewidth]{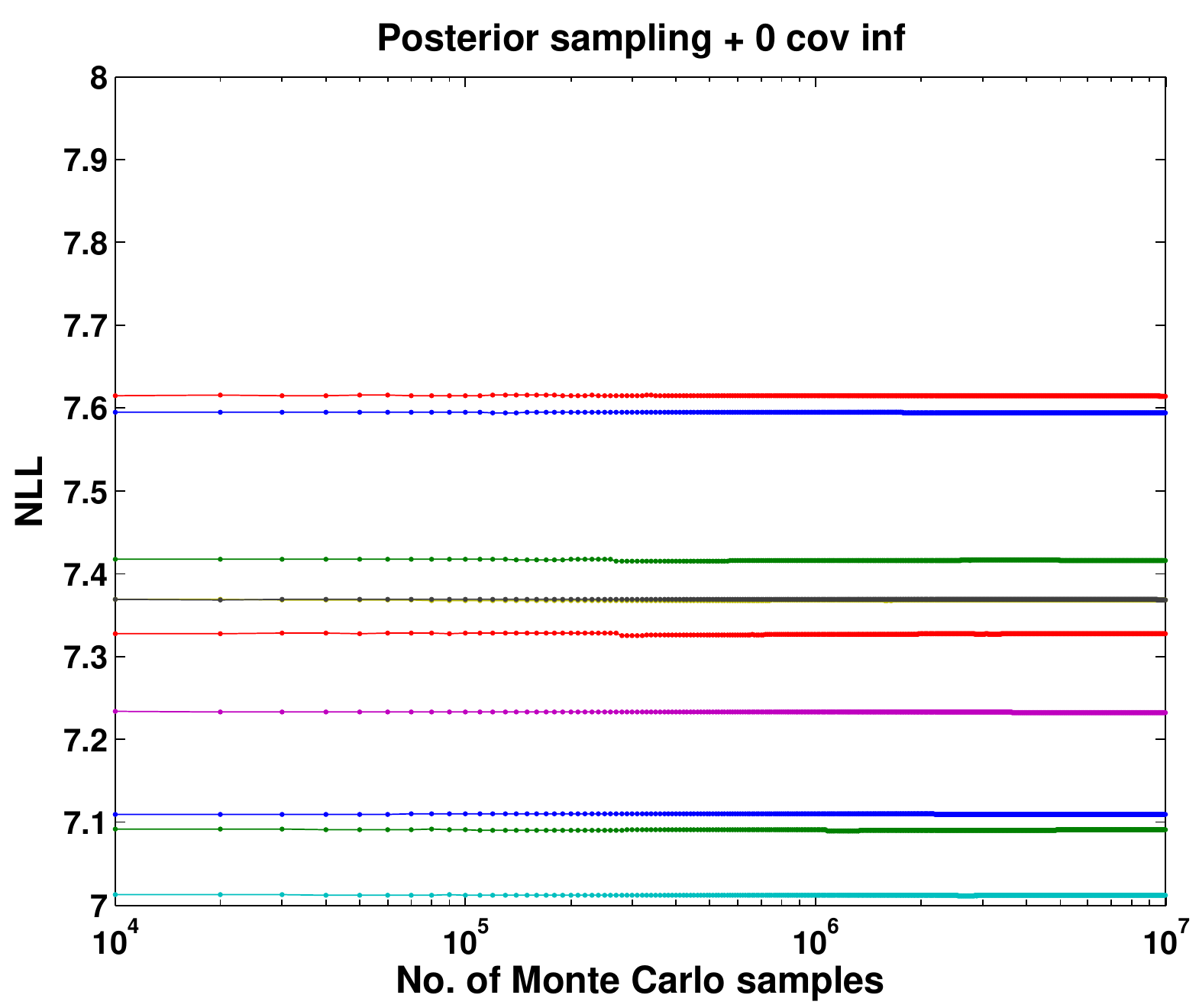}
\end{center}
\end{minipage}
\begin{minipage}{\mpwsc}
\begin{center}
\includegraphics[width=0.99\linewidth]{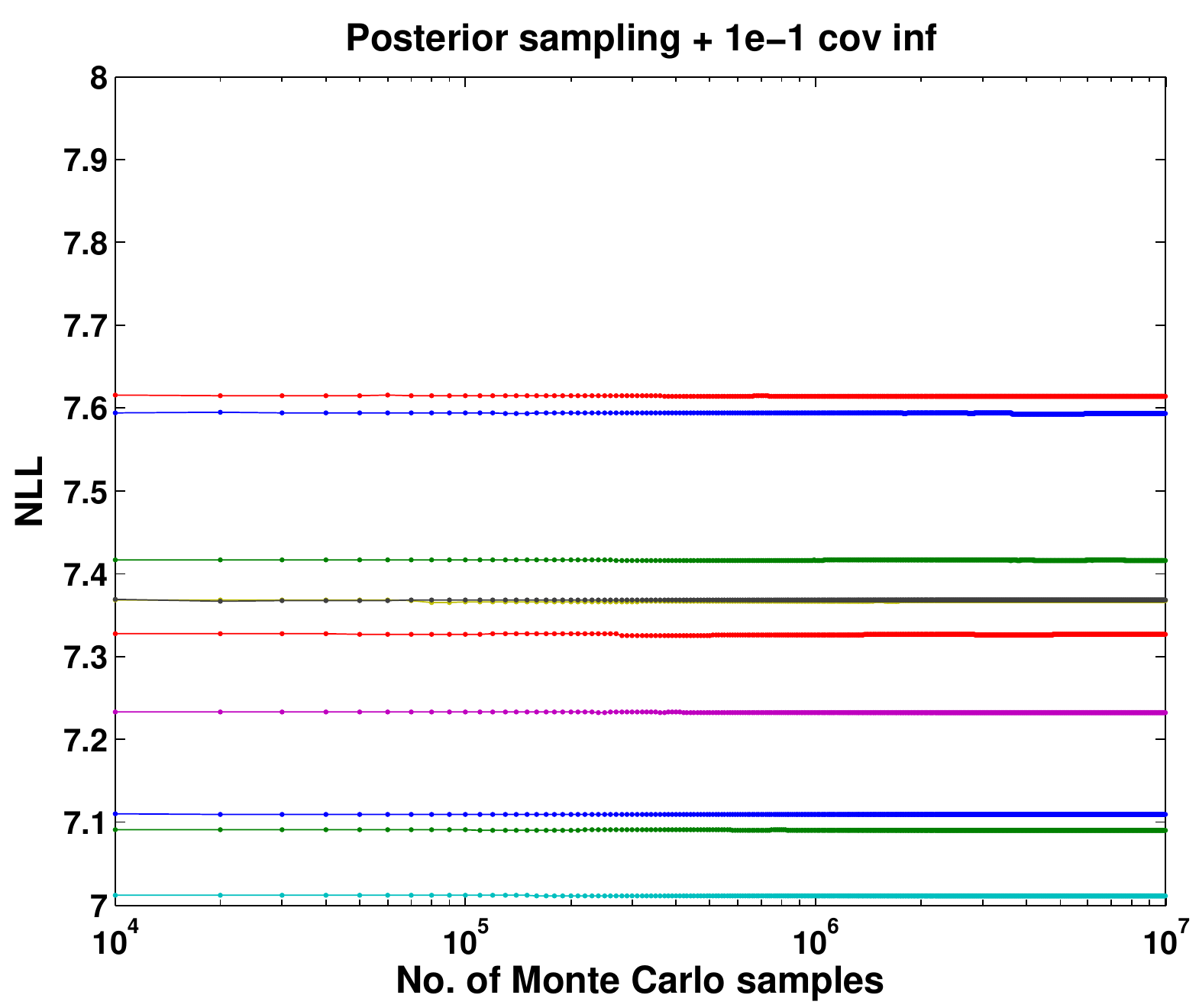}
\end{center}
\end{minipage}
\begin{minipage}{\mpwsc}
\begin{center}
\includegraphics[width=0.99\linewidth]{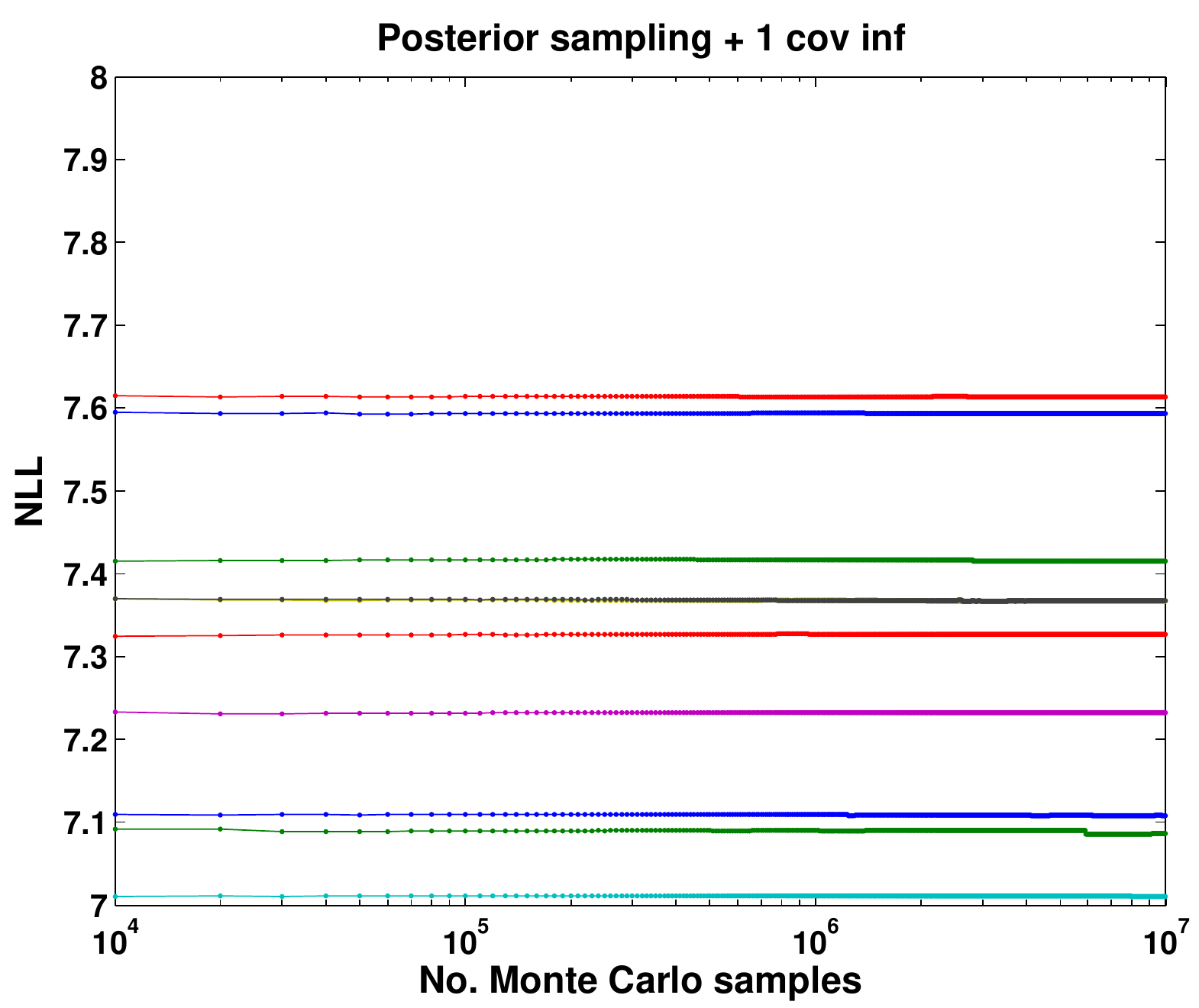}
\end{center}
\end{minipage}

\begin{minipage}{\mpwsc}
\begin{center}
\includegraphics[width=0.99\linewidth]{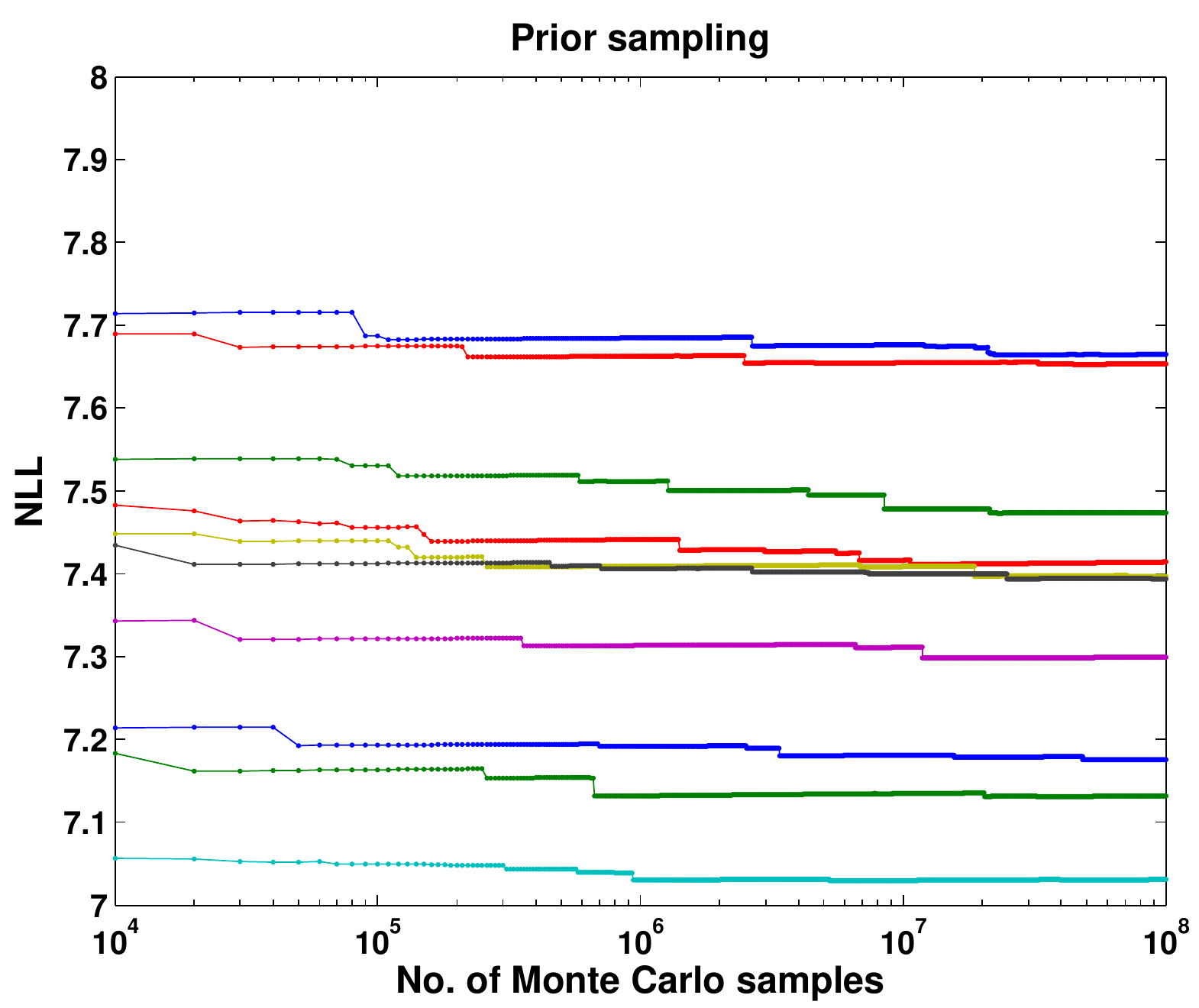}
\end{center}
\end{minipage}
\begin{minipage}{\mpwsc}
\begin{center}
\includegraphics[width=0.99\linewidth]{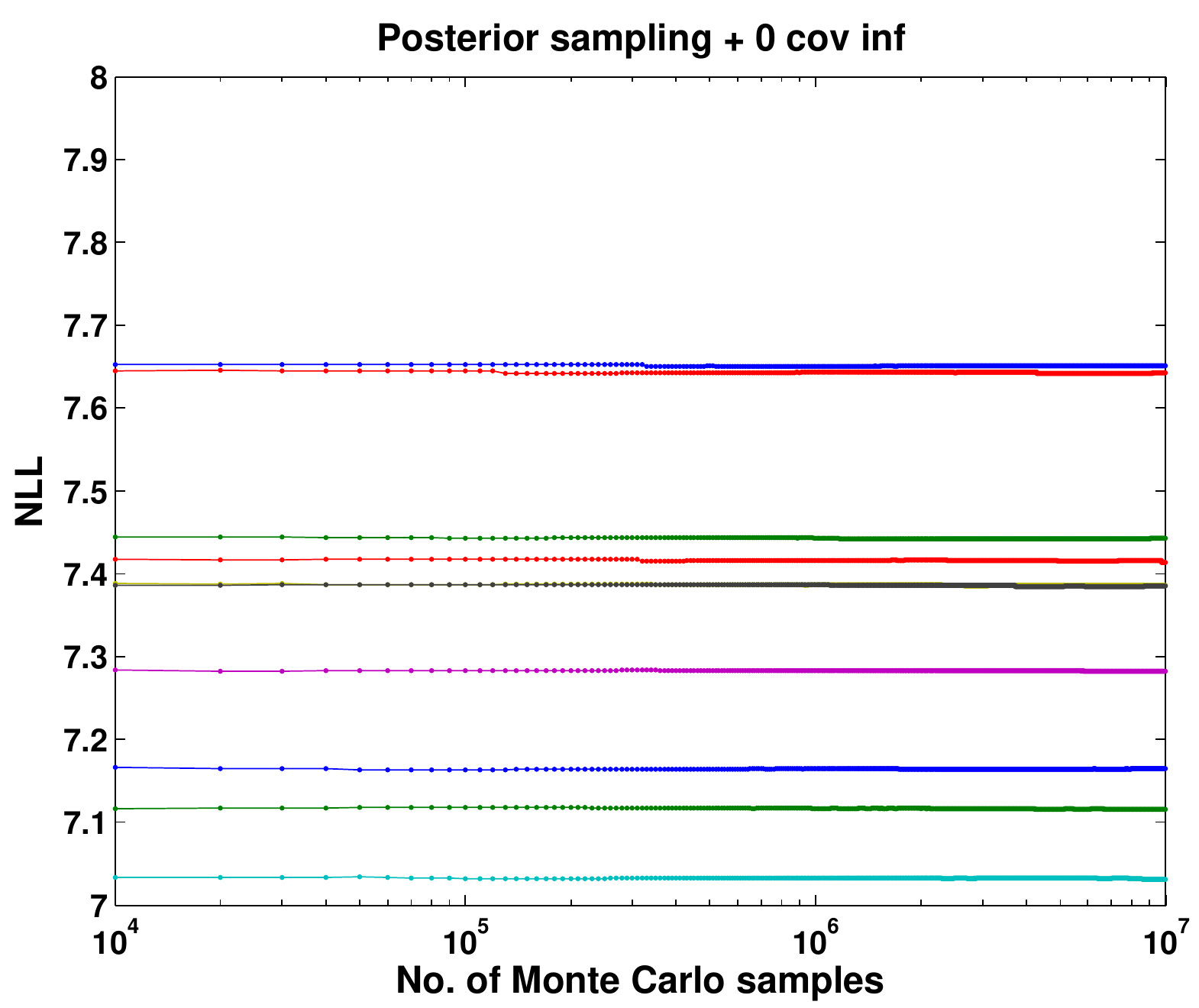}
\end{center}
\end{minipage}
\begin{minipage}{\mpwsc}
\begin{center}
\includegraphics[width=0.99\linewidth]{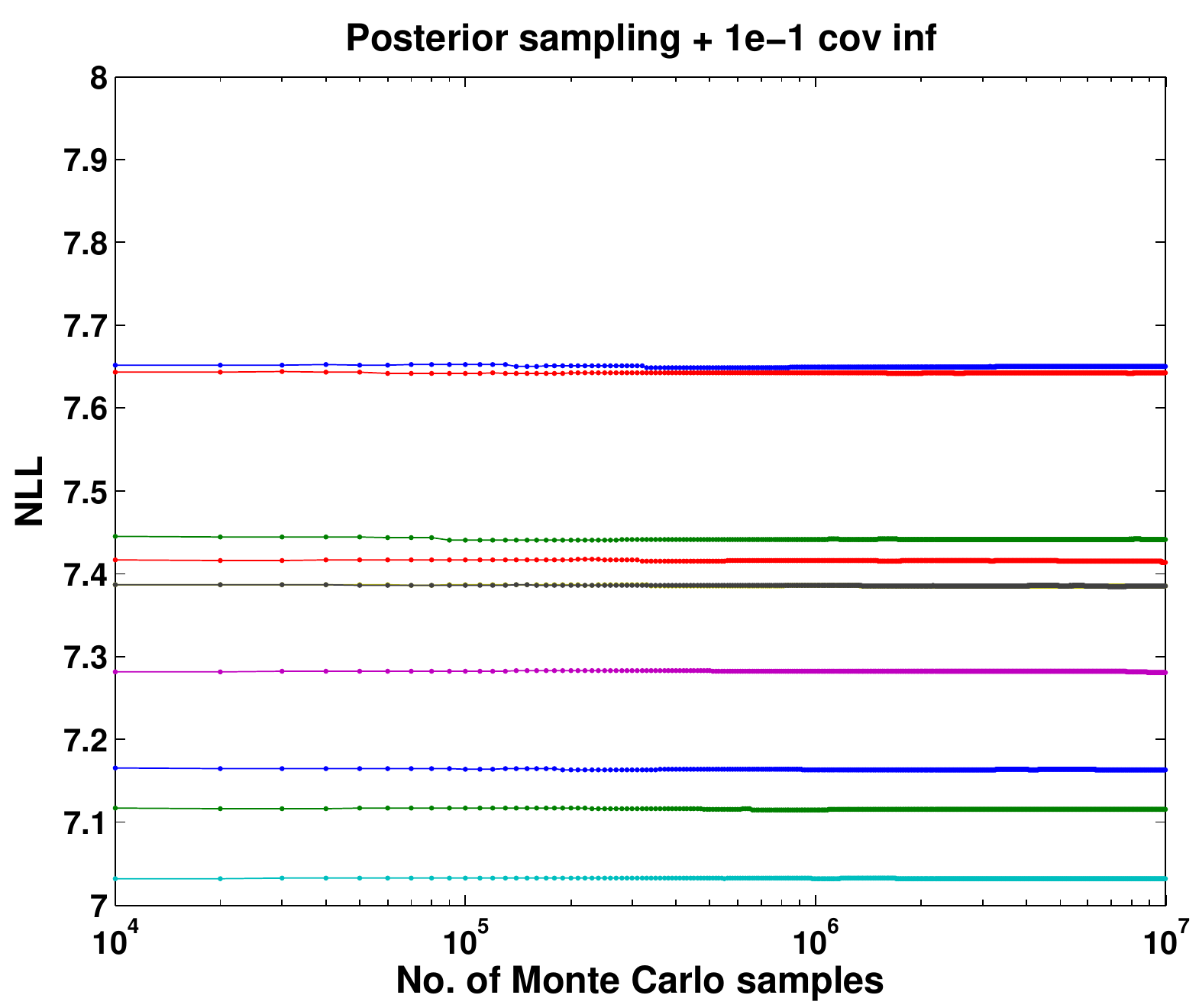}
\end{center}
\end{minipage}
\begin{minipage}{\mpwsc}
\begin{center}
\includegraphics[width=0.99\linewidth]{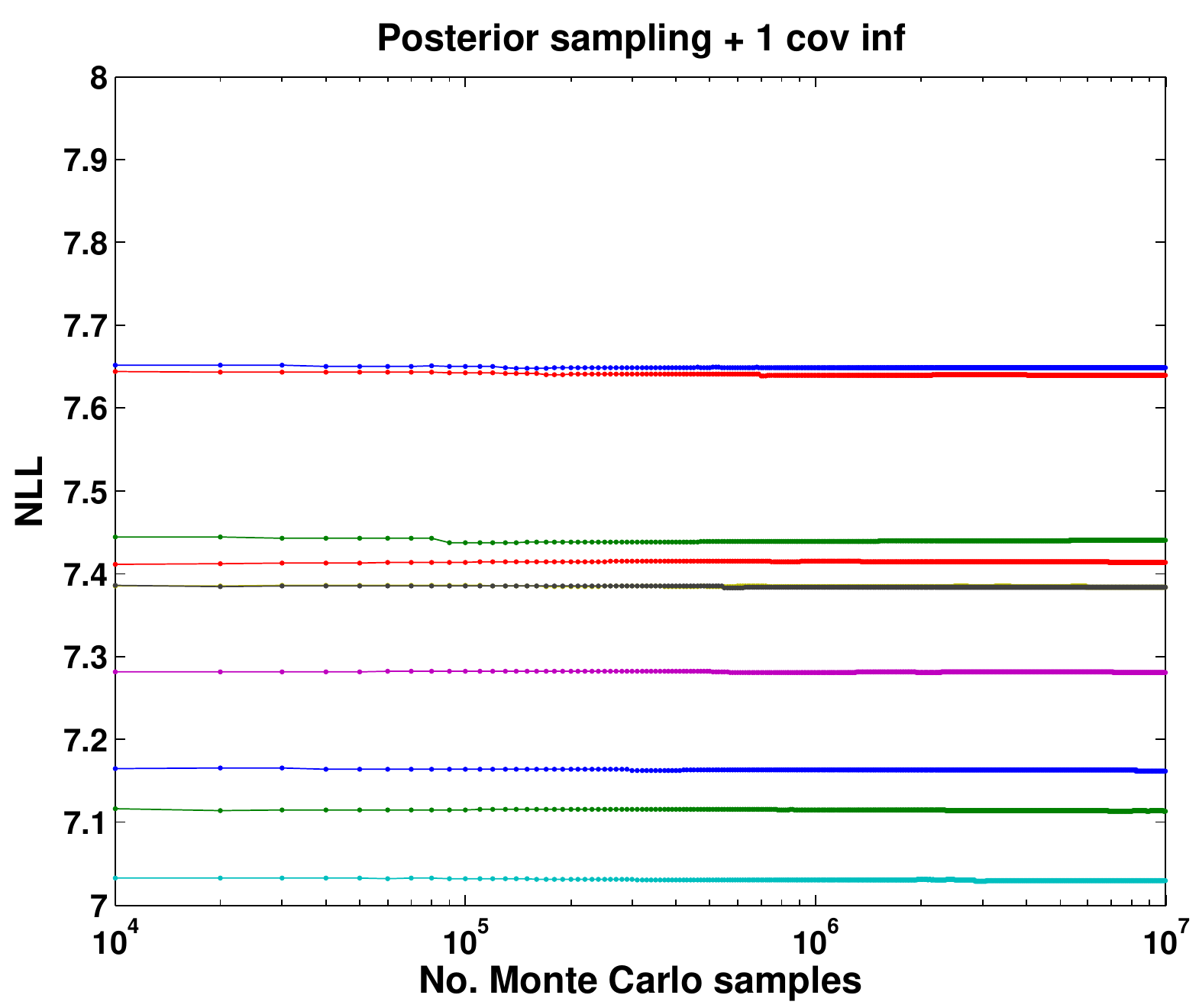}
\end{center}
\end{minipage}

\caption{
CTM normalized NLL values (\ref{eq:ctmnll}) of first 10 test documents of \emph{nips} with $K=50$ as a function of algorithm (rows) and importance sampling scheme (columns).
Top row represents mean field. 
Middle row represents H-MC-SSVI. 
Bottow row represents S-DSVI. 
First column represents prior sampling. 
Second column represents posterior sampling. 
Third column represents posterior sampling + $10^{-1} I$ covariance inflation. 
Fourth column represents posterior sampling + $I$ covariance inflation. 
Each color within a subplot represents a different document and the color coding is consistent across subplots.
All calculations performed with final learned parameters. 
Test document posteriors calculated with 100 iterations of H-MC-SSVI.
}

\label{fig:staircaseplots}
\end{figure*}
\end{landscape}

}

\newcommand{\putCTMKPlots}{
\begin{figure*}[h]

\begin{minipage}{\mpw}
\begin{center}
\includegraphics[width=0.99\linewidth]{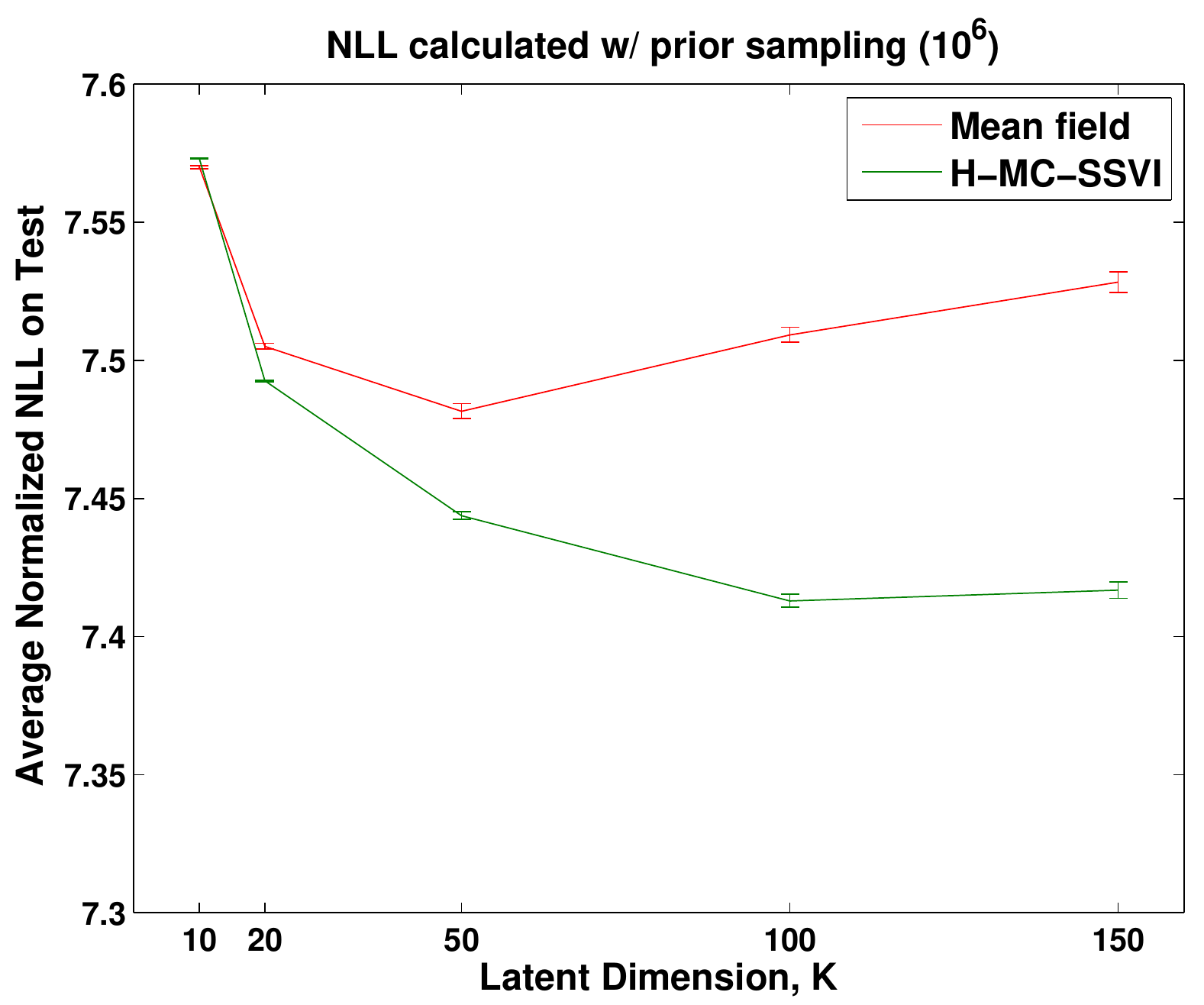}
\end{center}
\end{minipage}
\begin{minipage}{\mpw}
\begin{center}
\includegraphics[width=0.99\linewidth]{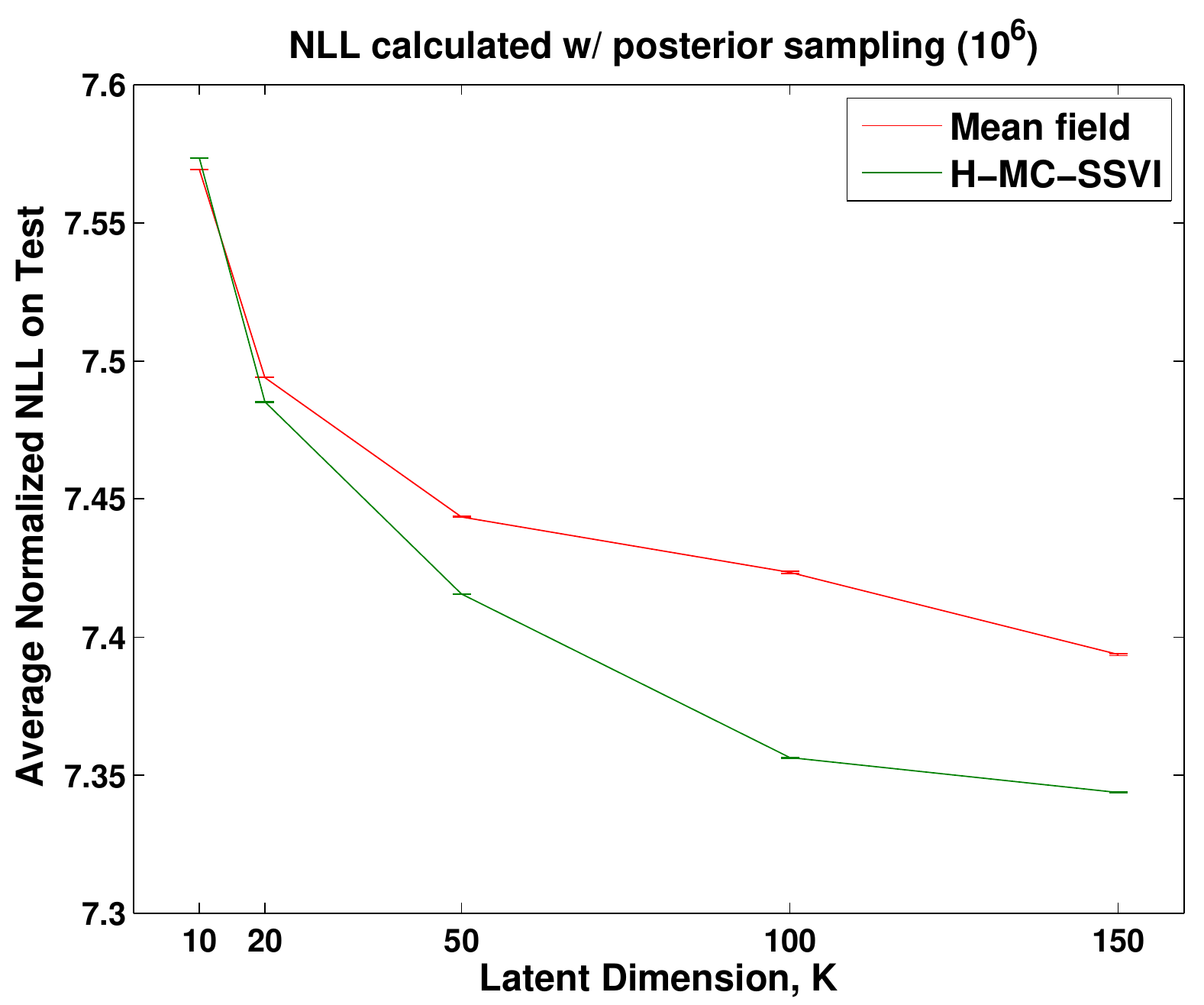}
\end{center}
\end{minipage}

\vspace{2mm}

\begin{minipage}{\mpw}
\begin{center}
\includegraphics[width=0.99\linewidth]{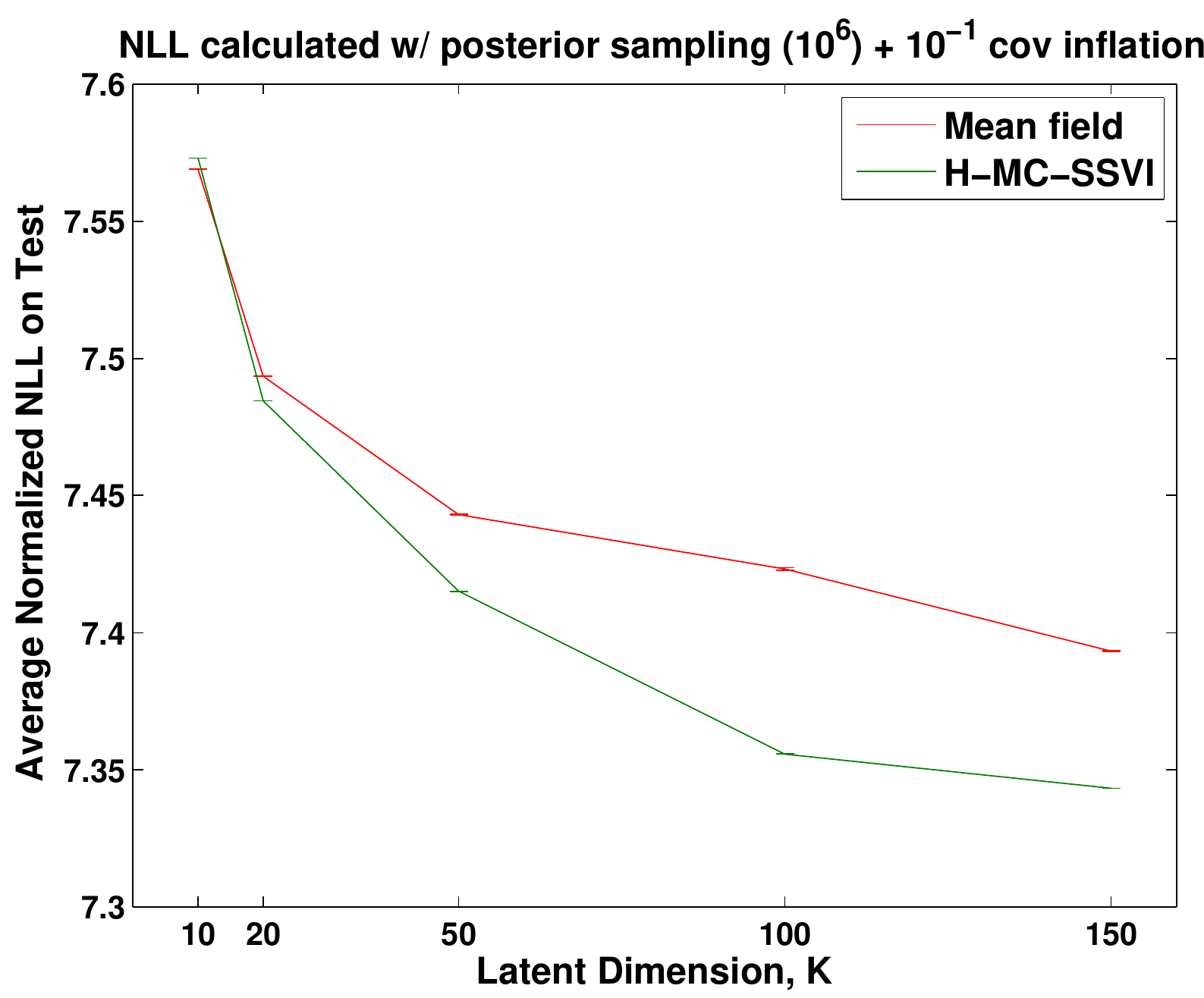}
\end{center}
\end{minipage}
\begin{minipage}{\mpw}
\begin{center}
\includegraphics[width=0.99\linewidth]{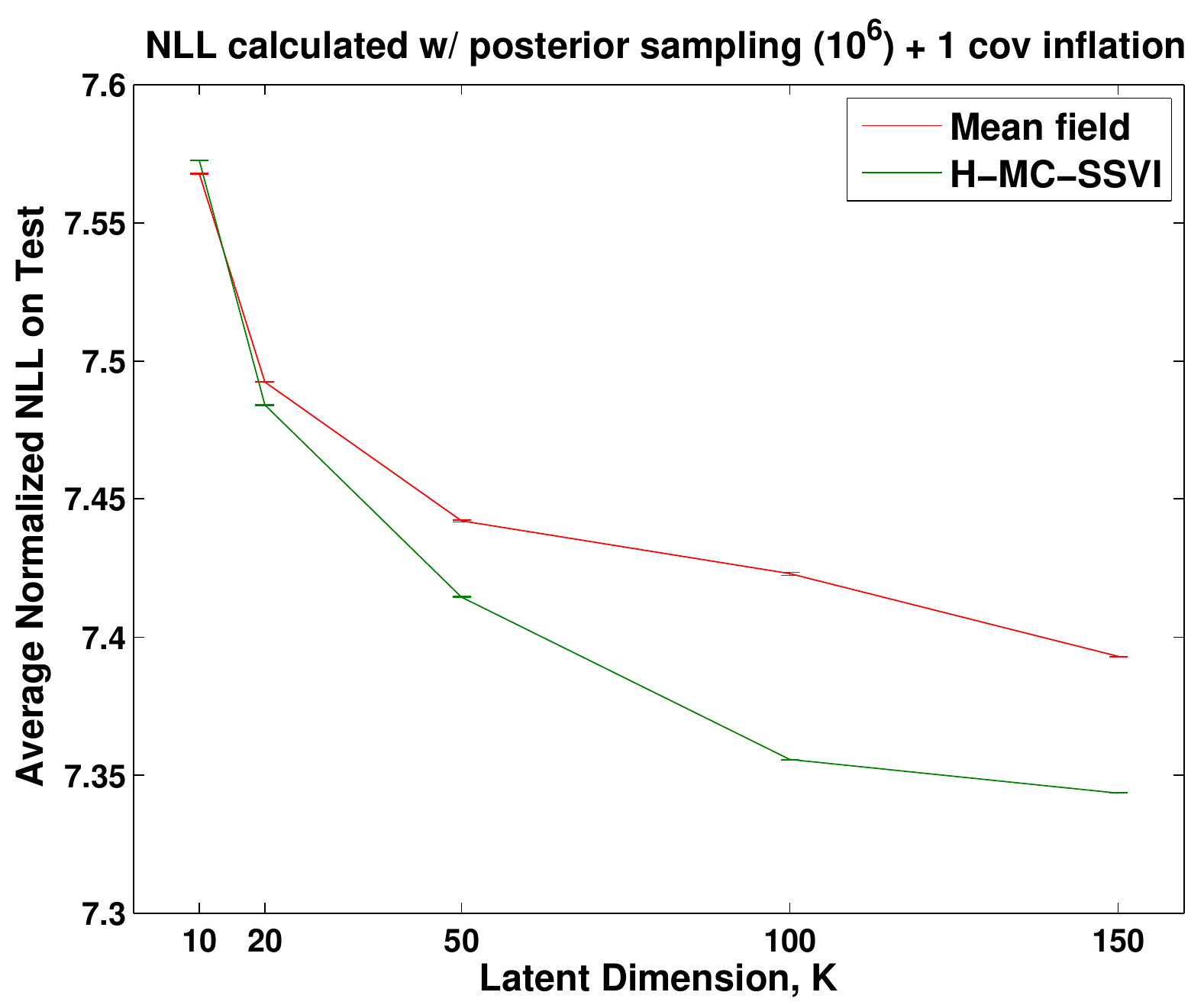}
\end{center}
\end{minipage}

\caption{
Performance in CTM on \emph{nips} dataset as a function of latent dimensionality. Test set normalized NLL was computed using the four variants described in Section \ref{sec:appendix_exp_details}. 
}
\label{fig:ctmkplots}
\end{figure*}
}

\newcommand{\putCTMPlotsNLL}{
\begin{figure*}[h]

\begin{minipage}{\mpw}
\begin{center}
\includegraphics[width=0.99\linewidth]{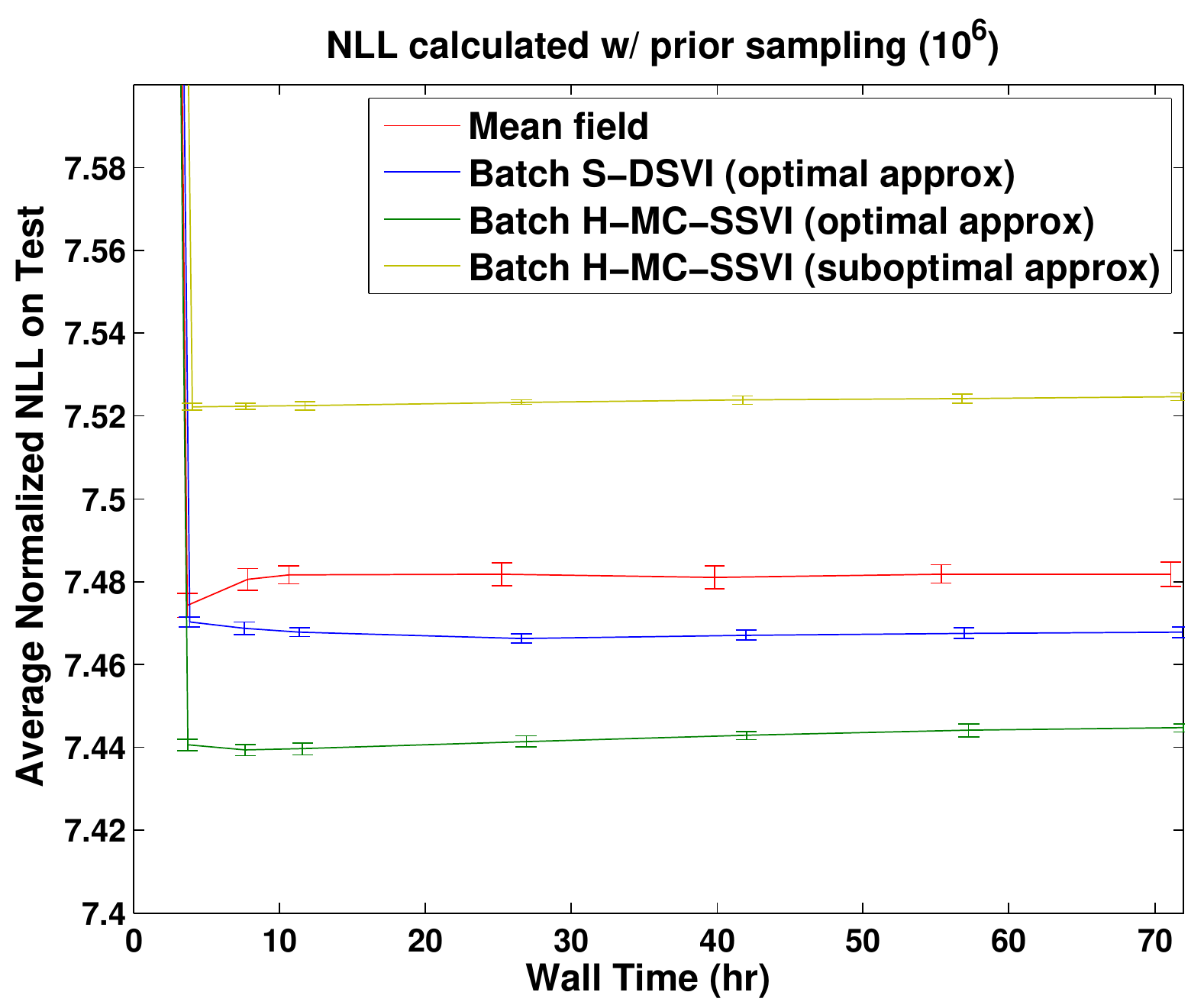}
\end{center}
\end{minipage}
\begin{minipage}{\mpw}
\begin{center}
\includegraphics[width=0.99\linewidth]{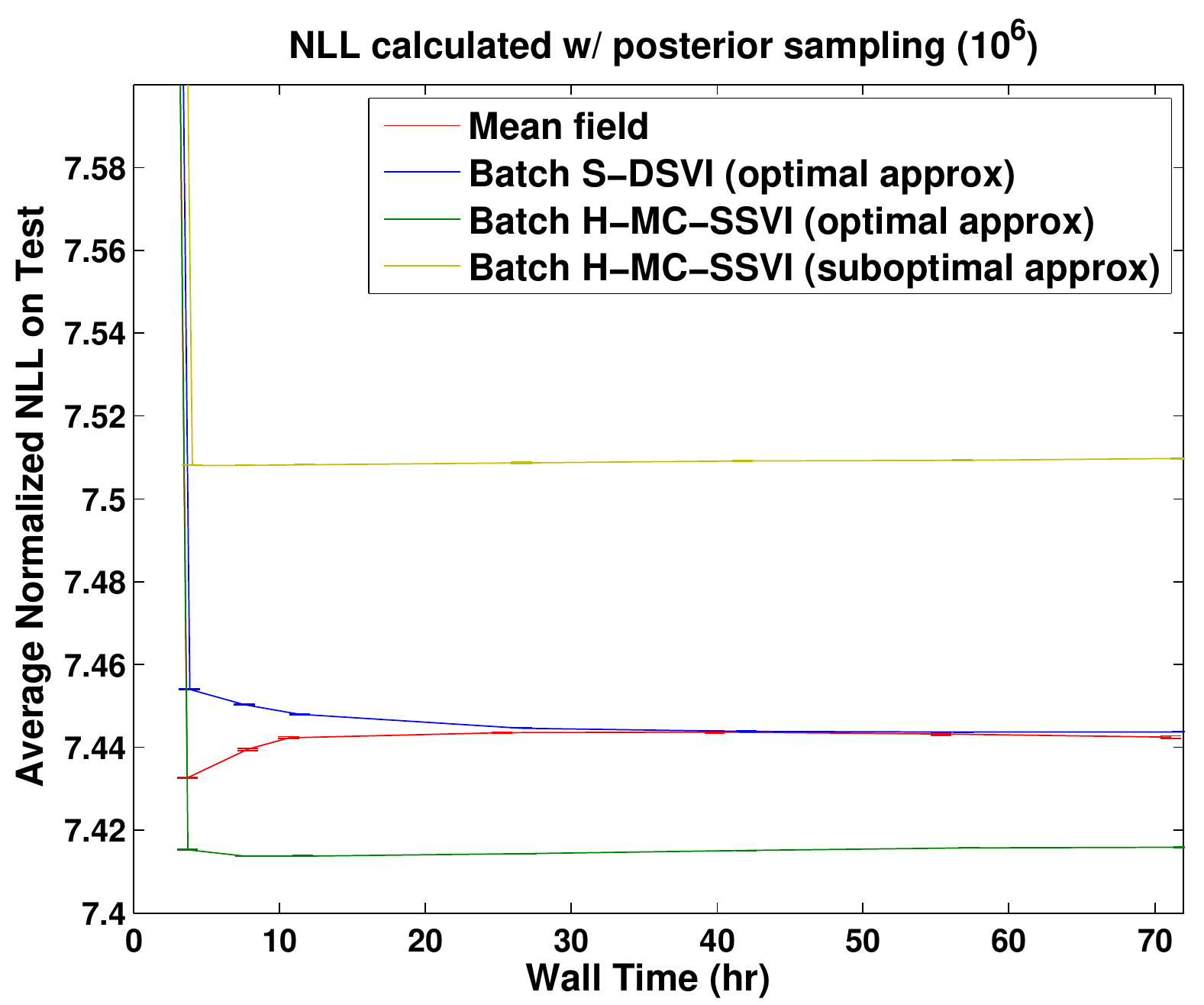}
\end{center}
\end{minipage}

\vspace{2mm}

\begin{minipage}{\mpw}
\begin{center}
\includegraphics[width=0.99\linewidth]{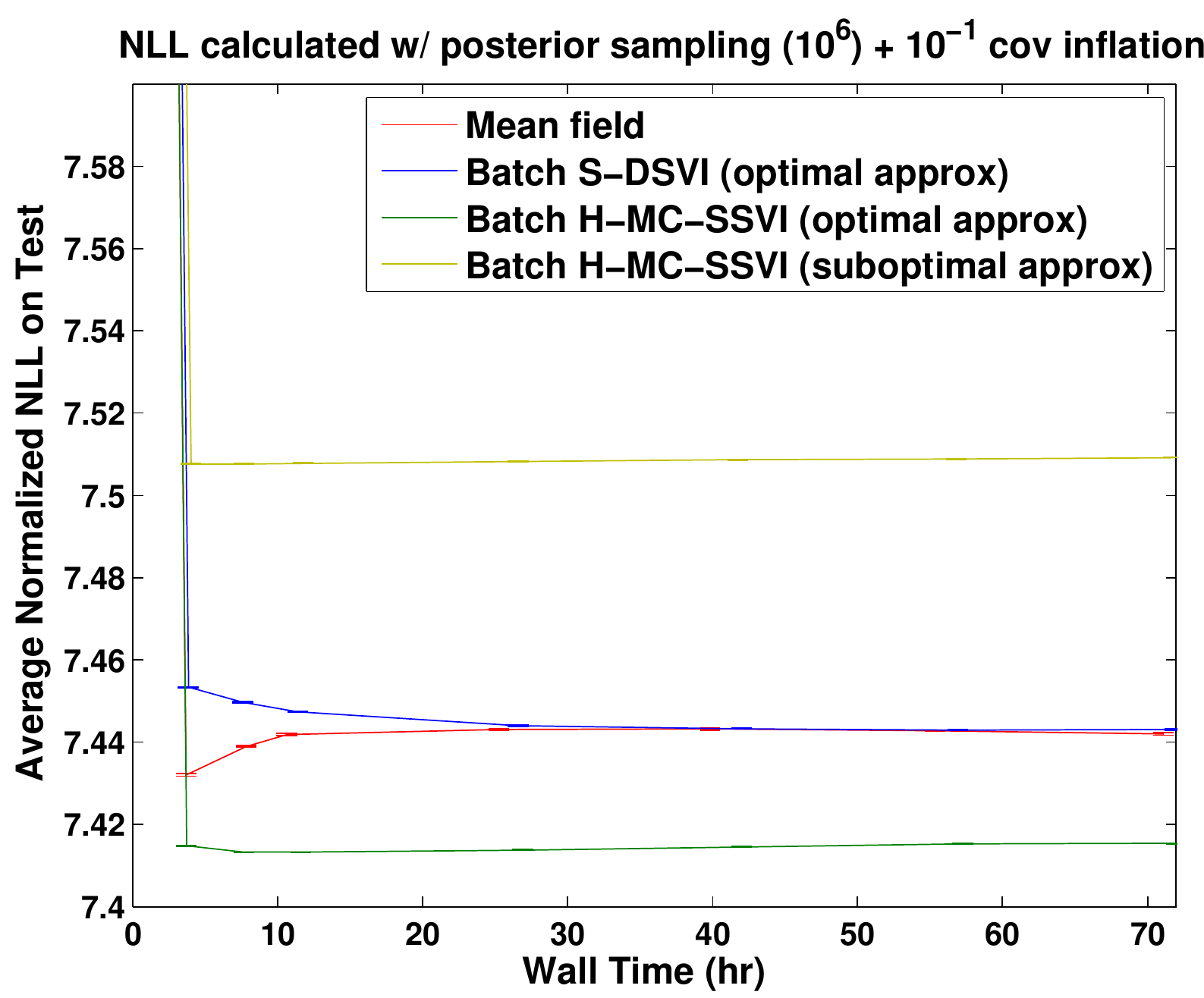}
\end{center}
\end{minipage}
\begin{minipage}{\mpw}
\begin{center}
\includegraphics[width=0.99\linewidth]{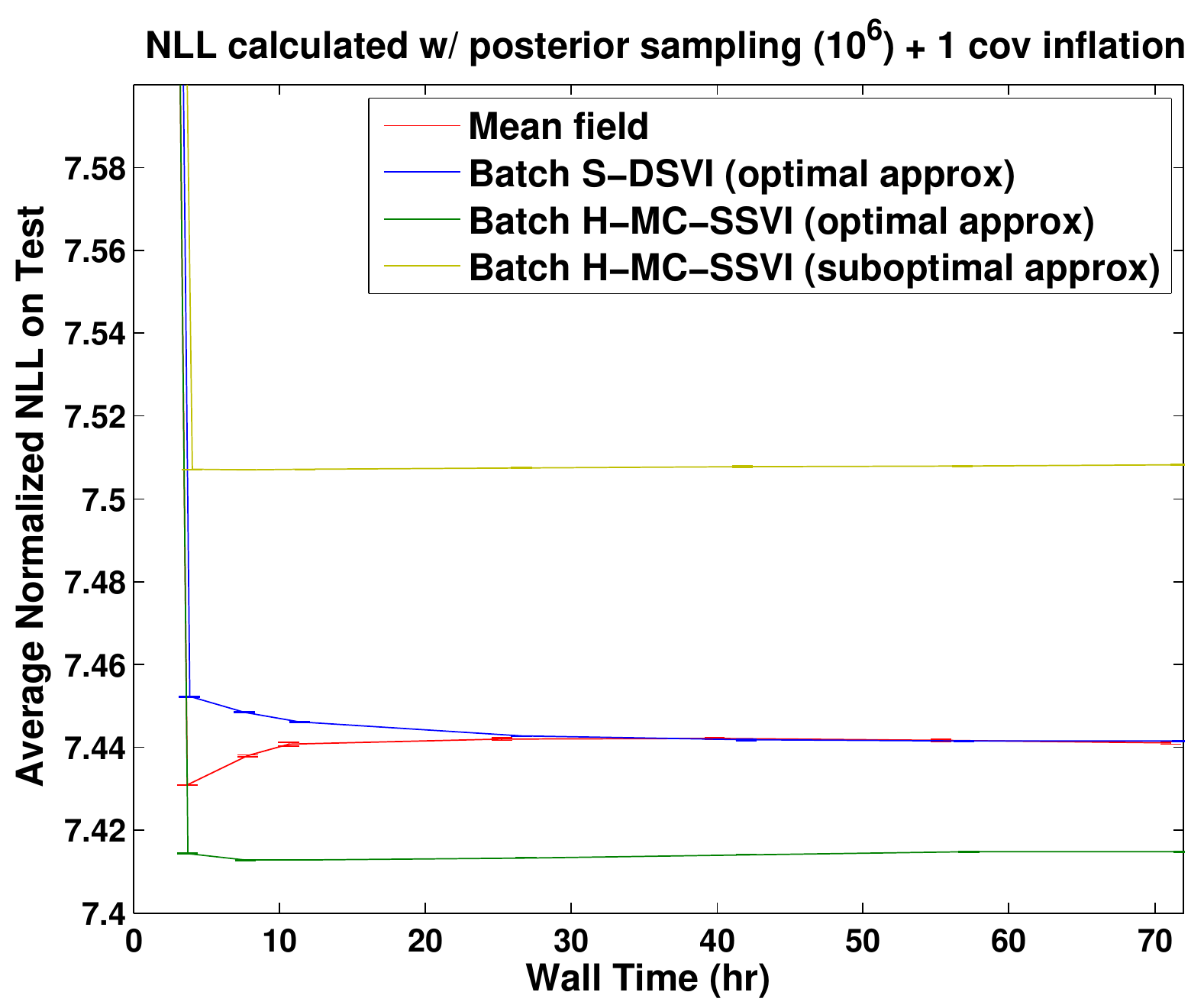}
\end{center}
\end{minipage}

\caption{
Test set performance of different batch variational approximations for CTM (with diagonal covariances). Test set normalized NLL was computed using the four variants described in Section \ref{sec:appendix_exp_details}. 
}
\label{fig:ctmplotsnll}
\end{figure*}
}

\newcommand{\putPMFPlotsErrB}{
\begin{figure*}[t]

\begin{center}
\includegraphics[width=0.45\linewidth]{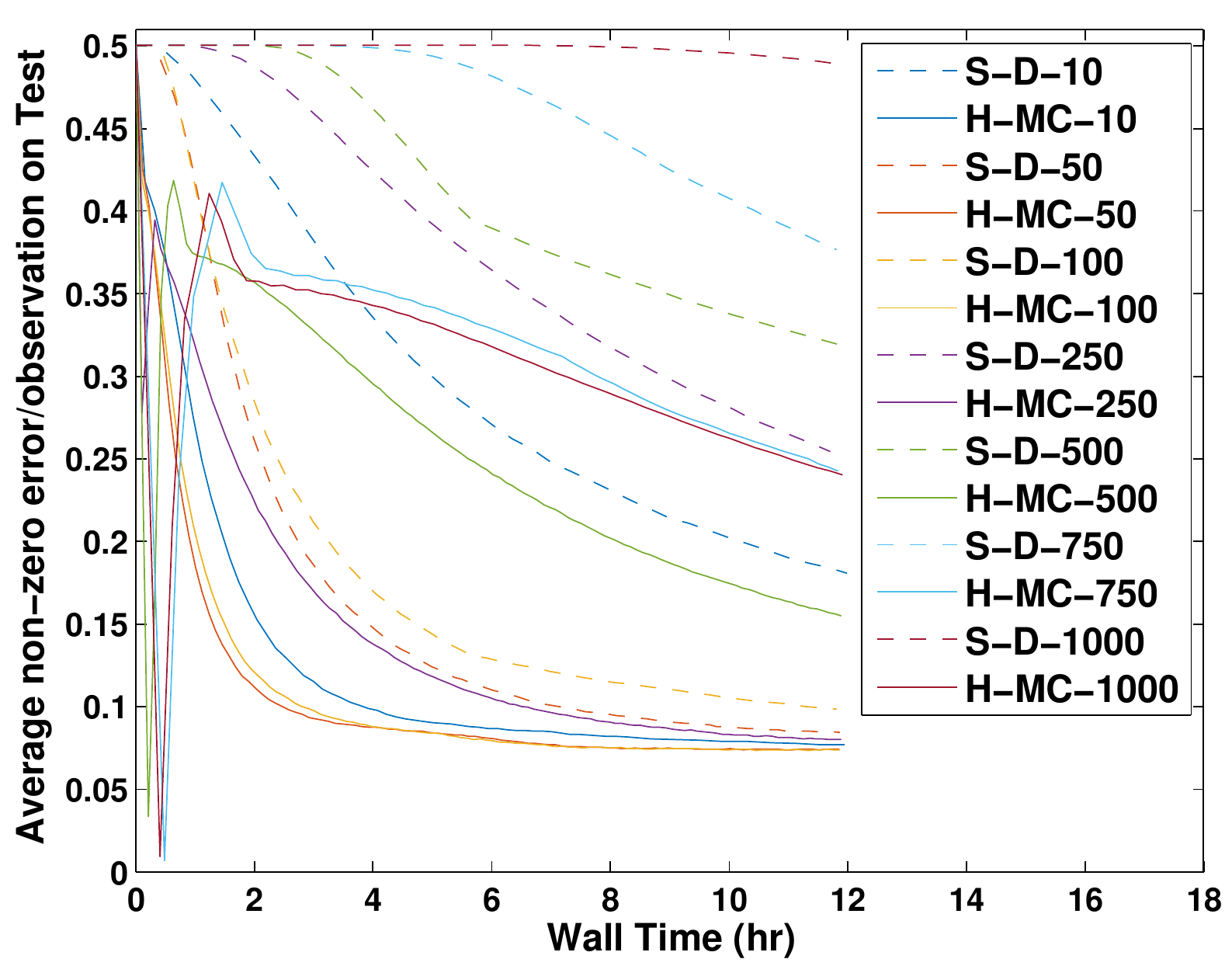}
\end{center}

\caption{
Additional error metric for the artificial dataset in PMF with count likelihood. S-D denotes S-DSVI and H-MC denotes H-MC-SSVI. 
}
\label{fig:pmfplotserrb}
\end{figure*}
}

\newcommand{\putPMFPlotsErr}[2]{
\begin{figure*}[p]

\begin{minipage}{\mpw}
\begin{center}
\includegraphics[width=0.99\linewidth]{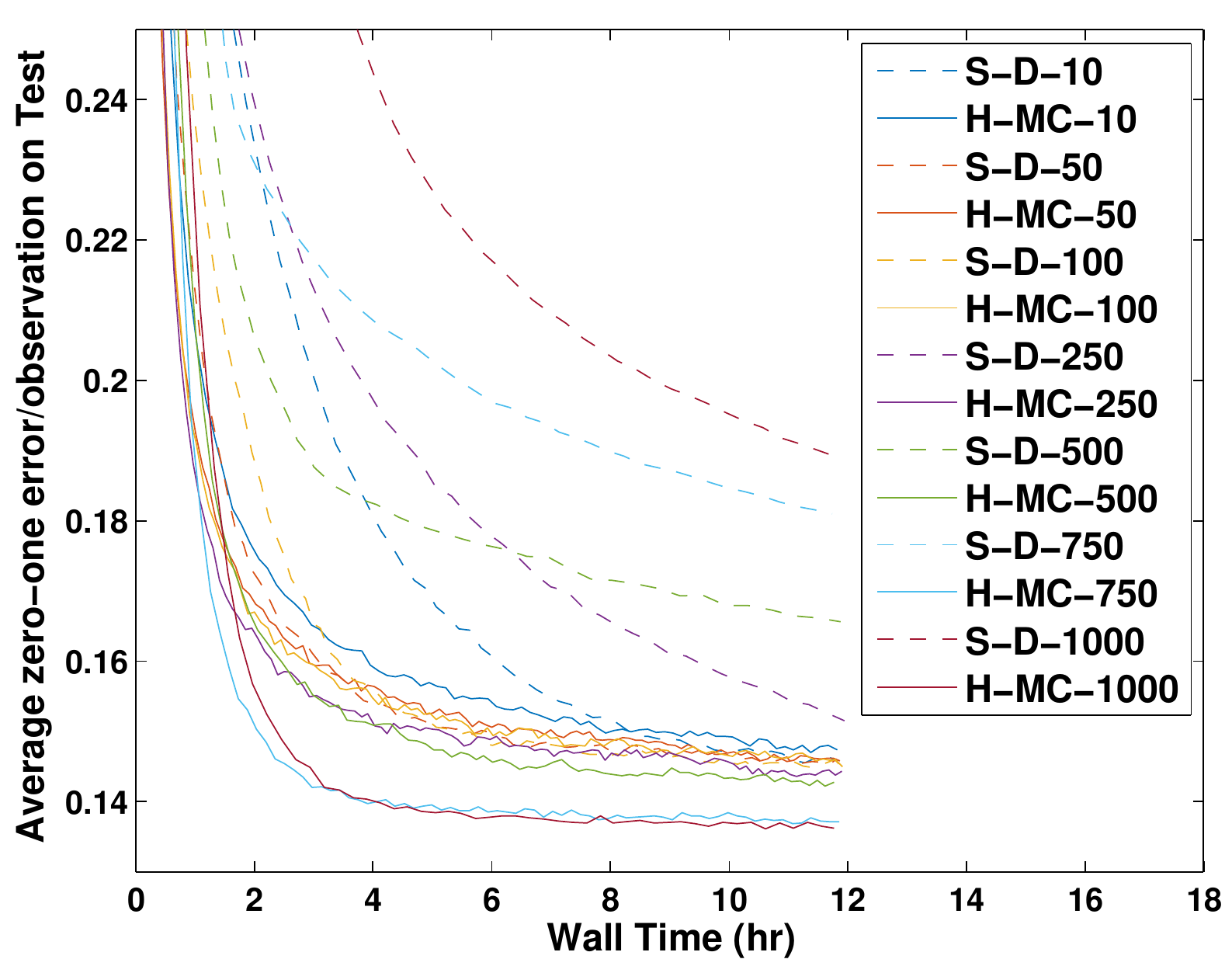}
\end{center}
\end{minipage}
\begin{minipage}{\mpw}
\begin{center}
\includegraphics[width=0.99\linewidth]{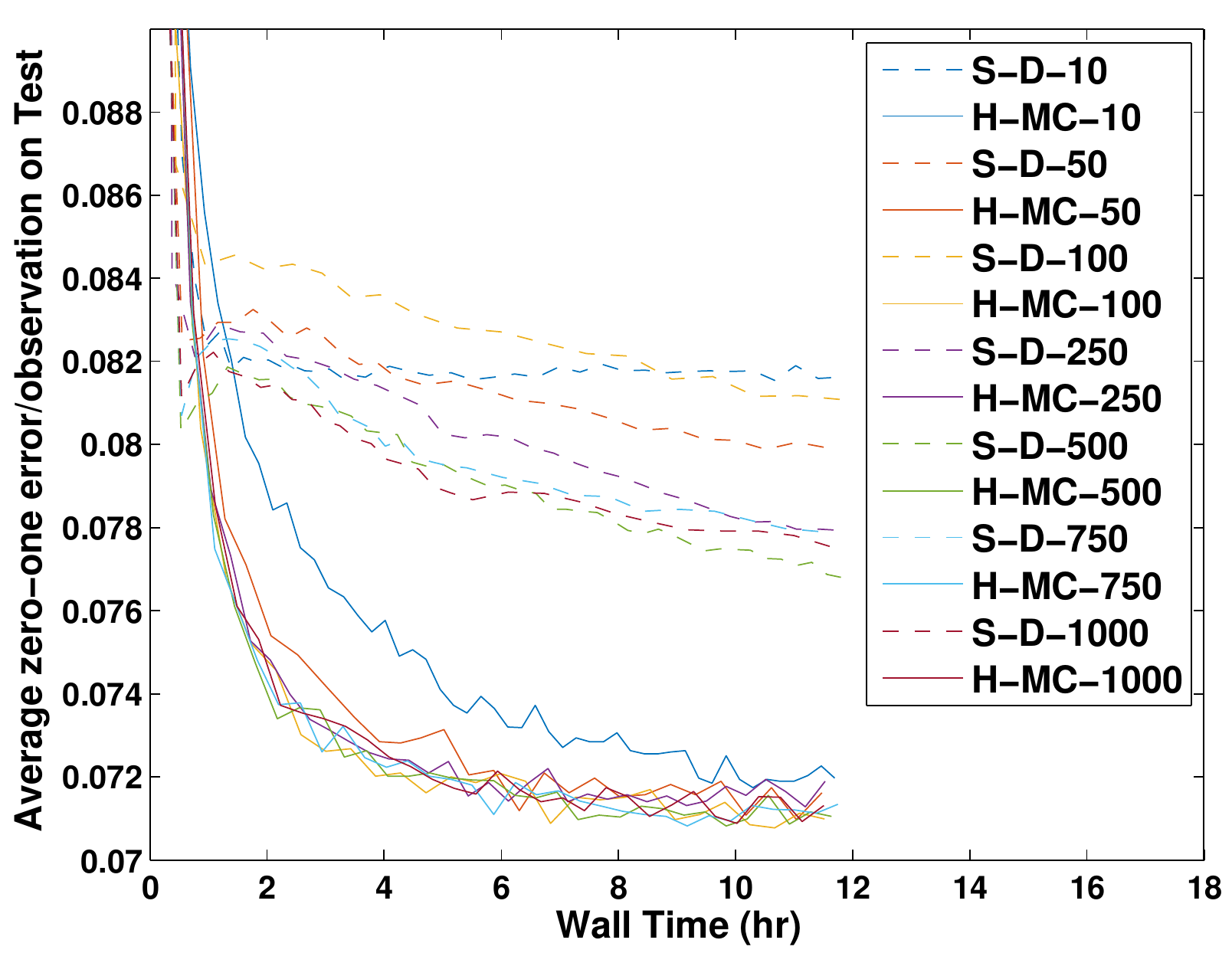}
\end{center}
\end{minipage}

\vspace{2mm}

\begin{minipage}{\mpw}
\begin{center}
\includegraphics[width=0.99\linewidth]{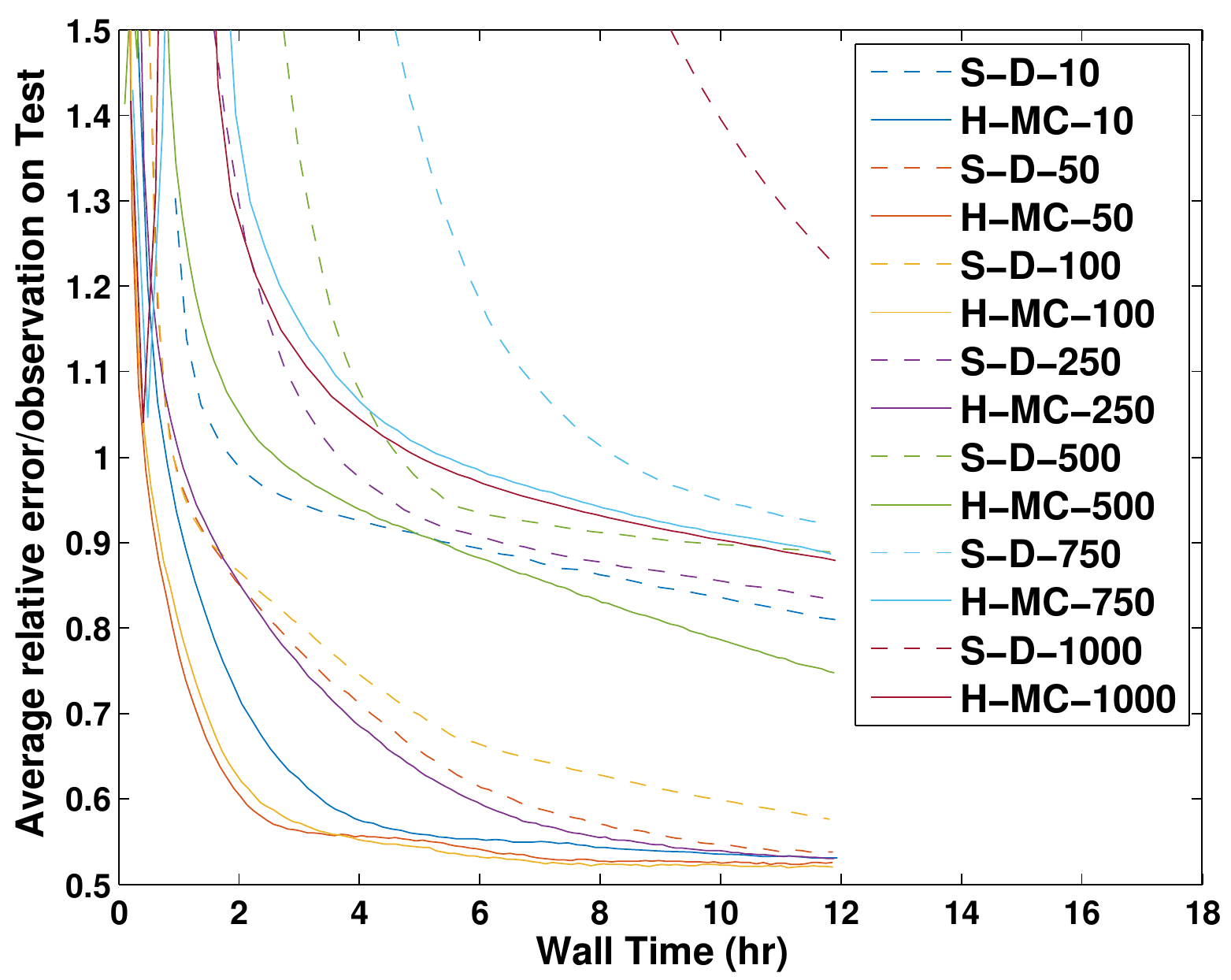}
\end{center}
\end{minipage}
\begin{minipage}{\mpw}
\begin{center}
\includegraphics[width=0.99\linewidth]{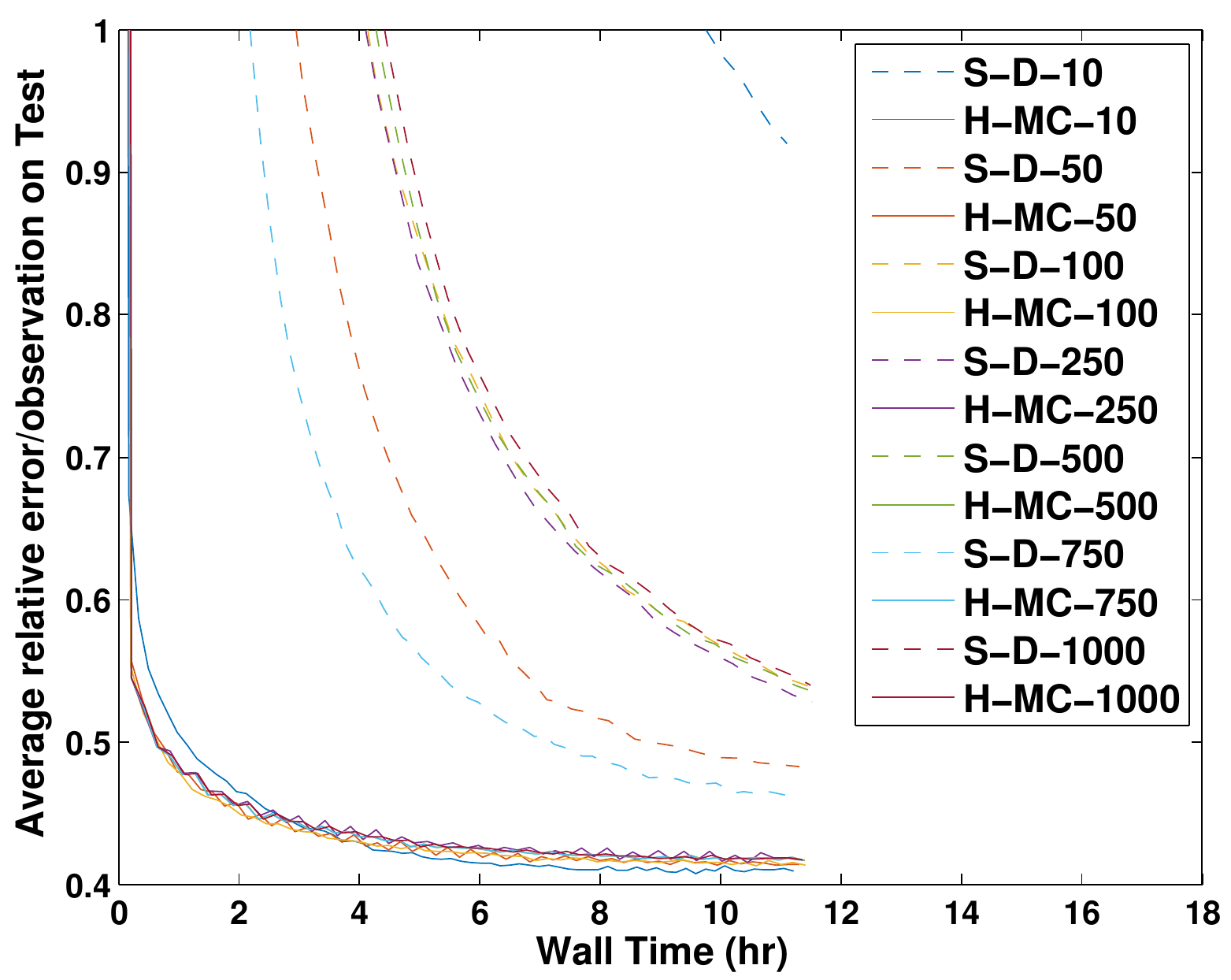}
\end{center}
\end{minipage}

\vspace{2mm}

\begin{minipage}{\mpw}
\begin{center}
\includegraphics[width=0.99\linewidth]{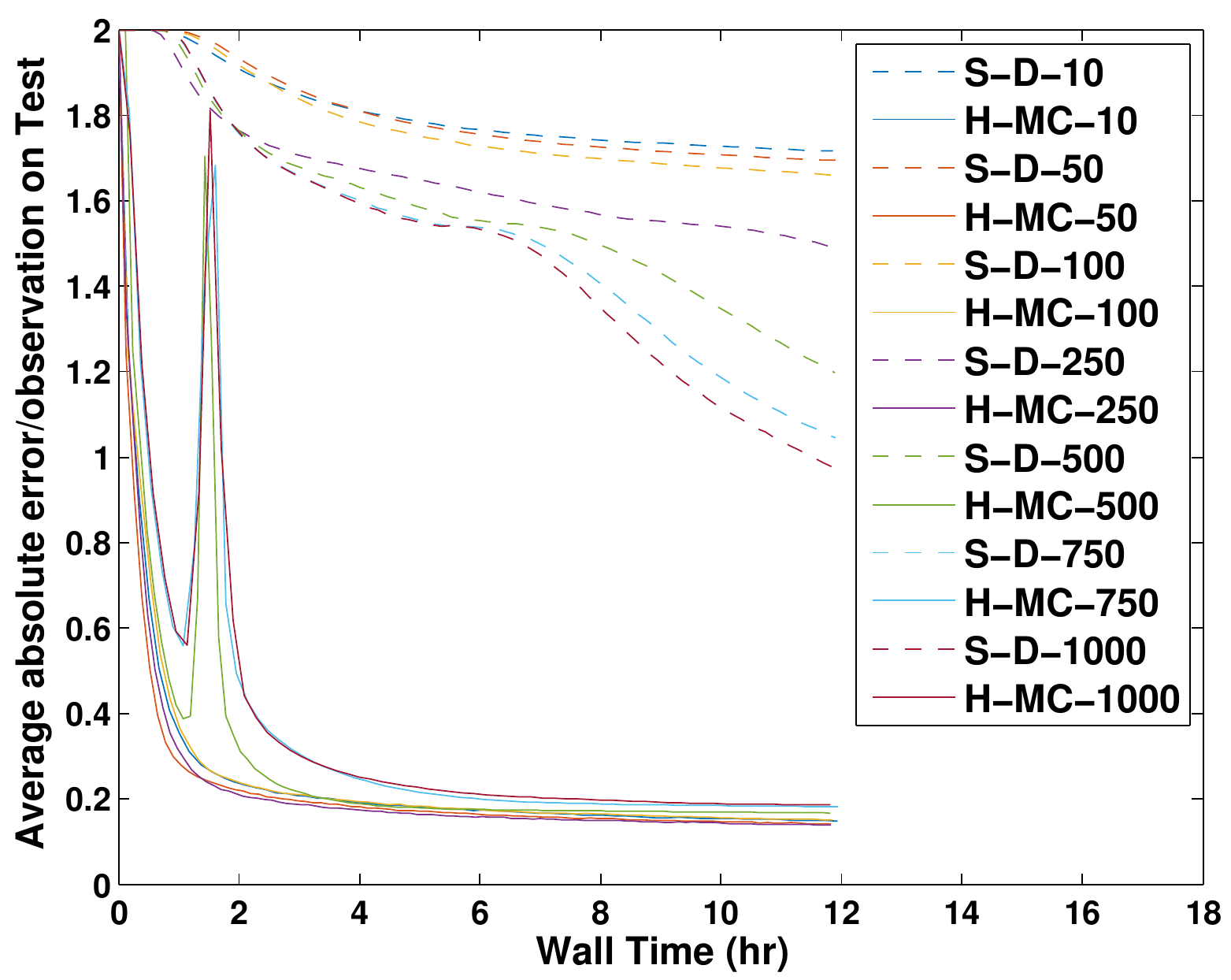}
\end{center}
\end{minipage}
\begin{minipage}{\mpw}
\begin{center}
\includegraphics[width=0.99\linewidth]{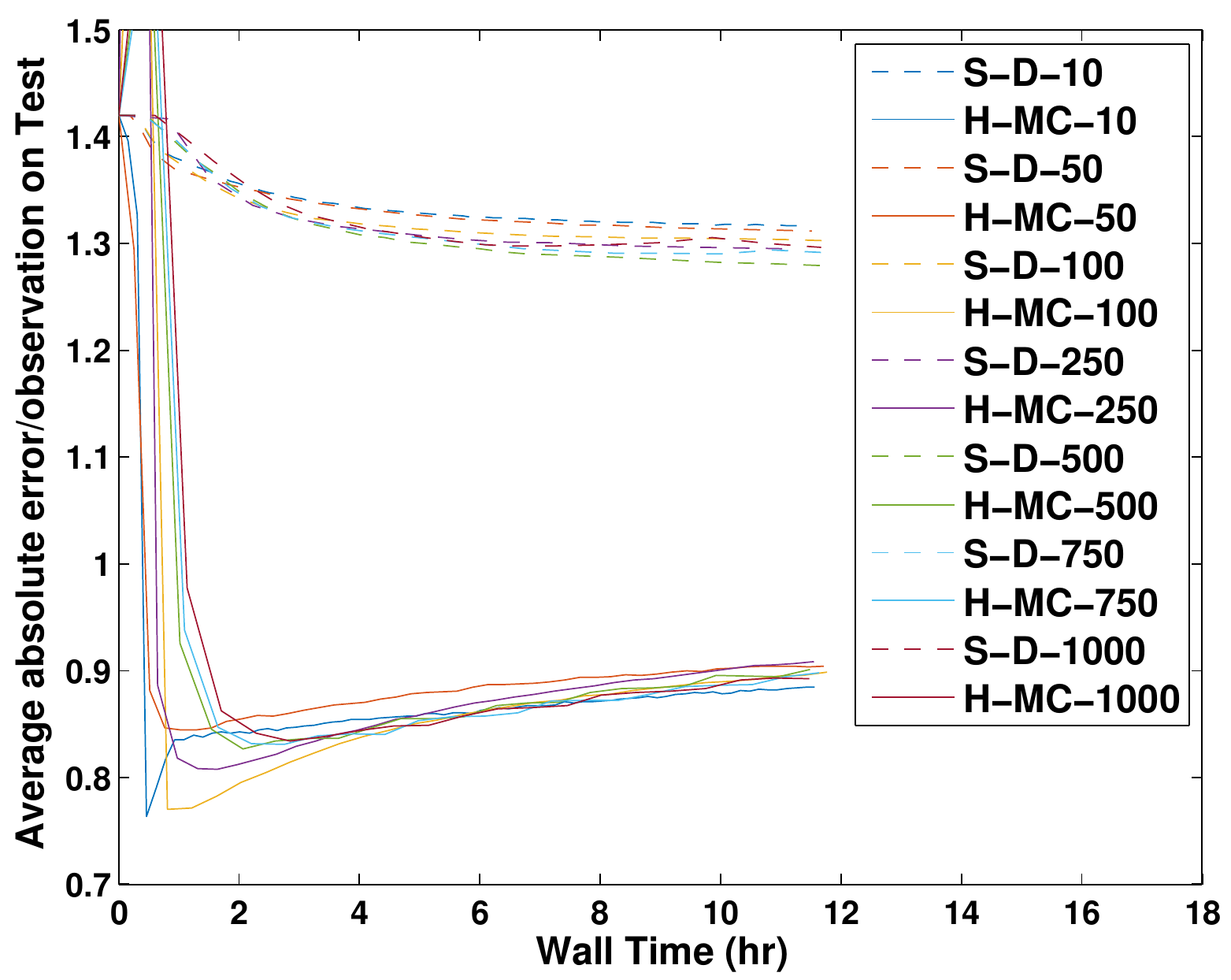}
\end{center}
\end{minipage}

\caption{#2}
\label{#1}
\end{figure*}
}

\newcommand{\putPMFPlotsNLLTrials}{
\begin{figure*}[t]

\begin{minipage}{\mpw}
\begin{center}
\includegraphics[width=0.99\linewidth]{artificial_binary_L1e6_v2_traintest_D100_nlltrials}
\end{center}
\end{minipage}
\begin{minipage}{\mpw}
\begin{center}
\includegraphics[width=0.99\linewidth]{ml1M_bin_traintest_D100_nlltrials}
\end{center}
\end{minipage}

\vspace{2mm}

\begin{minipage}{\mpw}
\begin{center}
\includegraphics[width=0.99\linewidth]{artificial_count_L1e6_v2_traintest_D100_nlltrials}
\end{center}
\end{minipage}
\begin{minipage}{\mpw}
\begin{center}
\includegraphics[width=0.99\linewidth]{lastfm_count_traintest_D100_nlltrials}
\end{center}
\end{minipage}

\vspace{2mm}

\begin{minipage}{\mpw}
\begin{center}
\includegraphics[width=0.99\linewidth]{artificial_ordinal_L1e6_v2_traintest_D100_nlltrials}
\end{center}
\end{minipage}
\begin{minipage}{\mpw}
\begin{center}
\includegraphics[width=0.99\linewidth]{ml1M_ord_traintest_D100_nlltrials}
\end{center}
\end{minipage}

\caption{
Experimental results across several likelihood functions (from top to bottom: binary, count, ordinal-5) and artificial (left) and real (right) data sets showing average NLL per observation on test with respect to training time. Lower values indicate better performance.
}
\label{fig:pmfplotsnlltrials}
\end{figure*}
}

\newcommand{\putPMFPlotsNLL}[2]{
\begin{figure*}[p]

\begin{minipage}{\mpw}
\begin{center}
\includegraphics[width=0.99\linewidth]{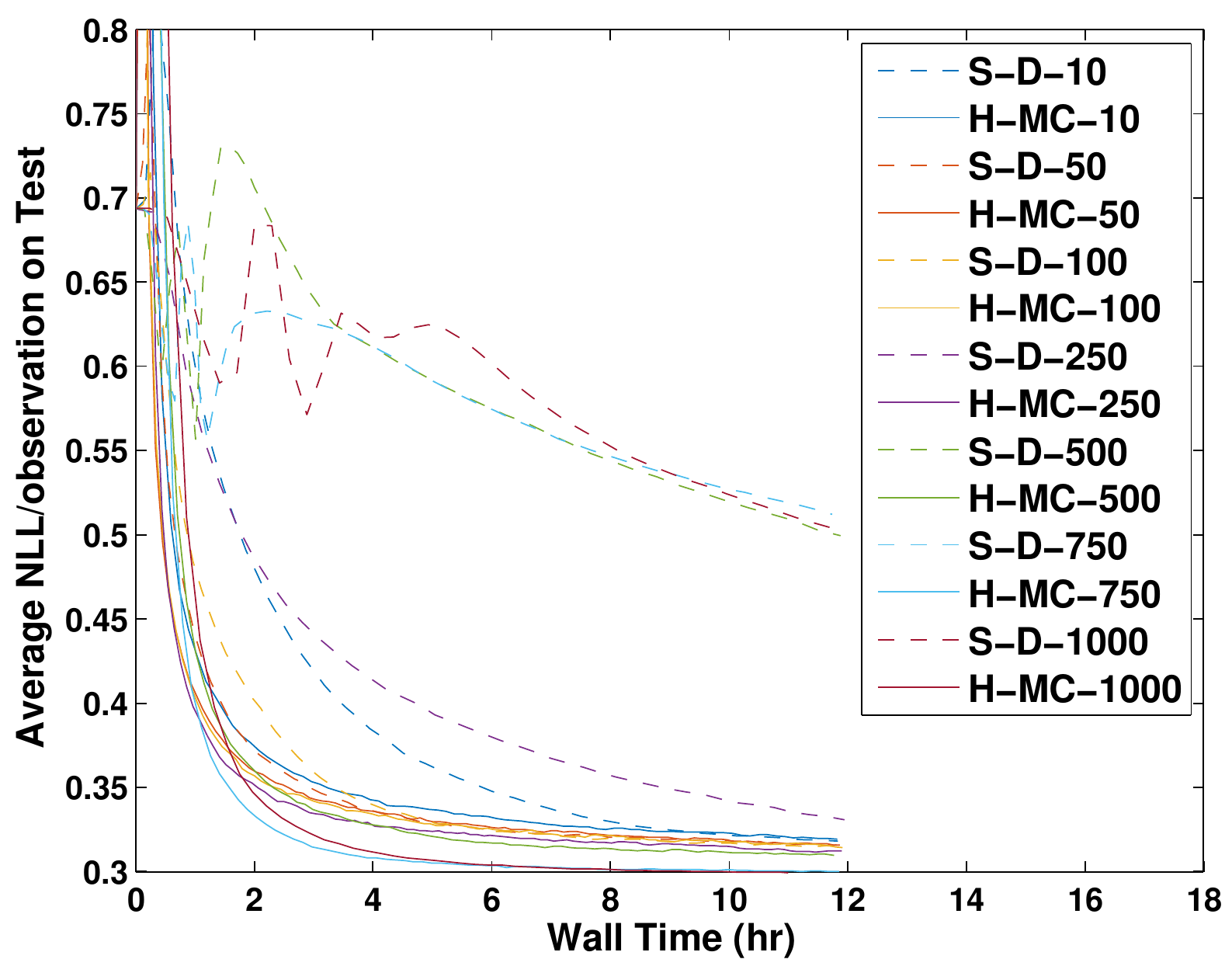}
\end{center}
\end{minipage}
\begin{minipage}{\mpw}
\begin{center}
\includegraphics[width=0.99\linewidth]{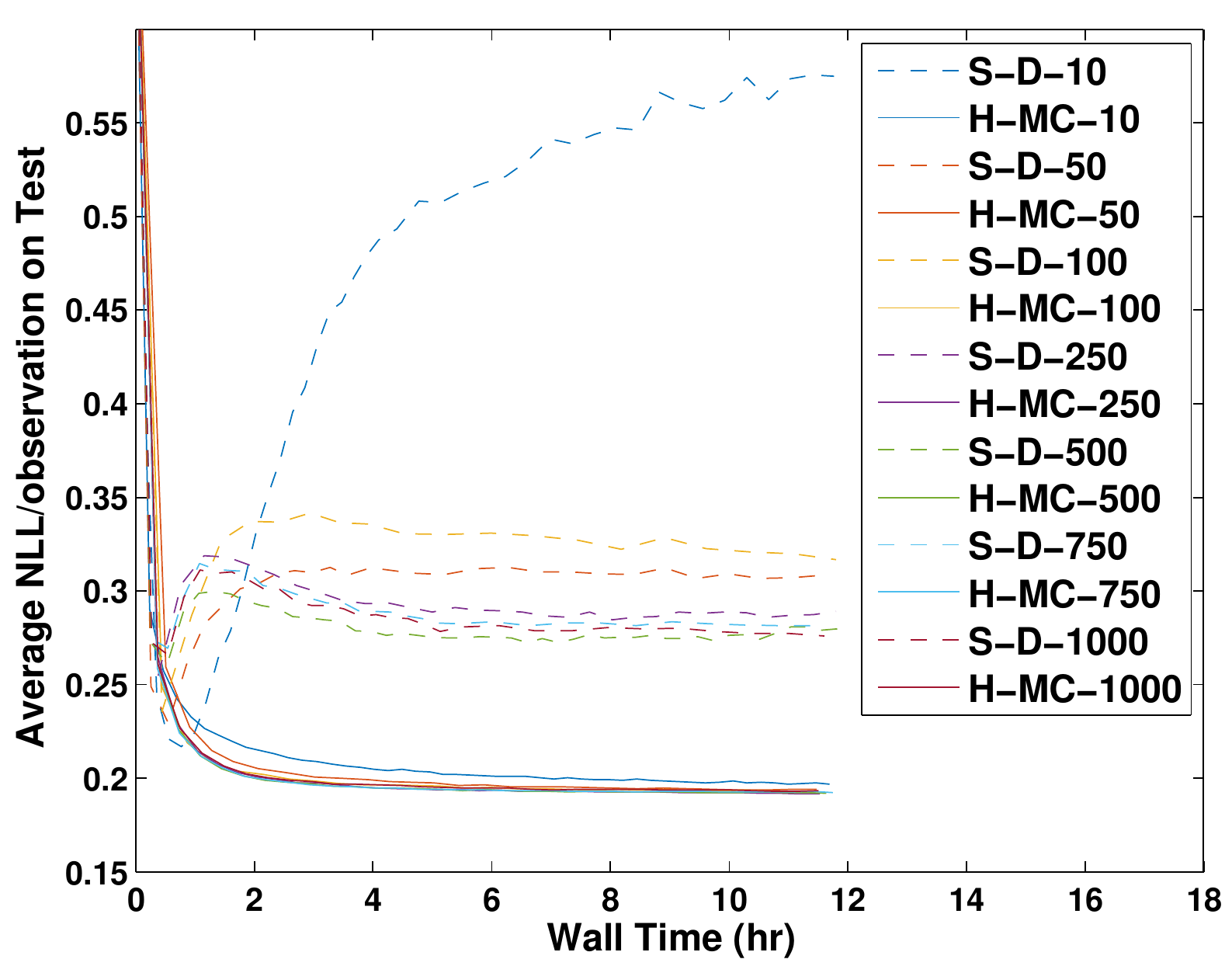}
\end{center}
\end{minipage}

\vspace{2mm}

\begin{minipage}{\mpw}
\begin{center}
\includegraphics[width=0.99\linewidth]{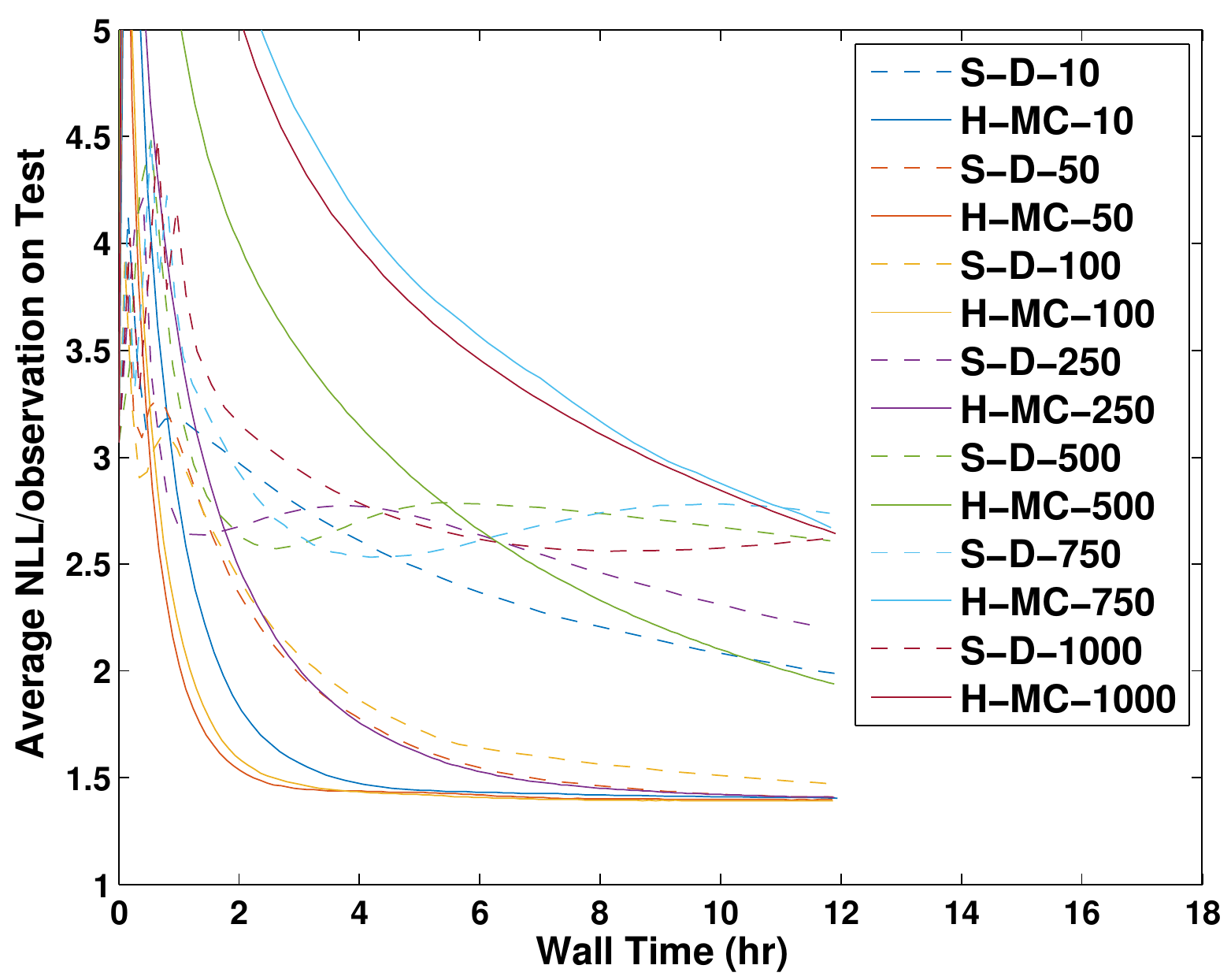}
\end{center}
\end{minipage}
\begin{minipage}{\mpw}
\begin{center}
\includegraphics[width=0.99\linewidth]{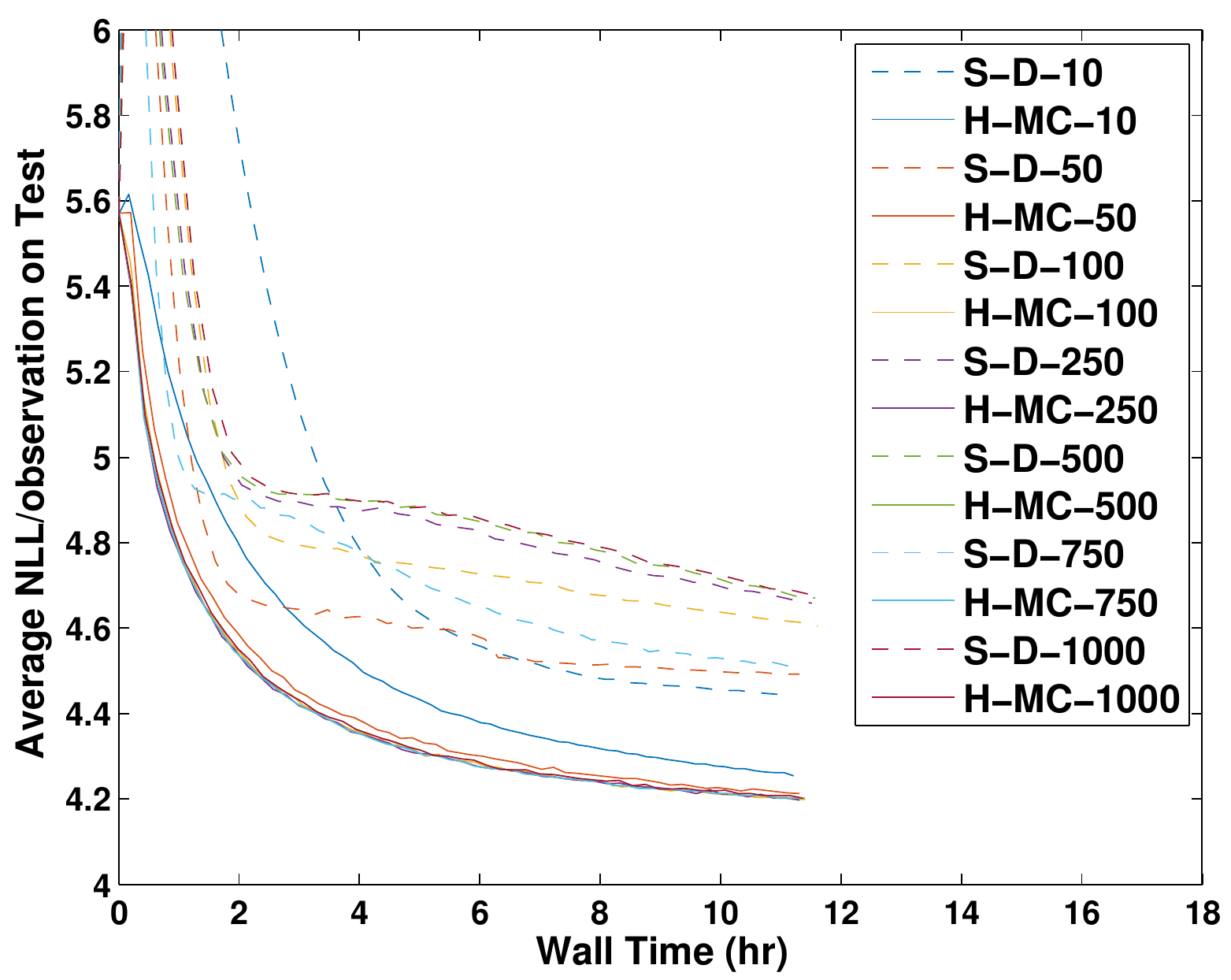}
\end{center}
\end{minipage}

\vspace{2mm}

\begin{minipage}{\mpw}
\begin{center}
\includegraphics[width=0.99\linewidth]{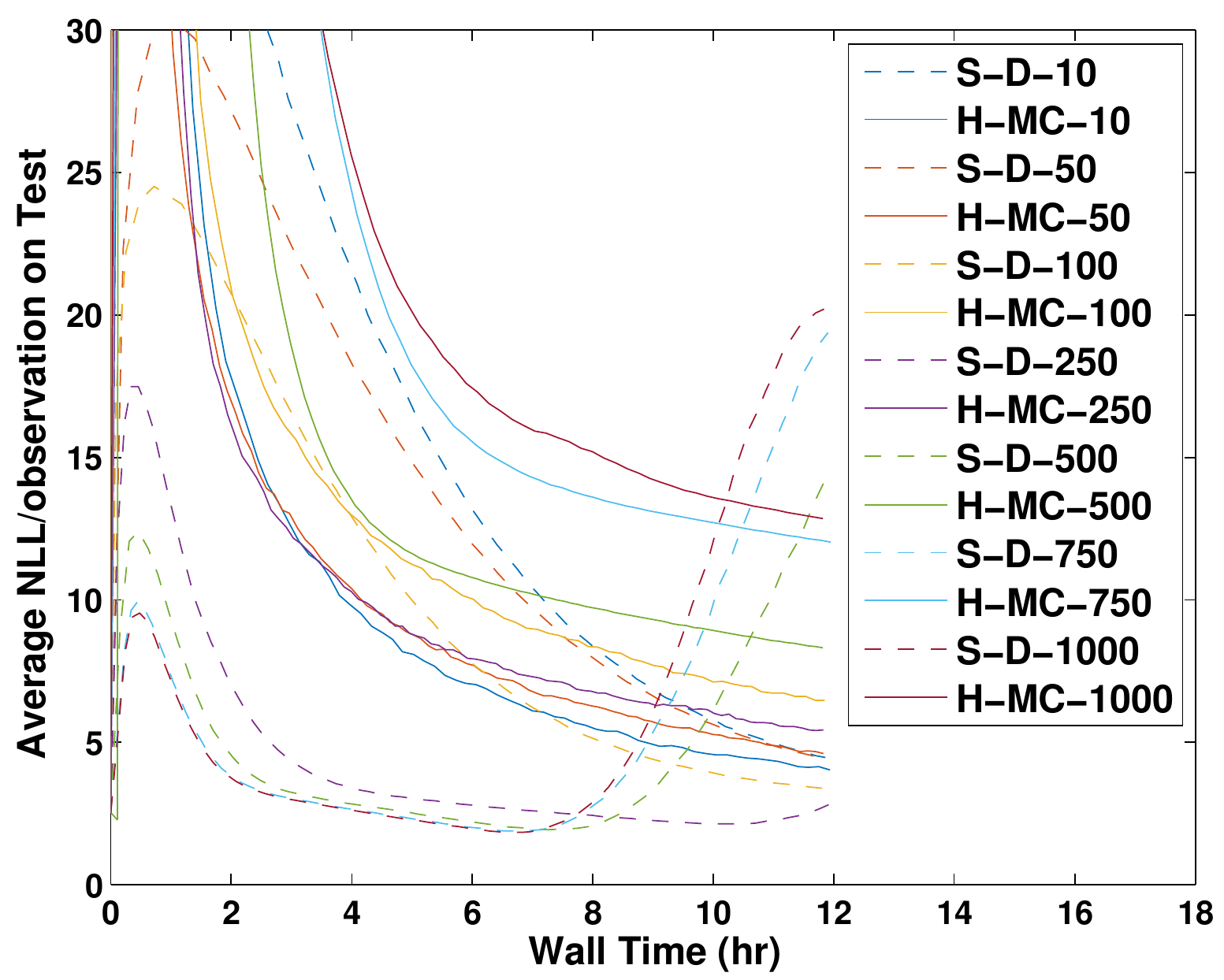}
\end{center}
\end{minipage}
\begin{minipage}{\mpw}
\begin{center}
\includegraphics[width=0.99\linewidth]{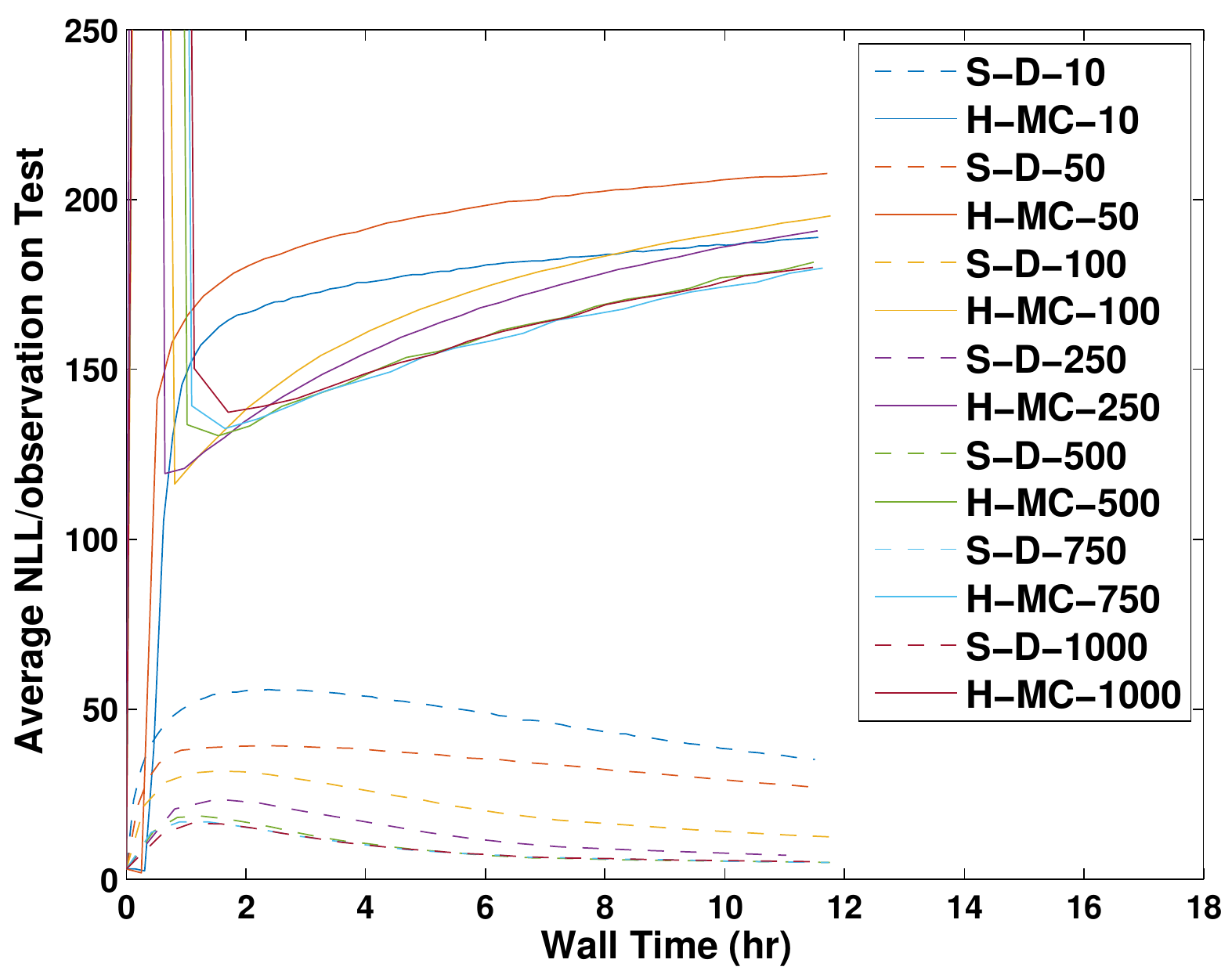}
\end{center}
\end{minipage}

\caption{#2}
\label{#1}
\end{figure*}
}

\newcommand{\putPMFPlotsVLBZoom}{
\begin{figure*}[h]

\begin{minipage}{\mpw}
\begin{center}
\includegraphics[width=0.9\linewidth]{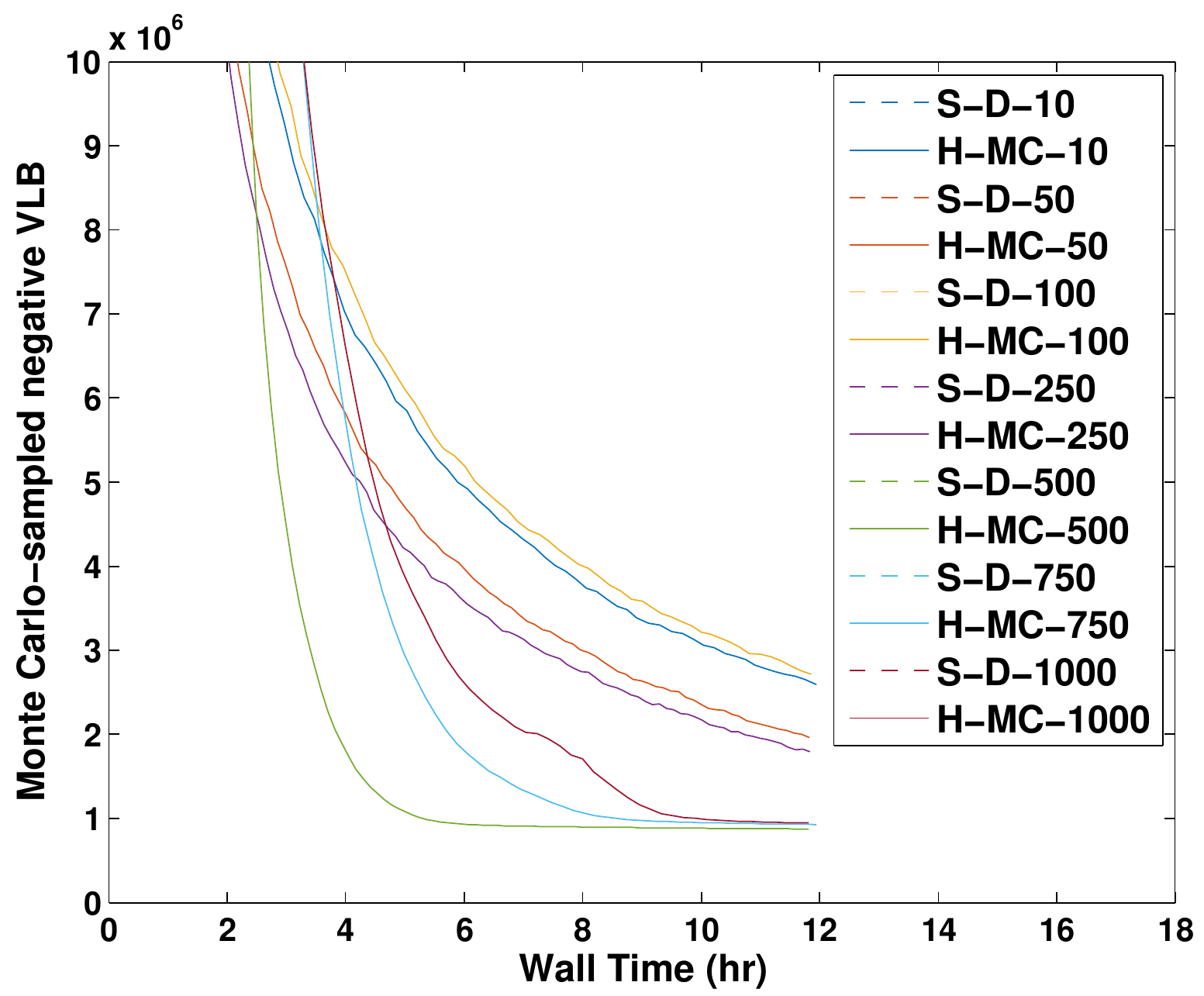}
\end{center}
\end{minipage}
\begin{minipage}{\mpw}
\begin{center}
\includegraphics[width=0.9\linewidth]{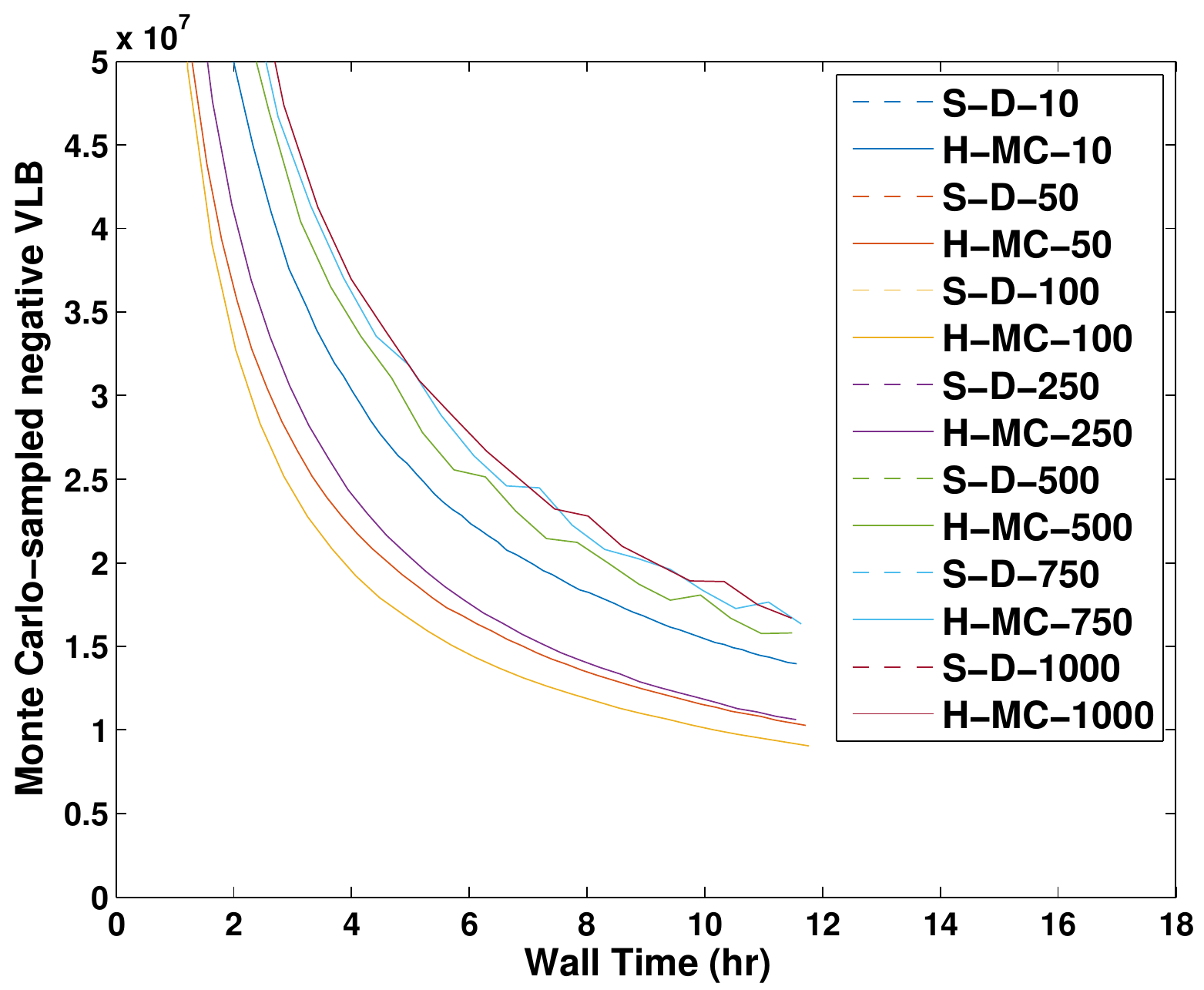}
\end{center}
\end{minipage}

\caption{
Zoom-in of bottom row of Figure \ref{fig:pmfplotsvlb}. Note, at this scale, \dsvi{} performance is not visible.
}
\label{fig:pmfplotsvlbzoom}
\end{figure*}
}

\newcommand{\putPMFPlotsVLB}[2]{
\begin{figure*}[p]

\begin{minipage}{\mpw}
\begin{center}
\includegraphics[width=0.98\linewidth]{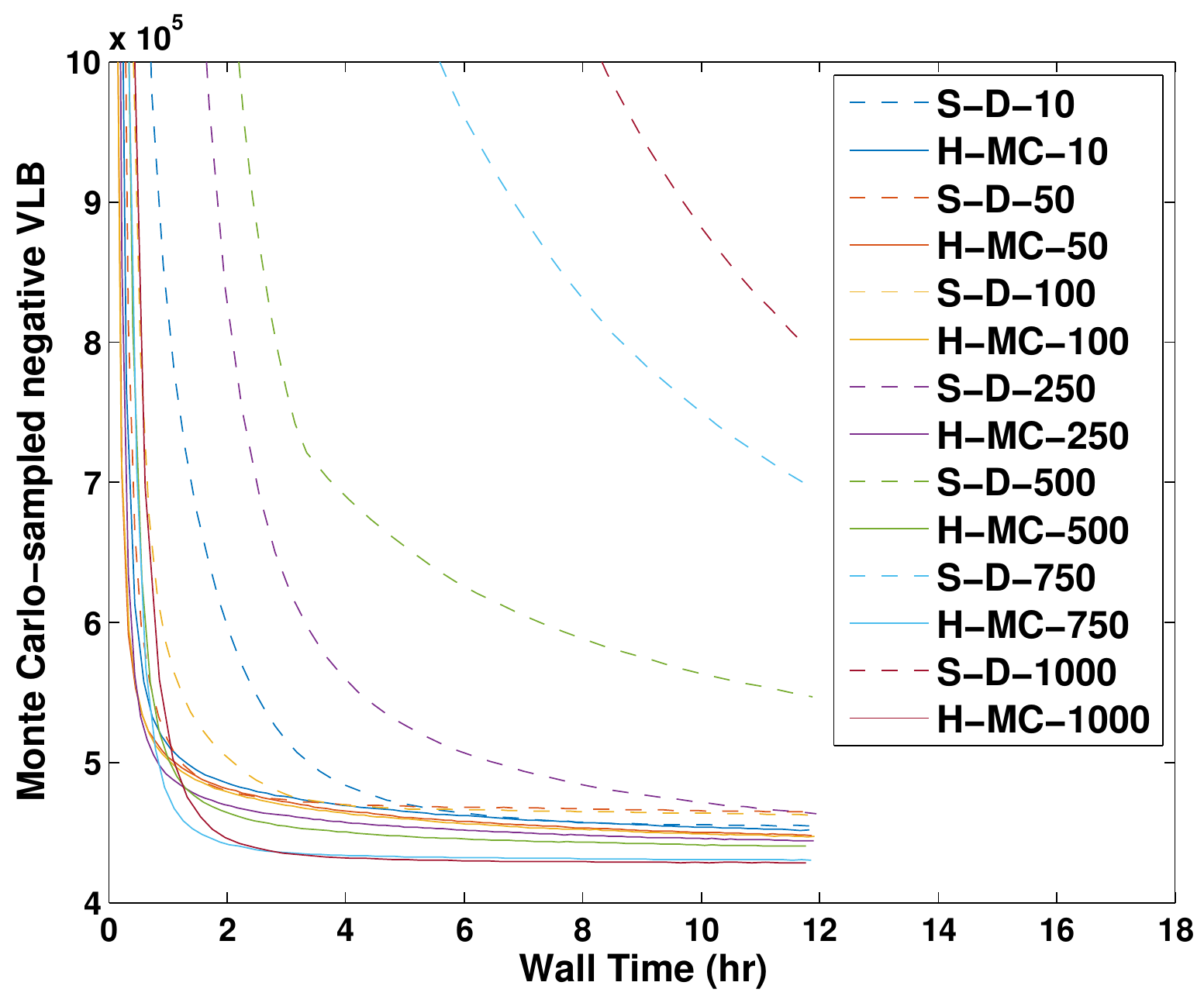}
\end{center}
\end{minipage}
\begin{minipage}{\mpw}
\begin{center}
\includegraphics[width=0.98\linewidth]{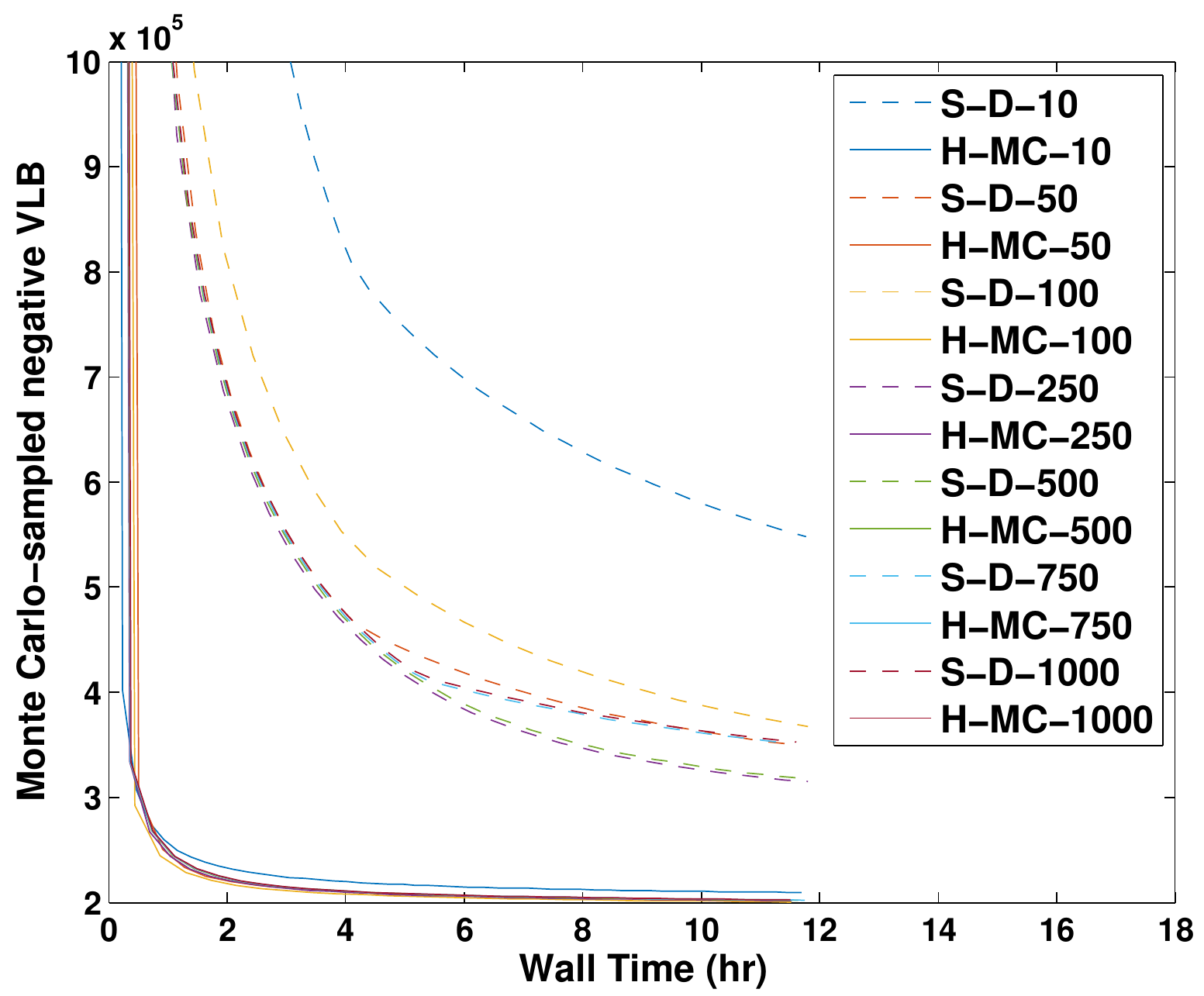}
\end{center}
\end{minipage}

\vspace{2mm}

\begin{minipage}{\mpw}
\begin{center}
\includegraphics[width=0.98\linewidth]{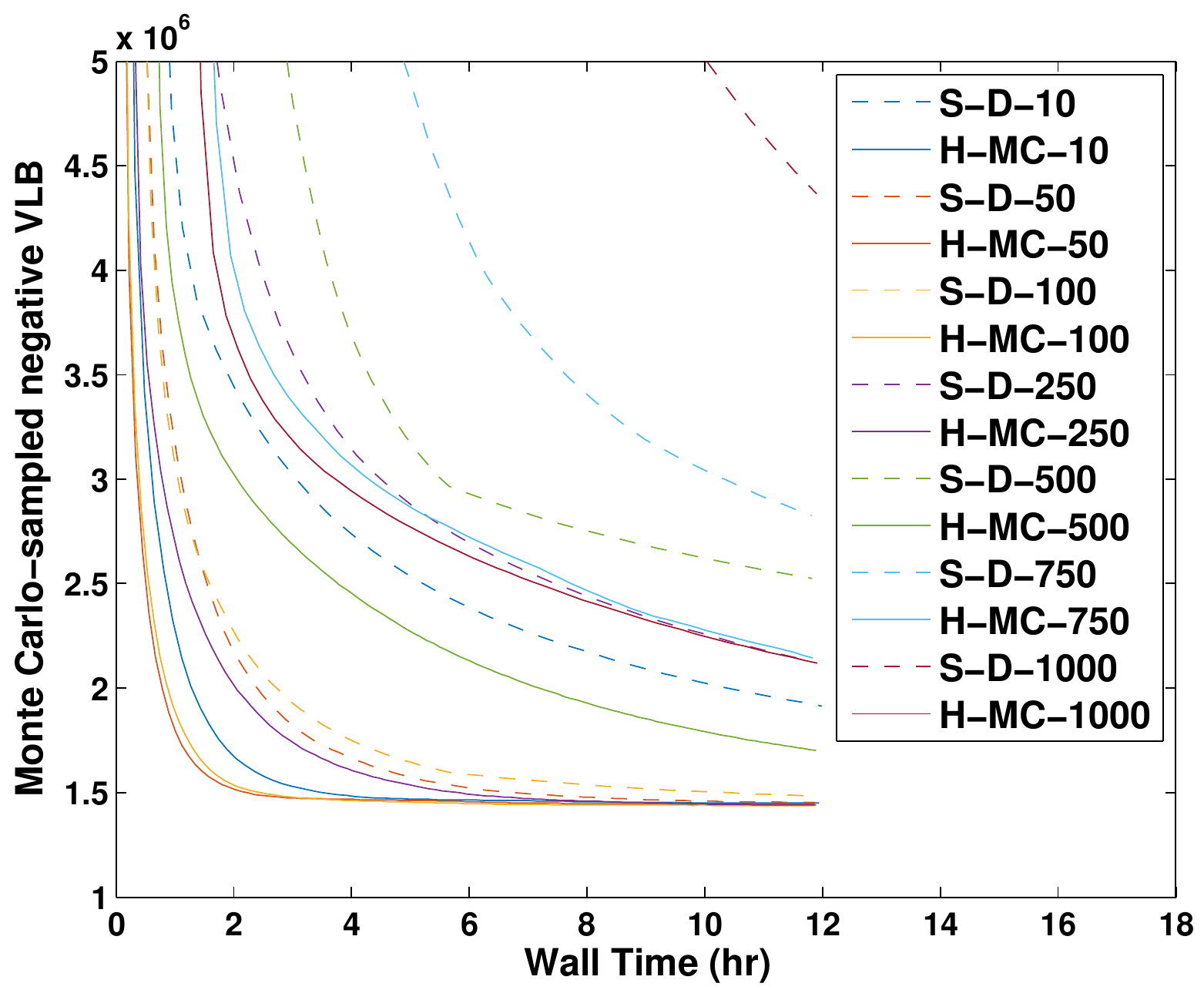}
\end{center}
\end{minipage}
\begin{minipage}{\mpw}
\begin{center}
\includegraphics[width=0.98\linewidth]{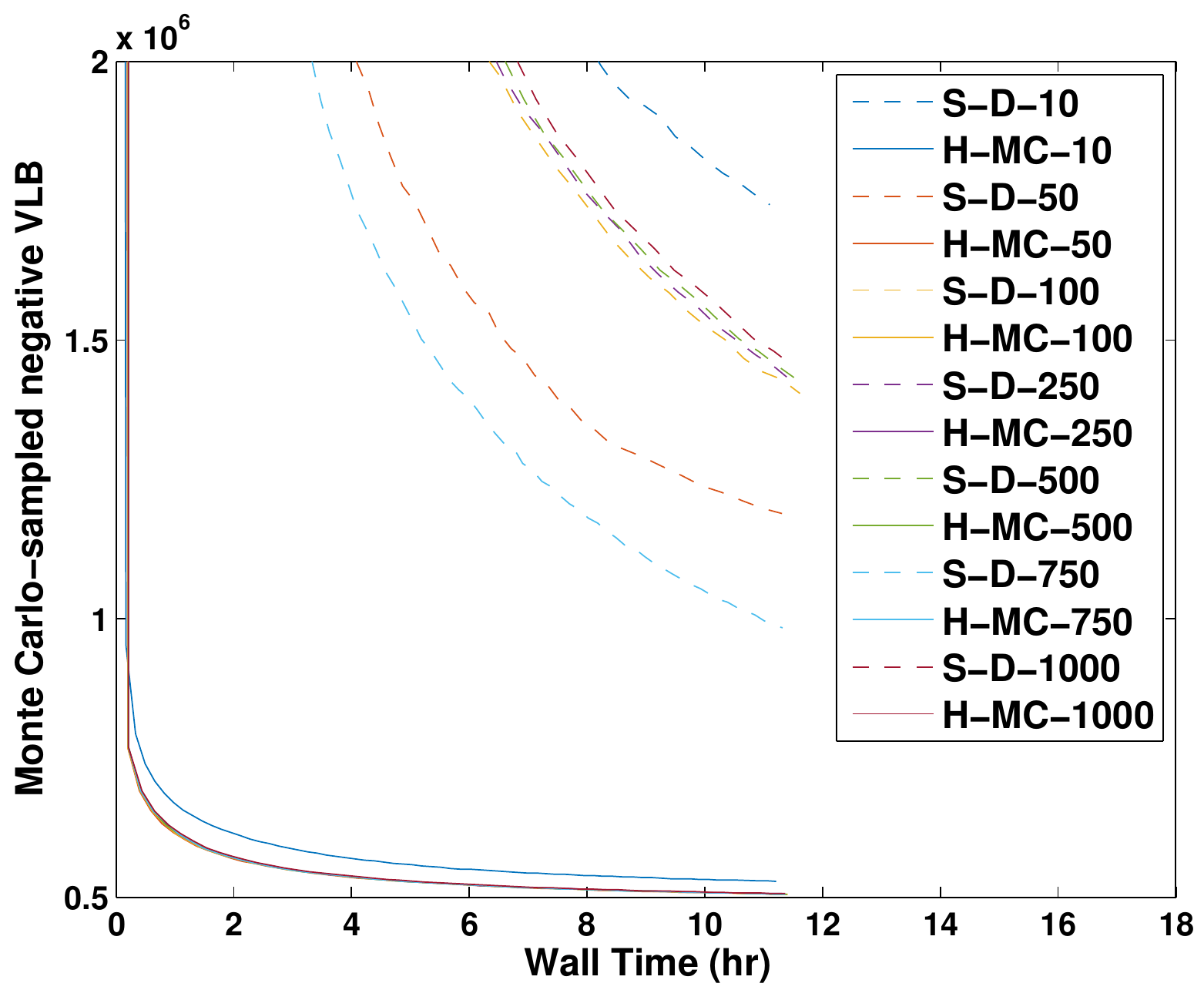}
\end{center}
\end{minipage}

\vspace{2mm}

\begin{minipage}{\mpw}
\begin{center}
\includegraphics[width=0.98\linewidth]{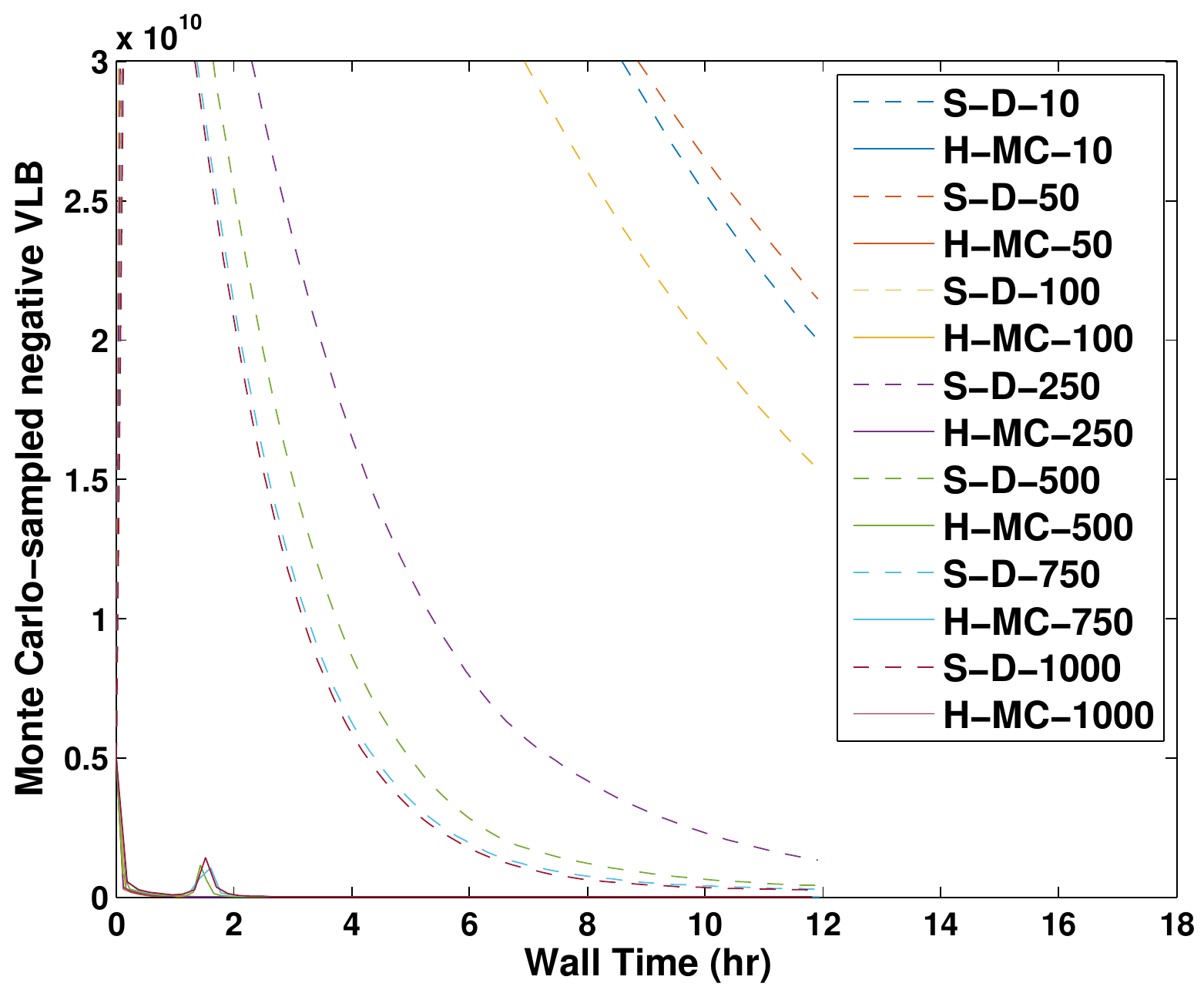}
\end{center}
\end{minipage}
\begin{minipage}{\mpw}
\begin{center}
\includegraphics[width=0.98\linewidth]{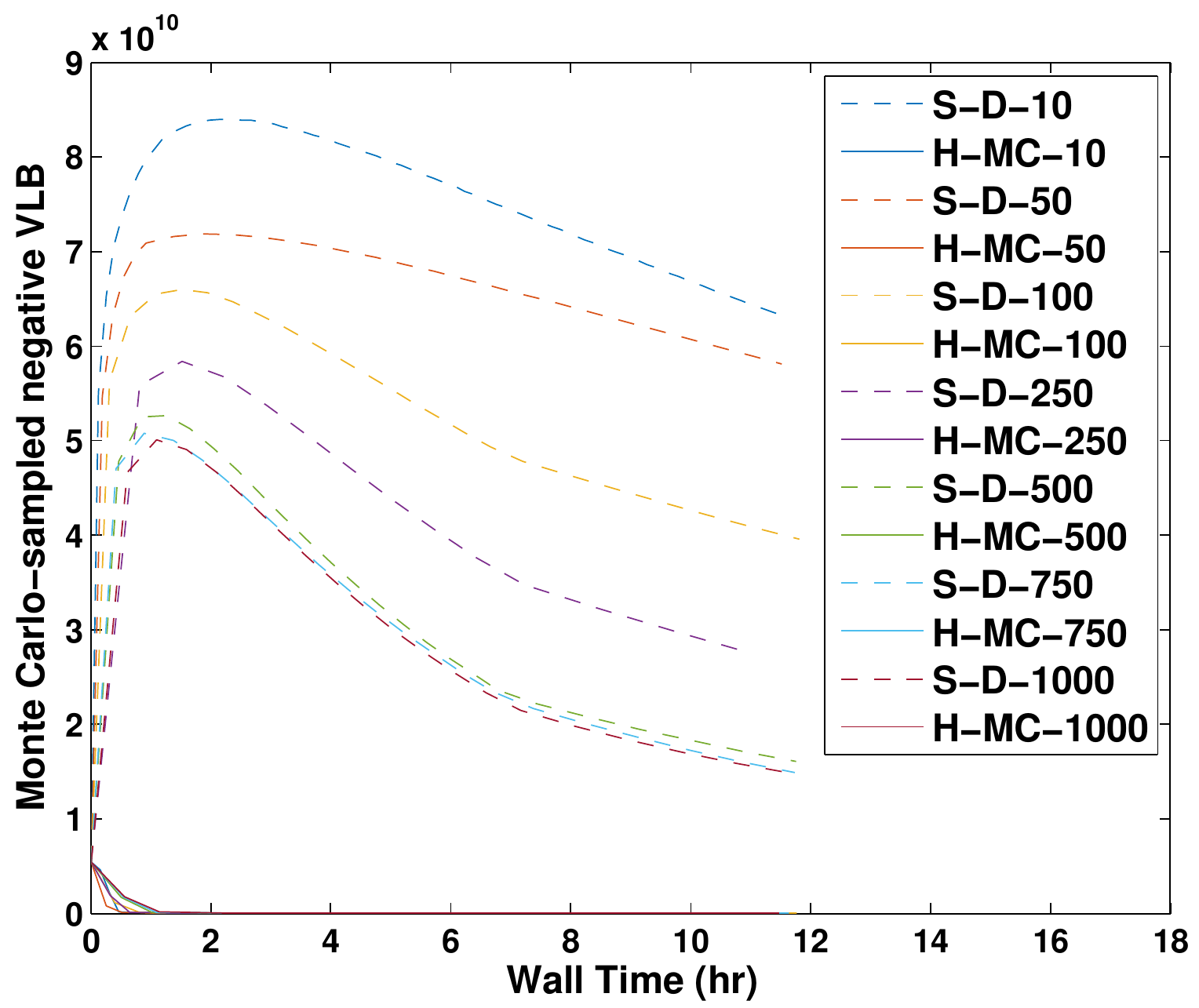}
\end{center}
\end{minipage}

\caption{#2}
\label{#1}
\end{figure*}
}

\newcommand{\putPMFPlotsOld}{
\begin{figure*}[t]

\begin{minipage}{\mpw}
\begin{center}
\includegraphics[width=0.99\linewidth]{art_bin_D100-grad-sngd}
\end{center}
\end{minipage}
\begin{minipage}{\mpw}
\begin{center}
\includegraphics[width=0.99\linewidth]{ml1M_bin_D100-grad-sngd}
\end{center}
\end{minipage}

\vspace{2mm}

\begin{minipage}{\mpw}
\begin{center}
\includegraphics[width=0.99\linewidth]{art_count_D100-grad-sngd}
\end{center}
\end{minipage}
\begin{minipage}{\mpw}
\begin{center}
\includegraphics[width=0.99\linewidth]{lastfm_count_D100-grad-sngd}
\end{center}
\end{minipage}

\vspace{2mm}

\begin{minipage}{\mpw}
\begin{center}
\includegraphics[width=0.99\linewidth]{art_ord_D100-grad-sngd}
\end{center}
\end{minipage}
\begin{minipage}{\mpw}
\begin{center}
\includegraphics[width=0.99\linewidth]{ml1M_ord_D100-grad-sngd}
\end{center}
\end{minipage}

\caption{
Experimental results across several models and likelihood functions, showing objective function values with respect to training time. The y-axis is Monte Carlo-sampled negative VLB.
}
\label{fig:pmfplots}
\end{figure*}
}

\newcommand{\putZoomedPMFPlotsOld}{
\begin{figure*}[t]

\begin{minipage}{0.5\linewidth}
\begin{center}
\includegraphics[width=0.99\linewidth]{ml1M_bin_D100-grad-sngd_zoom}
\end{center}
\end{minipage}
\begin{minipage}{0.5\linewidth}
\begin{center}
\includegraphics[width=0.99\linewidth]{lastfm_count_D100-grad-sngd_zoom}
\end{center}
\end{minipage}

\vspace{2mm}

\begin{minipage}{0.5\linewidth}
\begin{center}
\includegraphics[width=0.99\linewidth]{art_ord_D100-grad-sngd_zoom}
\end{center}
\end{minipage}
\begin{minipage}{0.5\linewidth}
\begin{center}
\includegraphics[width=0.99\linewidth]{ml1M_ord_D100-grad-sngd_zoom}
\end{center}
\end{minipage}

\caption{
Zoomed experimental results across several models and likelihood functions, showing objective function values with respect to training time. The y-axis is Monte Carlo-sampled negative VLB.
}
\label{fig:zoomedpmfplots}
\end{figure*}
}

\newcommand{\putGLMPlot}[2]{
\begin{figure*}[t]

\begin{center}
\includegraphics[width=0.5\linewidth]{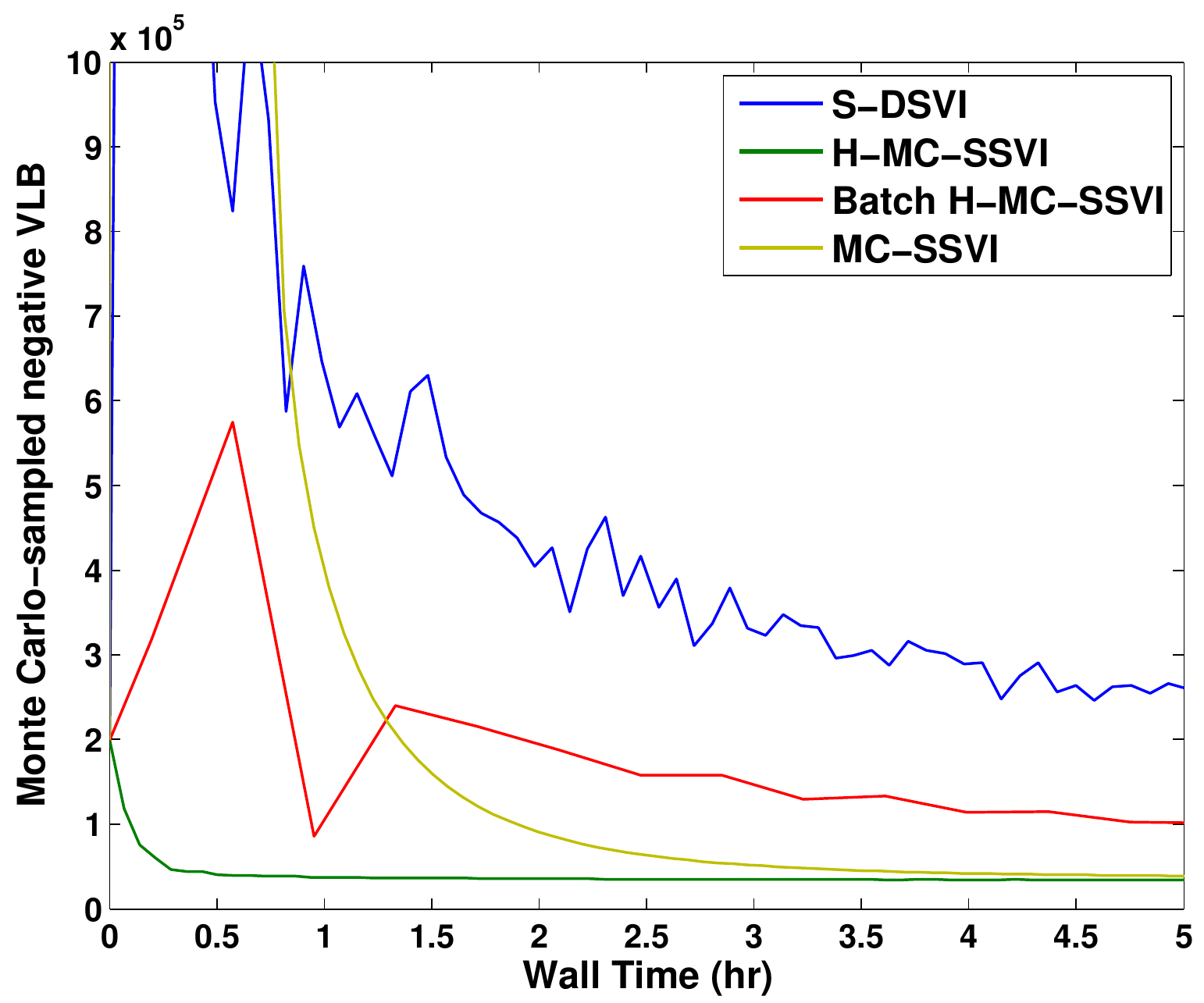}
\end{center}

\caption{#2}
\label{#1}
\end{figure*}
}

\newcommand{\putPMFPlotsA}[2]{
\begin{figure*}[p]

\begin{minipage}{0.5\linewidth}
\begin{center}
\includegraphics[width=0.99\linewidth]{art_bin_mcsvi}
\end{center}
\end{minipage}
\begin{minipage}{0.5\linewidth}
\begin{center}
\includegraphics[width=0.99\linewidth]{ml1m_ord_mcsvi}
\end{center}
\end{minipage}

\caption{#2}
\label{#1}
\end{figure*}
}

\newcommand{\putPMFPlotsB}[2]{
\begin{figure*}[p]

\begin{minipage}{0.5\linewidth}
\begin{center}
\includegraphics[width=0.99\linewidth]{art_bin_mcsvi_dsvi}
\end{center}
\end{minipage}
\begin{minipage}{0.5\linewidth}
\begin{center}
\includegraphics[width=0.99\linewidth]{art_count_mcsvi_dsvi}
\end{center}
\end{minipage}

\caption{#2}
\label{#1}
\end{figure*}
}

\newcommand{\putCTMStructurePlot}[2]{
\begin{figure*}[p]

\begin{minipage}{0.5\linewidth}
\begin{center}
\includegraphics[width=0.99\linewidth]{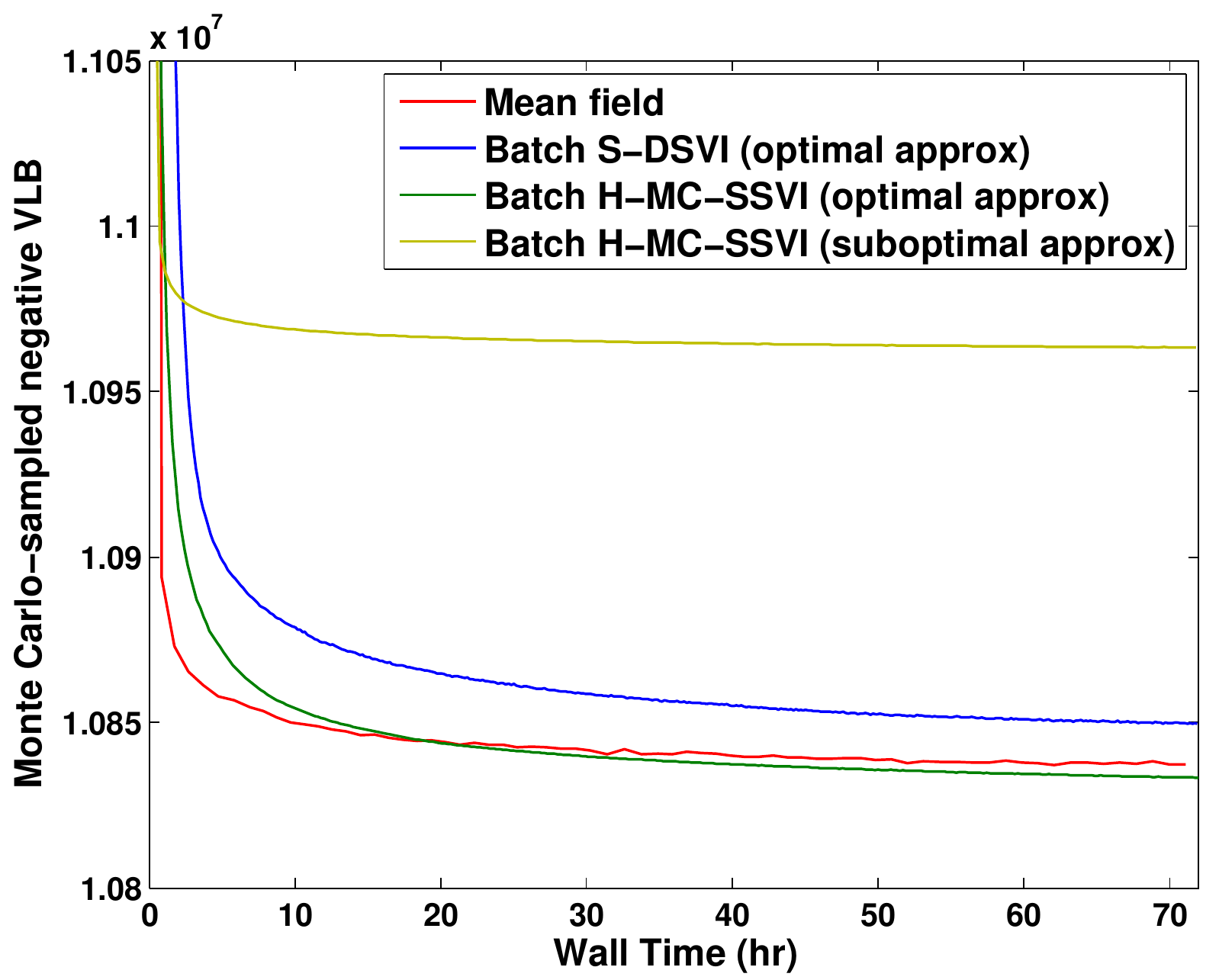}
\end{center}
\end{minipage}
\begin{minipage}{0.5\linewidth}
\begin{center}
\includegraphics[width=0.99\linewidth]{nips_v3_traintest_K50_nll_A2}
\end{center}
\end{minipage}

\caption{#2}
\label{#1}
\end{figure*}
}

\newcommand{\putCTMVLBStructurePlot}[2]{
\begin{figure*}[p]

\begin{center}
\includegraphics[width=0.65\linewidth]{nips_v3_traintest_K50_vlb_diag_batch}
\end{center}

\caption{#2}
\label{#1}
\end{figure*}
}

\newcommand{\putCTMVLBZoomPlot}[2]{
\begin{figure*}[p]

\begin{center}
\includegraphics[width=0.65\linewidth]{nips_v3_traintest_K50_vlb_zoom}
\end{center}

\caption{#2}
\label{#1}
\end{figure*}
}

\newcommand{\putCTMNLLStructurePlot}[2]{
\begin{figure*}[p]

\begin{center}
\includegraphics[width=0.65\linewidth]{nips_v3_traintest_K50_nll_diag_batch}
\end{center}

\caption{#2}
\label{#1}
\end{figure*}
}

\newcommand{\putCTMVLBPlotEnron}[2]{
\begin{figure*}[p]

\begin{center}
\includegraphics[width=0.5\linewidth]{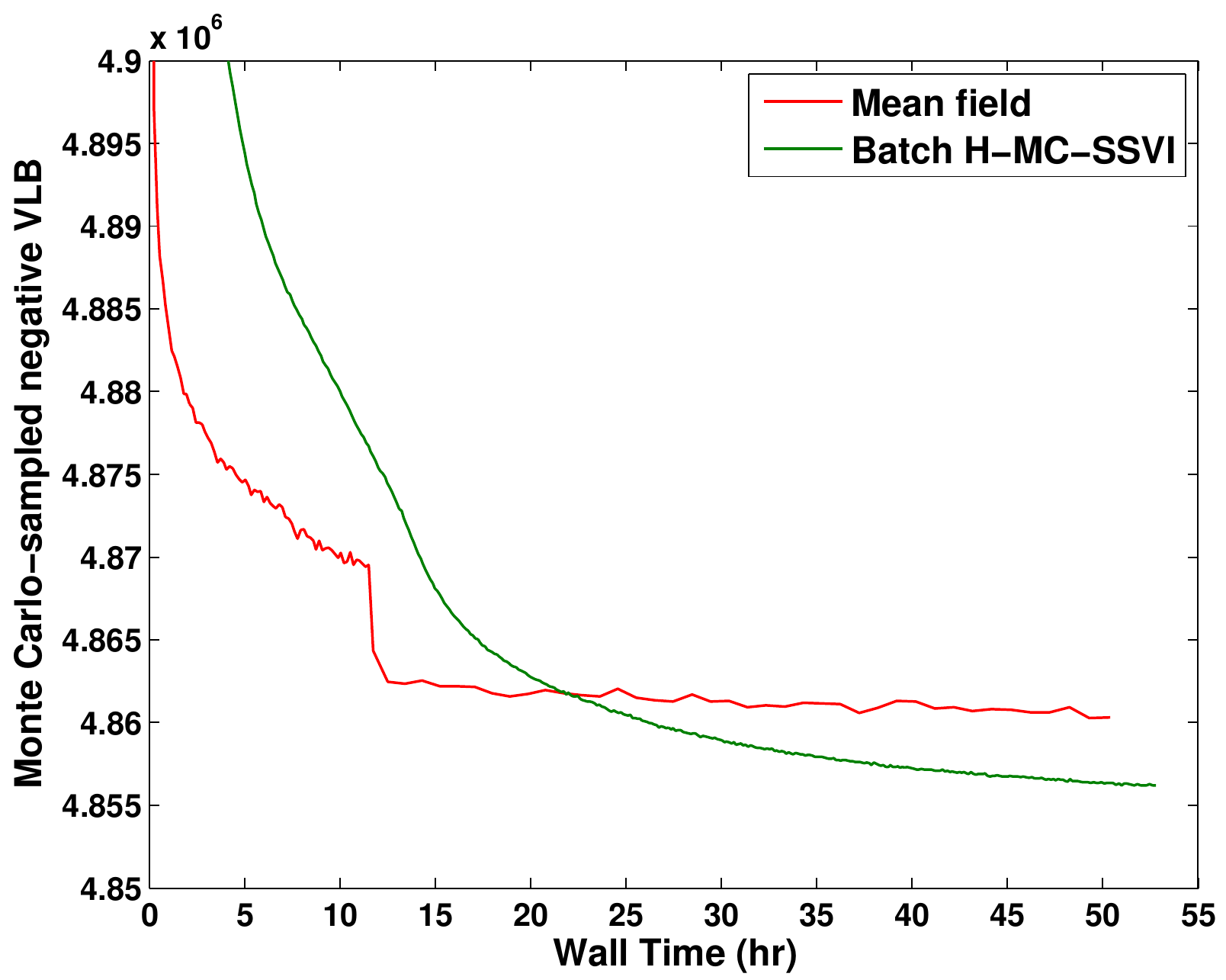}
\end{center}

\caption{#2}
\label{#1}
\end{figure*}
}

\newcommand{\putCTMKPlot}[2]{
\begin{figure*}[p]

\begin{center}
\includegraphics[width=0.5\linewidth]{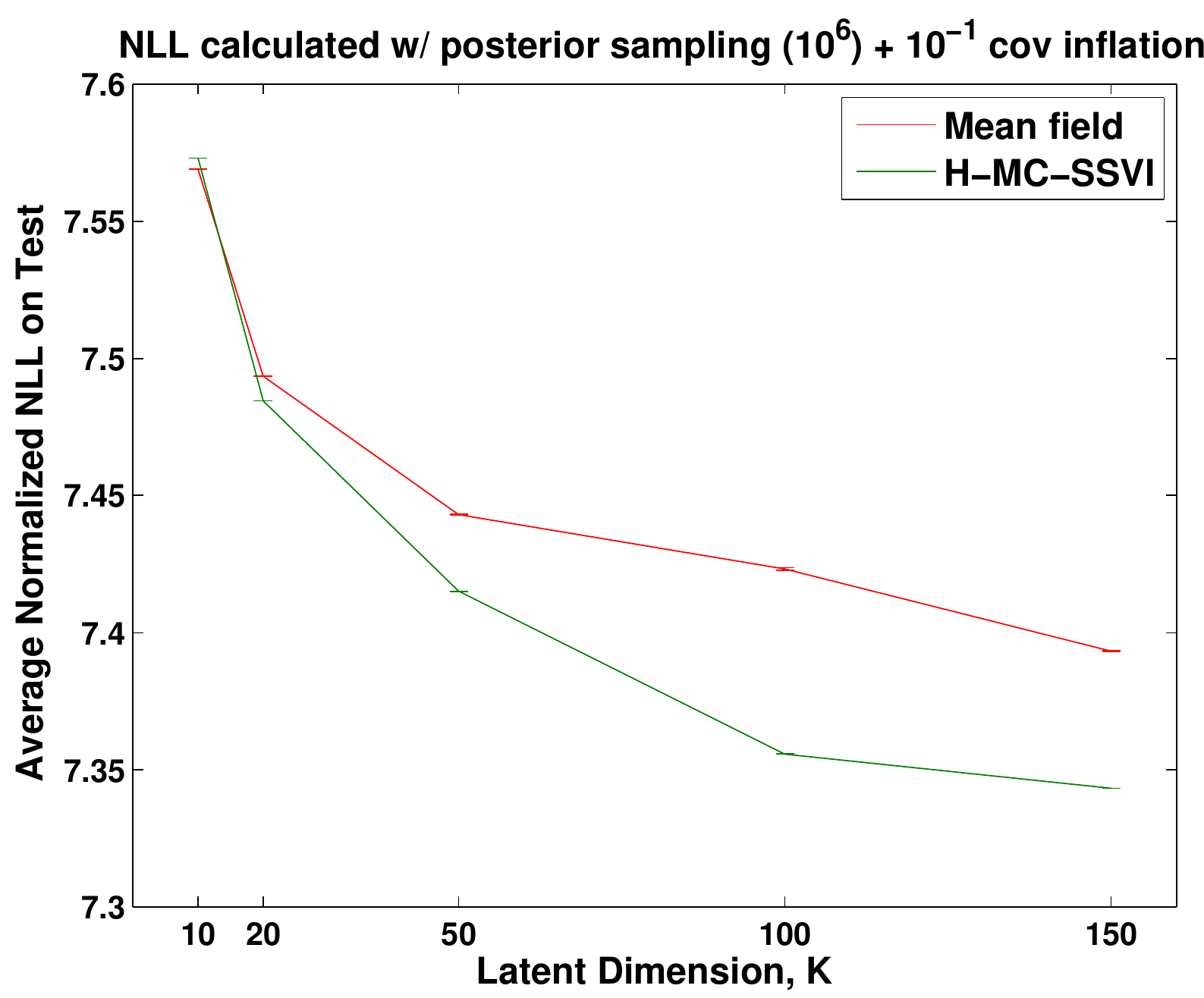}
\end{center}

\caption{#2}
\label{#1}
\end{figure*}
}

\section{Introduction}

Work over the last two decades developed powerful models in probabilistic machine leaning capturing, for example,  structure in text documents, collaborative filtering applications and more. With complex probabilistic models, exact inference is rarely possible and much research has been devoted to develop successful algorithms for approximate inference in such models.
Variational approximations are one of the main paradigms for approximate efficient inference, but optimization of the variational approximation is still challenging. 
Multiple authors have recently introduced the idea of combining stochastic gradient ascent with variational approximations and, although they differ in details, several of these are known as instances of SVI (stochastic variational inference).  Two sources of stochastic estimates of gradients have been proposed. The first is 
using mini-batch samples from the data which avoids having to process the entire dataset for each gradient update. The second is using Monte-Carlo estimates of intractable expectations in complex models which otherwise prevent application of the variational paradigm.  This line of work has been very successful and it spans work in Gaussian processes \cite{Hensman2013}, topic models \cite{Hoffman2013}, matrix factorization \cite{Hernandez-LobatoHG14}, deep learning \cite{Kingma2014} and more. The type of variational approximations used in this work has varied from the simplest mean field approximation where all variables are assumed independent to more structured variants. 

The work of \cite{Hoffman2013} incorporates an additional idea which is applicable to models with  conjugate complete conditionals in the exponential family. In this case, the use of natural gradients \cite{Amari2000} on the natural parameters of the distribution,
combined with the mean-field approximation,
leads to especially simple update equations. Combining the advantages of natural gradients with stochastic updates, and the fact that such updates are simple computationally, leads to superior performance in practice. 
The work of \cite{Hoffman2015} extends this algorithm by using a better variational approximation, a structured approximation where a portion of the variational distribution is selected optimally. The improved approximation leads to significant improvement in predictive performance \cite{Hoffman2015}. We refer to this approach which includes a structured approximation, mini-batch data sampling, Monte-Carlo sampling, and natural gradients as \algfull \ (\alg).

The paper makes several contributions extending this line work. 
To obtain this generalization we introduce
a family of Bayesian latent variable models with two levels of hidden variables (\lvm{}), without any conjugacy requirements.
Many existing models in the literature belong to this family as special cases: generalized linear models (GLM), Mixed Effects GLM (GME), Gaussian processes (GP), sparse GPs, GP latent variable model (GPLVM), probabilistic matrix factorization (PMF), latent Dirichlet allocation (LDA), correlated topic models (CTM) and more, and all of these have been solved with variational approximations.

We start by 
reviewing and unifying the treatment from previous work by showing that the structured variational approximation \cite{Hoffman2015,LimTeh2007} is applicable
to the \lvm{}, 
and identifying when previous approximations in the literature are optimal in this sense.
This analysis also clarifies the the relations between the different variational approximations that have been used in different models, and suggests potential for improvement. 

Our main contribution is to show that \alg{} is applicable in the \lvm. 
Our analysis uses recent observations on fixed-point updates and their relation to natural gradients \cite{ShethKh2016} to show that \alg{}  can be applied whenever the prior and variational distributions over latent variables have the same form in the exponential family. 
The algorithm enjoys the same benefits listed above while expanding the scope of this paradigm to a large range of non-conjugate latent variable models. 
For models with one level of latent variables, \alg{} is identical to the mirror descent algorithm of \citeAY{Khan2017}. However, our derivation of the algorithm
(which was done independently and prior to the publication of \cite{Khan2017})
is simpler and more direct. More importantly, for models with more than one level of hidden variables, \citeAY{Khan2017} uses a mean field approximation and does not use the optimal structured 
approximation. 

Our second contribution, which is specific for 
models with latent Gaussian variables,
is a variant algorithm, \hsvifull{} (\hsvi), 
that uses the optimal approximation but combines natural gradients for the covariance with standard gradients on the mean. This algorithm leads to significant improvements in convergence in many models. 

Building on this we
develop the details of \alg\ for several challenging classes of models, 
including GME, sGP, PMF, and CTM,
all of which are members of the LGM sub-family of the \lvm{} where the first set of hidden variables is Gaussian.
For sGP, the optimal solution requires cubic time but we propose two variants which are efficient and provide a significant improvement over current best practice.
Previous work on PMF is restricted by developing algorithms for specific likelihoods (\cite{Hernandez-LobatoHG14, KoenigsteinP13,KoK14,KhanKS14,PaquetTW12}), making strong independence assumptions (mean-field, \cite{Hernandez-LobatoHG14}), weakening the variational bound (\cite{SeegerB12, Hernandez-LobatoHG14,KoenigsteinP13} use different local approximations to the bound) or working in the batch setting (all but \cite{Hernandez-LobatoHG14}) limiting applicability to big data. 
We show that fixed-point updates can be calculated for general non-conjugate PMF and that this yields an effective instance of \alg. 
The resulting algorithm is generic, i.e., applicable to any observation likelihood, enjoys the benefits of \alg{}, and 
avoids the limitations of previous work on PMF.
The CTM model \cite{Blei2006} is a non-conjugate extension of LDA \cite{Blei2003}. Previous work \cite{Blei2006} used the mean field approximation which is sub-optimal and 
the application of \alg{} to CTM was posed as an open problem in \cite{Hoffman2013}.
Here too we show that fixed point updates can be developed and that this yields an effective instance of \alg.

An experimental evaluation in the LGM sub-family illustrates the use of the approach across models. 
In particular, we demonstrate
(i)
the advantage of the optimal structured approximation over the mean field approximation and over a simpler structured approximation from previous work,
(ii) the advantage of natural gradients over standard gradients,
and
(iii) the improved performance of  \hsvi{} over ``pure" variants in latent Gaussian models in terms of convergence speed w.r.t.\ the training set optimization objective and w.r.t.\ test set error rate. 
Although the \alg{} approach is generic to the entire \lvm{} family, application to a specific model still requires a significant amount of effort in developing the details of derivatives and appropriate computational structure.
This is in contrast with ``black-box" algorithms which do not require such additional development. 
Similarly, our algorithm is limited to two level models, but one could conceivably seek applications more generally in graphical models, for example, in the spirit of variational message passing \cite{WinnB05}. We discuss these challenges and related work further in the concluding section of the paper. 

\section{Background}
\label{sec:background}

This section describes the model family and reviews technical preliminaries from previous work on variational lower bounds (slightly generalizing previous work), natural gradients, fixed-point updates and SVI. 

\subsection{Two-Level Latent Variable Model}

We consider a latent variable model with two levels of latent variables (\lvm{}) of the form
\begin{equation}
\label{eq:model}
w\sim p(w), \ \ \ \ \ f|w\sim p(f|w),   \ \ \ \ \ 
p(y|f)= \prod_i p(y_i|f_i).
\end{equation}
This generalizes several existing models including generalized linear models (GLM), Gaussian processes (GP), sparse GPs, GP latent variable model (GPLVM), probabilistic matrix factorization (PMF), latent Dirichlet allocation (LDA), correlated topic models (CTM) and more.
Concrete detailed instantiations are illustrated later in the paper.
The LGM sub-family, where the global latent variables $w$ are normally distributed, $p(w)={\cal N}(\mu,\Sigma)$, includes several of these models and has been studied extensively.  
Importantly, the \lvm{} described by (\ref{eq:model}) does not assume any conjugacy relationships between the \emph{global} latent variables $w$, the \emph{local} latent variables $f$, and the observations $y$.

For the sequel we define two special cases of this family. 
In the first, \lvm{}d, $f$ is a deterministic function of $w$. GLM, GP and PMF are families of models satisfying this restriction. In the second, \lvm{}i, $f$ is random and factors into conditionally independent components similar to $y$ so that $p(f|w)=\prod_i p(f_i|w)$. 
LDA and CTM  are families of models satisfying this restriction. 
\lvm{}d is a special case of \lvm{}i since due to determinism we can take a product of the delta functions giving $p(f_i|w)$. But the distinction is useful because \lvm{}d affords a simpler analysis.
Sparse GPs is an example of a family that does not fit in the restricted subsets. 

\subsection{Variational Lower Bounds}

Given observations $y$, the main task considered in this paper is to marginalize out the local latent variables and calculate the posterior on the global latent variables.
Since this is in general intractable, we use a variational approximation for the distribution over latent variables denoted $q(w,f)$. The variational approach leads to a lower bound approximation
(known as VLB or ELBO) on the marginal likelihood that can be used for choosing the best approximation and for model selection.  
However, 
to achieve such a simplification in inference  one has to restrict the form of $q(w,f)$, since otherwise the optimal choice yields $q(f,w)=p(f,w|y)$ which recovers the original problem.

Different choices have been made in previous work about this form. The simplest approach uses the mean field approximation which assumes that all components are independent, that is, $q(w,f)=\prod_j q(w_j)\prod_i q(f_i)$.
This approach has been used for example in the LDA and CTM models \cite{Blei2003,Blei2006}.

The most common approximation in LGM including GLM, GP, GPLVM, and sparse GP models (see e.g., \cite{Challis2013,Titsias2010,Titsias2009,Sheth2015,ShethKh2016}) captures more structure, using

\begin{equation}
\label{eq:SimplestructuredQ}
q(w,f)=q(w)p(f|w)
\end{equation} 
where the posterior marginal on $w$ is approximated by $q(w)$ and some dependence of $f$ on $w$ is captured by $p(f|w)$. 
However, this is rather limited since the dependence is fixed to its form in the prior $p(f|w)$ and $q()$ cannot adjust it.
Using this simple approximation 
one can derive a convenient lower bound  on the marginal likelihood:
\begin{eqnarray} %
\log p(y) %
& = & 
 \log \int  p(w) p(f|w) \prod_i p(y_i|f_i) df dw \nonumber \\
& \geq & 
\int  q(w,f) \log \left( \frac{ p(w) p(f|w)}{q(w,f)} \prod_i p(y_i|f_i) \right) df dw \nonumber \\
& = & -d_{KL}(q(w) \| p(w)) + \sum_i E_{q(w,f)} [\log p(y_i|f_i) ]   \nonumber \\
& = & -d_{KL}(q(w) \| p(w)) + \sum_i E_{q(f_i)} [\log p(y_i|f_i) ]  \label{eq:vlb}
\end{eqnarray}
where $d_{KL}$ is the Kullback-Leibler divergence. 
Note that the effect of $p(f|w)$ is implicitly captured through the marginal distribution $q(f_i)$, and that it does not affect the $d_{KL}$ term.

More recently, \cite{Hoffman2015} proposed the following so-called structured approximation for models within \lvm{}i which allows more flexibility in capturing the dependence between the global and local latent variables 
\begin{equation}
\label{eq:truestructuredQ}
q(w,f)=q(w)q(f|w)
\end{equation} 
where some assumptions on $q(w)$ and $q(f|w)$ (for example, restricting the form of $q(w)$) are used to obtain a simplification. A similar construction was developed in the collapsed variational bound of \cite{TehNW06}.

We note that mean field and the simple variational distribution of Eq~(\ref{eq:SimplestructuredQ}) are not comparable in that they make complementary assumptions on the distribution. But the form of Eq~(\ref{eq:truestructuredQ}) subsumes both of these and therefore has a potential to provide a strictly better bound.

As shown by \cite{Hoffman2015}, (\ref{eq:truestructuredQ}) can be used to derive a tighter lower bound where $q(f|w)$ is chosen optimally. The same argument holds more generally for the \lvm{}. We have

\begin{eqnarray} %
\log p(y) %
& = & 
 \log \int  p(w,f,y) \ df dw \nonumber \\
& \geq & 
\int  q(w)q(f|w) \log \left( \frac{ p(w) p(y|w)p(f|y,w)}{q(w)q(f|w)}  \right) df dw \nonumber \\
& = &  -d_{KL}(q(w) \| p(w))  +
\int  q(w) \left[\int  q(f|w) df\right]  \log p(y|w) dw 
\nonumber \\
&  & \ \ \  
- \int  q(w) \left[\int  q(f|w) \log \frac{q(f|w)}{p(f|y,w)} df\right]  dw 
\nonumber \\
\end{eqnarray}
where the second line decomposes $p(w,f,y)$, differently from the generative process, in a manner that is convenient for the argument. 
Next we observe that the
square brackets in the third line integrate to 1 and the term in square brackets in the fourth line is equal to $d_{KL}(q(f|w) \| p(f|w,y))$. 
We can optimally minimize this term, for every $w$, to get $d_{KL}=0$ by choosing $q(f|w)=p(f|w,y)$.
This yields
\begin{eqnarray} %
\log p(y)  & \geq & -d_{KL}(q(w) \| p(w)) + E_{q(w)} [\log p(y|w) ] 
\nonumber \\
& = & -d_{KL}(q(w) \| p(w)) + E_{q(w)} \left[  \log E_{p(f|w)} [ \prod_i p(y_i|f_i) ] \right] .
\label{eq:vlbT}
\end{eqnarray}
Comparing the two final VLBs we see that the expectation term in (\ref{eq:vlbT}) is more complex and less convenient than the one in (\ref{eq:vlb}). In particular, the decomposition into sum in  (\ref{eq:vlb}) which is amenable to stochastic gradients does not occur in (\ref{eq:vlbT}).
An alternative view of  (\ref{eq:vlbT}) is simply as a result of integrating $f$ out of the model at the outset and deriving a variational bound using $q(w)$. However, as the discussion below shows, the explicit use of $f$ in the bound is useful both computationally and as a tool to understand optimality of other approximations.
We make 3 additional observations on these bounds and their potential use.

First, note that applying Jensen's inequality, $\log E_{p(f|w)} [ \prod_i p(y_i|f_i) ] \geq E_{p(f|w)} [ \log \prod_i p(y_i|f_i) ] $, in (\ref{eq:vlbT}) yields exactly the bound of (\ref{eq:vlb}). This shows, as expected, that (\ref{eq:vlbT}) is tighter, and in addition that the simpler variational bound can be alternatively derived with this additional approximation step.
Second, note that in \lvm{}d where $p(f|w)$ is deterministic, the simple approximation is identical to the general one 
(because $p(f|w)=p(f|y,w)$). Therefore for \lvm{}d we can use (\ref{eq:vlb}) instead of (\ref{eq:vlbT}).
Third, we consider \lvm{}i where both $f$ and $y$ are products of independent factors. In this case, 
\begin{align} %
p(y|w)  
& =  \int_f p(f|w)p(y|f) df  = \int_f \prod_i p(f_i|w)p(y_i|f_i) df
\nonumber \\
& = \prod_i \int_{f_i}  p(f_i|w)p(y_i|f_i) d{f_i}  =  \prod_i E_{p(f_i|w)} [ p(y_i|f_i)] =  \prod_i p(y_i|w)
\nonumber \\
\end{align}
and the bound in (\ref{eq:vlbT}) simplifies to 
\begin{eqnarray} %
\log p(\mbox{data})  & \geq & -d_{KL}(q(w) \| p(w)) + \sum_i E_{q(w)} [\log p(y_i|w) ] 
\nonumber \\
& = & -d_{KL}(q(w) \| p(w)) + \sum_i  E_{q(w)} \left[  \log E_{p(f_i|w)} [ p(y_i|f_i) ] \right] 
\label{eq:vlbi}
\end{eqnarray}
which decomposes into a sum over $i$ and hence amenable to stochastic gradient ascent. 

In summary, we have the following observations:
\begin{itemize}
\item
An optimal bound (\ref{eq:vlbT}) holds for the \lvm{}. Unfortunately, this bound  does not in general decompose into a sum over examples. 
\item
An optimal decomposable bound (\ref{eq:vlb}) is available for \lvm{}d, and an optimal decomposable bound  (\ref{eq:vlbi}) is available for \lvm{}i. %
\item
Alternatively a decomposable bound is obtained as (\ref{eq:vlb}) by making a stronger approximation. This approach was used in previous work on sparse GPs and GPLVM. 
The tighter bound in (\ref{eq:vlbT})  potentially offers a better approximation. 
\end{itemize}

\subsection{Fixed-Point Updates}

Consider the case where the prior distribution $p(w)$ and the assumed marginal posterior $q(w)$ are of the same exponential family type:
\begin{align}
    p(w) & = \exp \left( t(w)^T \theta_p - F(\theta_p) \right) h(w) \\
    q(w) & = \exp \left( t(w)^T \theta_q - F(\theta_q) \right) h(w)
\end{align}
where $\theta_{p}$ and $\theta_q$ denote the canonical (natural) parameters of $p$ and $q$.
Let $\eta_p$ and $\eta_q$ denote the expectation (mean) parameters of $p$ and $q$ (i.e., $E_p(t(w))$ and $E_q(t(w))$) and recall that in the exponential family we have
$\frac{\partial F(\theta)}{\partial\theta} = \eta$
and 
$\frac{\partial\eta}{\partial\theta} = I(\theta)$ where $I$ is the Fisher information matrix \citep{Amari2000}.

We next consider optimizing the canonical parameters of $q(w)$ using natural gradients \citep{Amari2000}.  
Natural gradients adapt to the geometry of the objective function and have been demonstrated to converge faster than standard gradients in some cases. 
The natural gradient pre-multiplies the standard gradient by the inverse of the Fisher information matrix $I$.
It has been shown \cite{Sato2001,HensmanRL12,Hensman2013} that the natural gradient update of the canonical parameters for $q(w)$, using a step size 1, is given by
\begin{align}
    \theta_q & \leftarrow  \theta_q + I(\theta_q)^{-1} \frac{\partial\text{VLB}}{\partial\theta_q} %
    = \theta_q + I(\theta_q)^{-1} \frac{\partial\eta_q}{\partial\theta_q} \frac{\partial\text{VLB}}{\partial\eta_q} %
    = \theta_q + \frac{\partial\text{VLB}}{\partial\eta_q}.
    \label{eq:update}
\end{align}
Following this, \cite{ShethKh2016} has shown that (\ref{eq:update}) corresponds to a fixed-point update of $\theta_q$. To see this recall that the VLB has two terms.
The Kullback-Liebler divergence between $q(w)$ and $p(w)$ is given by
\begin{equation}
    \text{KL}(q\Vert p) = \eta_q^T (\theta_q - \theta_p) - (F(\theta_q) - F(\theta_p))
\end{equation}
and the derivative 
is given by
\begin{align}
    \frac{\partial\text{KL}(q\Vert p)}{\partial \eta_q} & = \theta_q - \theta_p + \left(\frac{\partial\theta_q}{\partial\eta_q}\right)^T \eta_q - \left(\frac{\partial\theta_q}{\partial\eta_q}\right)^T \frac{\partial F(\theta_q)}{\partial\theta_q} %
    = \theta_q - \theta_p. 
\end{align}

Now, denote  the second term in the VLB expression (\ref{eq:vlbT})  by $T$
and denote
\[
G(\eta_q)= \frac{\partial }{\partial\eta_q} T(\eta_q)
\]  
where we have emphasized the dependence on $\eta_q$. Applying this notation to (\ref{eq:update}) yields the batch fixed-point update
\begin{align}
\label{eq:expfamilyUpdateT}
    \theta_q & \leftarrow   \theta_p + G(\eta_q).
\end{align}

Similarly, denoting the terms inside the sum in the VLB expressions (\ref{eq:vlb}) or (\ref{eq:vlbi}) by $T_i$
and denoting
\begin{equation}
G(\eta_q)=\sum_i G_i(\eta_q)=\sum_i \frac{\partial }{\partial\eta_q} T_i(\eta_q)
\label{eq:def_Gi}
\end{equation}
we get
\begin{align}
\label{eq:expfamilyUpdate}
    \theta_q & \leftarrow  \theta_q  - [ \theta_q - \theta_p] +\sum_i G_i(\eta_q) = \theta_p + \sum_i G_i(\eta_q).
\end{align}

Since $\eta_q$ is a function of $\theta_q$, this is a fixed-point update.
If the terms $G_i(\eta_q)$ do not depend on $\theta_q$ then this is a closed-form update. 
In this case, the update is equivalent to equating the gradient to zero and solving for $\theta_q$, i.e., a steepest ascent update. 
However, in general a dependence exists and the update needs to be repeated. 
The fixed-point update is a natural gradient step with a fixed step size and it does not in general guarantee an increase in the objective so convergence is not guaranteed. 
The work of \cite{ShethKh2016} identified conditions on LGM where the fixed point of this update equation, when applied to the covariance parameter, is the optimal variational solution.

\subsection{SVI}

SVI \cite{Hoffman2013} and structured SVI \cite{Hoffman2015} were developed for conjugate latent variable models in the exponential family. SVI works by applying stochastic updates in the natural gradients space.
In this sense, it is of course applicable in any model. 
However, the main observation in \cite{Hoffman2013} is that, in models with conjugate complete conditionals in the exponential family, natural gradient updates have a particularly simple closed form. 
The reason is that, similar to the argument above, size 1 natural gradient updates are equivalent to coordinate-wise steepest ascent optimization. 
From this, simple algebraic properties show that stochastic updates with any step size also have a simple form. 

Therefore, the SVI paradigm combines the advantage of having a simple and efficient update formula, the speed and convergence of stochastic gradients, and the advantages of using gradients in the natural space leading to superior performance in practice. 
However, to date, the application of this paradigm was limited to 
models with conjugate complete conditionals in the exponential family \cite{Hoffman2013,Hoffman2015}.

\section{\algfull{}}
\label{sec:MCSVI}

This section develops our main contributions. We start by making the observation that the paradigm of structured SVI is applicable more generally than previously observed and is applicable to the \lvm{} defined above. 
This turns out to be a straightforward application of the fixed-point updates and is achievable as long as 
an approximation for
$G$ in (\ref{eq:expfamilyUpdateT})
(or $G_i$ in (\ref{eq:expfamilyUpdate}))
is computable, for example, through Monte Carlo estimates.
 We then provide several applications demonstrating the generality of these ideas.

The first application starts with a relatively simple model, illustrating how 
the bound (\ref{eq:vlb}) 
can be applied
to the special case of LGM where $p(f|w)$ is linear Gaussian. This includes GLM as a special case and is optimal in that case since GLM is in \lvm{}d. We note that even this model is non-conjugate because $p(y_i|f_i)$ is not restricted.

The second application extends GLM by adding a Rayleigh mixture of Gaussian noise to the local latent variables $f_i$.
By explicitly modeling noise at the pre-observation level, the GME model can potentially infer narrow posteriors on the weight vector whereas GLM would transfer noise in $f_i$ to the posterior on the weight vector.
In this case, there is nonconjugacy between the likelihood $p(y_i|f_i)$ and local conditional $p(f_i|w)$ as well as between the local conditional $p(f_i|w)$ and the Gaussian-Rayleigh prior.
This model is in LVMi where the VLB (\ref{eq:vlbi}) is optimal.

The third application in non-conjugate PMF uses the bound (\ref{eq:vlb}) which is optimal since PMF is in \lvm{}d.
This is also a latent Gaussian model, but the relationship $p(f|w)$ between the global and local latent variables is not linear. This application yields a general algorithm for PMF with any observation likelihood function. 
The significance of this development is in being both efficient and generally applicable, removing the need to develop a special algorithm in each case of observation likelihood, common in previous work. 
In addition, when applied to conjugate PMF with Gaussian likelihood, our derivation yields the batch steepest ascent algorithm of \cite{LimTeh2007} as a special case. 

The fourth application in CTM uses (\ref{eq:vlbi}) which is optimal since CTM is in \lvm{}i.
This is also an instance of LGM. But, in this model, the variable corresponding to $f$ is discrete unlike the previous models.  The application of structured SVI to CTM was posed as an open problem in \cite{Hoffman2013}.

\subsection{\algfull{} Updates in \lvm{}}

Recall
that in our model the natural gradient has the form
$- [ \theta_q - \theta_p] + G(\eta_q)$
where in the cases with decomposable bound we have
$G(\eta_q) = \sum_i G_i(\eta_q)$.

To facilitate the discussion below, let
$F=\theta_p +G$
so that the natural gradient is equal to $-\theta_q+F$ and the update rule 
in (\ref{eq:expfamilyUpdate})
is 
\begin{equation}
\label{eq:steepest}
\theta_q\leftarrow \theta_q + (-\theta_q +F) = F.
\end{equation}
When $G(\eta_q)$ is not a function of $\theta_q$, $F$ is the steepest ascent optimizer for $\theta_q$. 
This is exactly the case for the conjugate model of \cite{Hoffman2013} where the SVI algorithm was developed. 
In the more general case, $F$ is a function of $\theta_q$ which provides the batch fixed-point update of the natural parameter.

To apply stochastic gradients \cite{robbins-monro}, a random sample of the gradient is obtained and a standard ascent step with step size $\rho$ is applied, where the schedule for the step size satisfies standard conditions. %
Since $\theta_p$ and $\theta_q$ are parameters, we only need an estimate $\hat{G}$ of $G$.
Letting $\hat{F}=\theta_p +\hat{G}$, the stochastic update is
\begin{equation}
\label{eq:SVIupdateGeneric}
\theta_q\leftarrow \theta_q + \rho (-\theta_q+\hat{F})=(1-\rho) \theta_q + \rho \hat{F}.
\end{equation}

Hoffman et.\ al.\ \cite{Hoffman2013} observed that, when $\hat{F}$ is obtained from a mini-batch sample, this update has an interesting and useful interpretation. 
For example, let $\hat{F} = \theta_p +\frac{N}{|\minibatch|}\sum_{i\in \minibatch} \hat{G}_i(\eta_q)$ be a stochastic gradient estimated by sampling a mini-batch $\minibatch$ uniformly at random from a dataset of size $N$. %
In this case, the optimal update for the mini-batch using (\ref{eq:steepest}) is $\theta_q\leftarrow\hat{F}$ and the update 
(\ref{eq:SVIupdateGeneric}) can be seen to interpolate between the old value of $\theta_q$ and the optimal solution for the mini-batch. Our main observation is that the same holds when $F$ is a fixed-point update and also when each term $\hat{G}_i(\eta_q)$ is a Monte Carlo estimate of $G_i$ (for example as in \cite{Titsias2014,Kingma2014,Rezende2014,Ranganath2014}).  
\emph{The conjugacy relation and the optimality of the update (\ref{eq:steepest}) are not required.}

Our first proposed algorithm, \algfull{} (\alg{}) performs updates on a natural parameter using a stochastic natural gradient:
\begin{equation}
\label{eq:generalSVIupdate}
\theta_q\leftarrow  
(1-\rho) \theta_q + \rho (\theta_p +\hat{G}(\eta_q)) = 
(1-\rho) \theta_q + \rho (\theta_p +\frac{N}{|\minibatch|}\sum_{i\in \minibatch} \hat{G}_i(\eta_q))
\end{equation}
where the right-most form is applicable when the bound is decomposable. 
When the model includes further parameters (or hyperparameters), we follow standard practice and perform a gradient (or closed-form, if possible) update of the parameters after each update of $\theta_q$, thus learning the parameters online together with the posterior for~$w$.

To summarize, we have shown that \alg{} is applicable wherever $p(w)$ and $q(w)$ have the same exponentially family form and the terms ${G_i}$ can be efficiently computed or approximated, e.g., via sampling.  
This weakens the condition for conjugate complete conditionals in prior work. 
\alg{} shares the advantages of SVI: a simple and efficient update formula, the speed and convergence of stochastic gradients, and the advantages of being able to use gradients in the natural space.

At this point it is worth emphasizing the similarity to the DSVI algorithm of 
\cite{Titsias2014} which also performs variational inference in non-conjugate models with the use of Monte Carlo samples. This algorithm is also related to algorithms in \cite{Rezende2014,Kingma2014} that aim at neural network models.
DSVI was developed for a model with one level of hidden variables  but the same idea is applicable here.\footnote{
This is similar to integrating out $f$ from our model leading to a more complex conditional probability for $y$ as expressed in the bound of (\ref{eq:vlbT}).
}
However, the main difference is that DSVI performs standard gradient ascent for parameter optimization in the standard space, whereas \alg{} updates incorporate natural gradients on natural parameters. 
Below, we use the name \dsvi{} to refer to the analogous approach using a structured variational approximation but  
using standard gradients in the standard parameter space.
Letting $\phi$ denote the standard parameters, the \dsvi{} update is given by
\begin{equation}
    \phi_q  \leftarrow  \phi_q + \rho \frac{\partial\text{VLB}}{\partial\phi_q} 
    \approx \phi_q  +\rho\left(-\frac{\partial d_{KL}(q(w)\Vert p(w))}{\partial \phi_q}  + \frac{N}{|\minibatch|}\sum_{i\in\minibatch} \hat{G}_i(\phi_q)\right)
    \label{eq:DSVIupdate}
\end{equation}
where $G_i(\phi_q)$ is defined as in (\ref{eq:def_Gi}), but with the derivative taken with respect to $\phi_q$.

In LGM, the standard parameterization of the variational distribution is in terms of the mean and Cholesky factor of the covariance
and $\phi_q$ refers to these parameters. 
For LGM we propose an additional algorithm which is motivated by the results of \cite{ShethKh2016}.
In particular, 
\cite{ShethKh2016} showed that, in LGM, fixed-point updates based on natural gradients for the covariance are very effective whereas the same type of updates for the mean are less stable and that occasionally they lead to degradation in performance. 
Based on this observation we propose a hybrid algorithm, \hsvi, which 
updates the covariance using natural gradients through (\ref{eq:generalSVIupdate}) but updates the mean using standard gradients through (\ref{eq:DSVIupdate}).

Our experiments in LGM models provide comparisons of \alg, \hsvi{} and \dsvi.
The experiments confirm the advantage of using natural gradients and show that \hsvi{} provides the best performance of the 3 variants. 

\subsection{Related Algorithms}
\citeAY{Khan2017} developed an algorithm using
stochastic mirror-descent in the mean-parameter space. For \lvmone{}, this update is identical to 
\alg{}. 
However, 
for \lvm{} their algorithm uses
the mean field approximation. In addition, the derivation through fixed points is more direct and makes the connection to \cite{Hoffman2015} obvious. 

The work by \citeAY{salimans2013fixed} proposes an explicit
estimation of the Fisher information matrix to perform natural
gradient updates in a 1L model. This yields a more general algorithm
for 1L at the expense of increased computational cost by ignoring the
structure of $p()$ and $q()$. In the hierarchical case,
\cite{salimans2013fixed} explicitly estimate $q(w)$ and $q(f|w)$ limiting
their form which results in a sub-optimal solution, whereas our
approach retains optimality in the 2L case.

Other related work includes ``black-box'' inference algorithms and
related sampling schemes as in \cite{Kingma2014,Titsias2014,Rezende2014,Ranganath2014,Kingma2015}
and more recent work by others.
For example, \cite{Ranganath2014} develops
a black-box sampling based method for the 2L model and
\cite{KucukelbirTRGB16} develops an extension combining sampling with
automatic differentiation for a large class of models. However, both
use the mean field approximation and standard gradients. Further work
would be required to handle the marginalization used in the structured
approximation in 2L models with the black box scheme.

\subsection{\alg{} for LGM with Linear Gaussian \texorpdfstring{$p(f|w)$}{p(f|w)} using the VLB (\ref{eq:vlb})}

This section applies \alg{} to a simple model thereby illustrating the key concepts used in the application to more complex models in the following sections. %
As discussed above, the application of VLB (\ref{eq:vlb}) to LGM is not always optimal but it has nonetheless been shown to be useful across several models. In the special case of GLM (as applied in the experimental section) it is optimal since GLM is in \lvm{}d.

In LGM, letting the standard parameters of mean and covariance be denoted by $(m,\cov)$,
the natural parameters $\theta$ are equal to 
$(\cov^{-1}m,\frac{1}{2}\cov^{-1})$, 
and the expectation parameters $\eta=(h,H)$ are equal to $(m,-(\cov+m m^T))$.
Focusing on the covariance parameter, $G_i$ evaluates to 
$\frac{\partial E_{q(f_i)} [\log p(y_i|f_i)] }{\partial H} = - \frac{\partial E_{q(f_i)} [\log p(y_i|f_i)] }{\partial \cov} \triangleq -D_i$.

When $p(f|w)$ is linear Gaussian, the global and local latent variables have a conjugate relationship (though the model is still not conjugate since $p(y|f)$ is not restricted in this manner).
In this case, \cite{ShethKh2016} showed that the terms $D_i$ 
 take the form 
\begin{equation}
\label{eq:GLMGi}
D_i =  
\frac{\partial E_{q(f_i)} [\log p(y_i|f_i)] }{\partial v_{i}}\frac{\partial v_{i} }{\partial \cov}
\end{equation}
where $q(f_i)=\mathcal{N}(m_i,v_i)$, $m_i(m)=a_i+h_i^T m$, and $v_i(\cov)=c_i + h_i^T \cov h_i$ with $a_i, c_i, h_i$ determined by the particular model instantiation. %
For example, in GLM, where $w$ represents the weight vector and $f=H^T w$ with $H$ representing the design matrix, $a_i=c_i=0$ and $h_i$ is a row of $H$.
As shown by   
\cite{ShethKh2016}, 
if $\log p(y_i|f_i)$ is differentiable, then the univariate derivative 
can be computed as
\begin{equation}
    \gamma_i = \frac{\partial E_{q(f_i)} [\log p(y_i|f_i)]}{\partial v_i} = \frac{1}{2} E_{{\cal N}(f_i|m_{i},v_{i})}\left[ \frac{\partial^2}{\partial f_i^2}  \log p(y_i|f_i)\right].  
\label{eq:rhoi-first}
\end{equation}
Now, the fixed-point update (\ref{eq:expfamilyUpdate}) for the natural parameter $\frac{1}{2}\cov^{-1}$ is given by
\begin{equation}
\label{eq:FP}
\frac{1}{2}\cov^{-1}\leftarrow
\frac{1}{2}\Sigma^{-1} - \sum_i D_i  = \frac{1}{2}\Sigma^{-1} - \sum_i \gamma_i h_i h_i^T 
\end{equation}
and
applying (\ref{eq:generalSVIupdate}) to this model gives the \alg{} update for the covariance:
\begin{equation}
\label{eq:LGM-SVI-update-V}
\cov^{-1}\leftarrow  (1-\rho) \cov^{-1} + \rho (\Sigma^{-1} -2\frac{N}{|\minibatch|}\sum_{i\in \minibatch} \hat{\gamma}_i h_i h_i^T)
\end{equation}
where the univariate $\hat{\gamma}_i$s are estimates of (\ref{eq:rhoi-first}), e.g., computed with Monte Carlo sampling.

A similar series of steps can be used to derive the \alg{} update for the mean:
\begin{equation}
\label{eq:LGM-SVI-update-mean}
    \cov^{-1} m \leftarrow  (1-\rho) \cov^{-1}m + \rho \left(\Sigma^{-1} \mu + \frac{N}{\vert \minibatch\vert}\sum_{i\in \minibatch} (\hat{\alpha}_i -2 (m^T h_i)\hat{\gamma}_i)h_i\right)
\end{equation}
where $\hat{\alpha}_i$ is an estimate of 
$\frac{\partial}{\partial m_i}E_{\mathcal{N}(f_i\vert m_i,v_i)}[\log p(y_i\vert f_i)]
=
E_{{\cal N}(f_i|m_{i},v_{i})}[ \frac{\partial}{\partial f_i}  \log p(y_i|f_i)]
$.

\paragraph{Concrete Algorithms for LGM:}
The discussion above yields concrete instances of the three algorithms for LGM. \alg{} uses (\ref{eq:LGM-SVI-update-V}) and (\ref{eq:LGM-SVI-update-mean}) to update the variational parameters. 
Similarly, the \hsvi{} algorithm 
updates the covariance using (\ref{eq:LGM-SVI-update-V}) but updates the mean using standard gradients (equation given in the appendix as (\ref{eq:LGM-dsvi-update-mean})).
The \dsvi{} uses
(\ref{eq:LGM-dsvi-update-C}) and (\ref{eq:LGM-dsvi-update-mean}) to update the Cholesky factor and mean parameters respectively.

\subsection{\alg{} for GME using the VLB (\ref{eq:vlbi})}

We consider a simple form of GME which extends GLM by adding to $f_i$ Gaussian noise with Rayleigh-distributed variance.
With an augmented global variable $w=(w_1,w_2)$, $f_i|w \sim \mathcal{N}(f_i|w_1^T x_i, w_2)$ where $w_1$ is Gaussian and $w_2 \sim \text{Rayleigh}(w_2|\tau^2)$.
Experiments demonstrate that this indeed holds and GME reduces the error (negative log likelihood on test) of the learned model when compared to GLM.

\newcommand{\word}{y}
\newcommand{\topic}{f}
\newcommand{\glob}{w}
\newcommand{\ftest}{f_*}
\newcommand{\xtest}{x_*}
\newcommand{\ytest}{y_*}
\newcommand{\D}{\text{data}}

The Rayleigh density with parameter $\tau^2$, has natural parameter $-\frac{1}{2\tau^2}$, and expectation parameter $g=2\tau^2$.
Let the variational distribution be $q(w_1) q(w_2)$ where $q(w_1)=\pdfnorm{w_1}{m}{\cov}$ and $q(w_2)$ is given by the Rayleigh density with parameter $\sigma^2$, $\text{Rayl}(w_2|\sigma^2)$.
The Kullback-Leibler divergence from $\text{Rayl}(w_2|\sigma^2)$ to $\text{Rayl}(w_2|\tau^2)$ is equal to $\log\frac{\tau^2}{\sigma^2} + \frac{\sigma^2}{\tau^2} - 1$.
For optimization of the optimal VLB, derivatives of $\expec{q(w_1)q(w_2)}{\log \phi(w_1,w_2)}$ w.r.t. expectation parameters $h, H$, and $g$ are needed, where 
\begin{equation}
    \phi(w_1, w_2) = \expec{\pdfnorm{f}{w_1^T x}{w_2}}{p(y\vert f)}
    \label{eq:opt-phi}
    .
\end{equation}
By the multivariate normal identities 
(\ref{eq:mvn_iden_m}) and (\ref{eq:mvn_iden_V}) and the matrix chain rule,
\begin{align}
    \pd{}{h}\sbr{
        \expec{q(w_1)q(w_2)}{
            \log \phi(w_1, w_2)
        }
    }
    & = 
    \expec{q(w_1)q(w_2)}{
        \nabla_{w_1}
        \log \phi(w_1, w_2)
        -
        \nabla_{w_1}^2
        \log \phi(w_1, w_2)
        m
    }
    \label{eq:opth}
    \\
    \pd{}{H}\sbr{
        \expec{q(w_1)q(w_2)}{
            \log \phi(w_1, w_2)
        }
    }
    & = 
    -\frac{1}{2}
    \expec{q(w_1)q(w_2)}{
        \nabla_{w_1}^2
        \log \phi(w_1, w_2)
    }
    \label{eq:optH}
    ,
\end{align}
where
\begin{align}
    \nabla_{w_1}
    \log \phi
    & = 
    \frac{1}{\phi}
    \nabla_{w_1} \phi
    \label{eq:opt-phi-first-deriv-wone}
    ,
    \\
    \nabla_{w_1}^2
    \log \phi
    & = 
    \frac{1}{\phi}
    \del{
        \phi \nabla_{w_1}^2 \phi
        -
        \del{\nabla_{w_1} \phi}
        \del{\nabla_{w_1} \phi}^T
    }
    \label{eq:opt-phi-second-deriv-wone}
    ,
\end{align}
and
\begin{align}
    \nabla_{w_1} \phi
    & = 
    x
    \expec{\pdfnorm{f}{w_1^T x}{w_2}}{\pd{}{f} p(y\vert f)}
    \\
    \nabla_{w_1}^2 \phi
    & = 
    x x^T
    \expec{\pdfnorm{f}{w_1^T x}{w_2}}{\pd[2]{}{f} p(y\vert f)}
    .
\end{align}
The reparameterization trick \citep{Kingma2014} can be used to re-write the expectation w.r.t. $q(w_2)$ as %
\begin{align}
    \expec{q(w_2)}{
        \expec{q(w_1)}{
            \log \phi(w_1, w_2)
        }
    }
    & = 
    \expec{\text{Rayl}(\alpha\vert 1)}{
        \expec{q(w_1)}{
            \log \phi(w_1, \alpha\sigma)
        }
    }
    \label{eq:expqrayl}
    .
\end{align}
Letting $\psi(w_2) = \expec{q(w_1)}{\log \phi(w_1, w_2)}$, the derivative of (\ref{eq:expqrayl}) w.r.t. $g$ is given by
\begin{align}
    \expec{\text{Rayl}(\alpha\vert 1)}{
        \eval{
            \pd{}{w_2}
            \psi(w_2)
        }_{w_2=\alpha\sigma}
        \frac{\alpha}{4\sigma}
    }
    = 
    \expec{q(w_2)}{
        \frac{w_2^2}{4\sigma^2}
        \pd{}{w_2}
        \psi(w_2)
    }
    \label{eq:derivexpqrayl}
    ,
\end{align}
where in the LHS of (\ref{eq:derivexpqrayl}), $\alpha\sigma=\frac{\alpha}{\sqrt{2}}\sqrt{2\sigma^2}$ is used.
In summary,
\begin{align}
    \pd{}{g}
    \expec{q(w_1)q(w_2)}{
        \log \phi(w_1, w_2)
    }
    & =
    \expec{q(w_1)q(w_2)}{
        \frac{w_2^2}{4\sigma^2} \pd{}{w_2}\log\phi
    }
    \label{eq:opt-phi-first-deriv-wtwo}
    ,
\end{align}
where
\begin{equation}
    \pd{}{w_2}\log \phi
    =
    \frac{1}{2\phi} 
    \expec{\pdfnorm{f}{w_1^T x}{w_2}}{\pd[2]{}{f} p(y\vert f)}
    .
\end{equation}

Note, when only $\ell(f) = \log p(y|f)$ and its derivatives are available, the required derivatives can be computed as 
\begin{align}
    \pd{}{f} p(y|f)
    & = 
    \exp\del{\ell(f)} \pd{}{f} \ell(f)
    \\
    \pd[2]{}{f} p(y|f)
    & = 
    \exp\del{\ell(f)} 
    \del{
        \pd[2]{}{f} \ell(f) + \del{\pd{}{f} \ell(f)}^2
    }
    .
\end{align}

Using the shorthand notation $E_i=E_*[p(y_i|f_i)]$, 
$E'_i=E_*[\frac{\partial}{\partial f_i}p(y_i|f_i)]$,
$E''_i=E_*[\frac{\partial^2}{\partial f_i^2}p(y_i|f_i)]$, 
where the expectations $E_*$ are w.r.t.\ ${p(f_i|w_1^Tx_i,w_2)}$.
The optimal approximation update equations for standard and natural parameters are given by
\begin{align}
    m 
    & \leftarrow 
    m 
    + 
    \rho \del{
        -\Sigma^{-1}\del{m-\mu}
        +
        \expec{q(w_1)q(w_2)}{
            \frac{N}{|\minibatch|}  
            \sum_{i\in \minibatch}
            x_i \frac{E'_i}{E_i}
        }
    },
    \label{eq:opt-m}
    \\
    C 
    & \leftarrow 
    C 
    + 
    \rho \del{
        -\text{triu}\del{C \Sigma^{-1}}
        +\del{C \circ I}^{-1}
        +
        \text{triu}\del{
            C 
            \expec{q(w_1)q(w_2)}{
                \frac{N}{|\minibatch|}  
                \sum_{i\in \minibatch}
                x_i x_i^T \left( E''_i-\frac{(E'_i)^2}{E_i} \right)
            }
        }
    },
    \label{eq:opt-C}
    \\
    \sigma 
    & \leftarrow 
    \sigma 
    + 
    \rho \del{
        -\del{
            \frac{2\sigma}{\tau^2}
            -\frac{2}{\sigma} 
        }
        +\frac{1}{\sigma}
        \expec{q(w_1)q(w_2)}{
            \frac{N}{|\minibatch|}  
            \sum_{i\in \minibatch}
            \frac{w_2^2}{2} \frac{E''_i}{E_i}
        }
    },
    \label{eq:opt-sigma}
    \\
    \cov^{-1}m 
    & \leftarrow 
    \del{1-\rho} \cov^{-1}m 
    + 
    \rho \del{
        \Sigma^{-1}\mu
        +
        \expec{q(w_1)q(w_2)}{
            \frac{N}{|\minibatch|}  
            \sum_{i\in \minibatch}
            x_i \frac{E'_i}{E_i}
            -
            x_i x_i^T m \left( E''_i - \frac{(E'_i)^2}{E_i} \right)
        }
    },
    \label{eq:opt-nat-m}
    \\
    \cov^{-1}
    & \leftarrow 
    (1-\rho) \cov^{-1} 
    + 
    \rho \del{
        \Sigma^{-1}
        -\expec{q(w_1)q(w_2)}{
            \frac{N}{|\minibatch|}  
            \sum_{i\in \minibatch}
            x_i x_i^T \left( E''_i-\frac{(E'_i)^2}{E_i} \right)
        }
    },
    \label{eq:opt-nat-V}
    \\
    -\frac{1}{2\sigma^2}
    & \leftarrow 
    \del{1-\rho} \del{-\frac{1}{2\sigma^2}}
    + 
    \rho \del{
        -\frac{1}{2\tau^2}
        +\frac{1}{4\sigma^2}
        \expec{q(w_1)q(w_2)}{
            \frac{N}{|\minibatch|}  
            \sum_{i\in \minibatch}
            \frac{w_2^2}{2} \frac{E''_i}{E_i}
        }
    }
    \label{eq:opt-nat-sigma}
    ,
\end{align}
where $\minibatch$ is a subset of indices.
Here, triu$(\cdot)$ is a mask that zeros the lower-left portion of the input matrix (below the diagonal), and $\circ$ denotes element-wise product.

The updates are obtained through sampling to replace the expectations over $w_1, w_2, f_i$.
Note that the innermost expressions include $1/E_i$ and $(E'_i)^2$ whose Monte Carlo estimates are biased. 
The implementation uses a relatively large number of samples (100) for $f_i$ to mitigate against this bias.

\paragraph{Concrete Algorithms for GME}
S-DSVI utilizes (\ref{eq:opt-m}),(\ref{eq:opt-C}),and (\ref{eq:opt-sigma}).
MC-SSVI utilizes (\ref{eq:opt-nat-m}),(\ref{eq:opt-nat-V}),and (\ref{eq:opt-nat-sigma}).
H-MC-SSVI utilizes (\ref{eq:opt-m}),(\ref{eq:opt-nat-V}),and (\ref{eq:opt-nat-sigma}).
Additional update equations for the sub-optimal VLB (\ref{eq:vlbT}) and mean-field approximation are provided in \ref{app:subsec-gme-subopt} and \ref{app:subsec-gme-mf}.

\subsection{\alg{} for sGP using the VLB (\ref{eq:vlbT})}

Following \cite{Titsias2009} the current standard variational solution for sGP uses
the VLB (\ref{eq:vlb}). However, (\ref{eq:vlb})
is not optimal for sparse GP. As we show below, the solution for the optimal VLB (\ref{eq:vlbT}) can be developed. 
Unfortunately, the solution requires cubic time so it does not immediately provide a fast and improved solution for sGP. 
We therefore propose two variants of the VLB which can be computed efficiently, and which are motivated by theoretical analysis of variational approximations \cite{ShethKh2017} where they are shown to be regularized loss minimization algorithms. 
Our experiments show that the optimal VLB and the two variants all improve over the current standard solution.

Since the optimal VLB requires cubic time, in this section we develop the equations for the conjugate case where its solution has a closed form and does not require further costly iterative optimization. The general case where $p(y_i|f_i)$ is not Gaussian can be developed along similar lines except that we need Monte Carlo estimates of the gradients. The development here suffices to show the potential to improve over the accepted best practice \cite{Titsias2009} even in the conjugate case. 

In sGP $w$ captures the pseudo inputs, $f$ are the local site potentials, and  $y$ are the observations. 
Following standard practice, we assume a zero-mean prior, and denote the covariance matrix between two sets of variables $a$ and $b$ (w.r.t.\ some specified kernel function) as $K_{ab}$.
The loss term in (\ref{eq:vlbT}) is equal to 
$E_{q(w)} \left[  \log E_{p(f|w)} [ p(y|f) ] \right] = E_{q(w)} \left[  \log p(y|w) \right]$.
In the conjugate case $\log p(y|w)$ can be calculated in closed form as $\log {\cal N}(y|K_{fw}K_{ww}^{-1}w,K_{ff}-Q_{ff}+\sigma^2 I)$
where 
$Q_{ab}=K_{aw}K_{ww}^{-1}K_{wb}$.
To optimize the VLB we need derivatives w.r.t. parameters which can be obtained by applying
the multivariate identities provided by \citep{Rezende2014} which hold when integrating a smooth and integrable real-valued function, $\xi(w)$ and where $q(w) ={\cal N}(w|m,\cov)$:
\begin{align}
    \nabla_{m_i} \left[ \int q(w) \xi(w) dw \right] & = \int q(w) \nabla_{w_i} \left[\xi(w)\right]  dw \label{eq:mvn_iden_m} \\
    \nabla_{\cov_{ij}} \left[ \int q(w) \xi(w) dw  \right] & = \frac{1}{2} \int q(w) \nabla^2_{w_i,w_j} \left[\xi(w)\right]  dw. \label{eq:mvn_iden_V}
\end{align}

We therefore need the expectations of derivatives of $\log p(y|w)$ w.r.t.\ $w$.
The derivatives are
$\frac{\partial}{\partial w} \log p(y|w)=-Aw+b$
and $\nabla^2_w \log p(y|w)=-A$ where
$A=K_{ww}^{-1}K_{wf} (K_{ff}-Q_{ff}+\sigma^2 I)^{-1} K_{fw}K_{ww}^{-1}$
and
$b= K_{ww}^{-1}K_{wf} (K_{ff}-Q_{ff}+\sigma^2 I)^{-1}y$.
Combining this with the derivatives for the KL term (with $\mu=0$) gives %
\begin{equation}
\frac{\partial \mbox{VLB}}{\partial \cov} = \frac{1}{2}(\cov^{-1}-K_{ww}^{-1}-A)
\nonumber
\end{equation}

\begin{equation}
\frac{\partial \mbox{VLB}}{\partial m} = -K_{ww}^{-1}m -Am +b
\nonumber
\end{equation}
and equating to zero 
yields the closed form solution:
$\cov=(K_{ww}^{-1}+A)^{-1}$ and $m=\cov b$. As mentioned above the matrix inversion in computing $A$ is not efficient.

We propose two variants as follows. The first variant {\em V1} applies %
(\ref{eq:vlbi}) to sGP even though 
sGP is not in 2L-LVMi. Thus, the bound is not valid. 
In this case the loss term is $\sum_i  E_{q(w)} \left[  \log E_{p(f_i|w)} [ p(y_i|f_i) ] \right] = \sum_i E_{q(w)} \left[  \log p(y_i|w) \right] $,
where 
$p(y_i|w)={\cal N}(y_i|K_{iw}K_{ww}^{-1}w,K_{ii}-Q_{ii}+\sigma^2 )= {\cal N}(y_i|h_i^Tw,c_i)$ with $h_i=K_{ww}^{-1}K_{wi}$, $c_i=K_{ii}-Q_{ii}+\sigma^2$
and $i$ refers to an individual component of $f$.
The solution can be derived following the same steps as in the optimal model with
$A=\sum_i\frac{1}{c_i} h_i h_i^T$ and $b=\sum_i \frac{y_i}{c_i} h_i$.

The second variant {\em V2} replaces the loss term of the VLB with 
$\sum_i  \log E_{q(w)} \left[  E_{p(f_i|w)} [ p(y_i|f_i) ] \right] = \sum_i  \log E_{q(w)} \left[   p(y_i|w)  \right]$.
Now $p(y_i|w)$ is as in the previous paragraph and 
$\log E_{q(w)} \left[   p(y_i|w)  \right]=
\mbox{const}-\frac{1}{2}\log(c_i+h_i^T \cov h_i)-\frac{1}{2}\frac{(y_i-h_i^T m)^2}{c_i+h_i^T \cov h_i}$.
The derivatives of the loss term w.r.t.\ $m,\cov$ are respectively
$d=\sum_i\frac{y_i-h_i^T m}{c_i+h_i^T \cov h_i}h_i$ 
and 
$-\frac{1}{2}A=-\frac{1}{2}\left[ \sum_i\frac{(c_i+h_i^T \cov h_i)-(y_i-h_i^T m)^2}{(c_i+h_i^T \cov h_i)^2}h_i h_i^T \right]$.
The derivatives are given by $\frac{\partial \mbox{VLB}}{\partial \cov}$ above and $\frac{\partial \mbox{VLB}}{\partial m}=-K_{ww}^{-1}m + d$.
The optimal solution can be found with coordinate ascent.

\emph{V1, V2} are not lower bounds on the marginal likelihood, but the objective and its derivatives can be computed efficiently.  
This yield updates with a structure similar to the solution of (\ref{eq:vlbT}) (closed-form in the case of \emph{V1}, coordinate ascent needed for \emph{V2}).

\subsection{\alg{} for PMF using  the VLB (\ref{eq:vlb})}

Probabilistic matrix factorization (PMF) \citep{SalakhutdinovM07,SalakhutdinovM08a} is a generative model for 
a sparsely observed data matrix $Y$ of dimension $N_U\times N_V$ as follows.
First, matrices $U$ (dimension $K\times N_U$) and $V$ (dimension $K\times N_V$) are drawn by drawing each column of $U$ and $V$ independently from a Gaussian prior, ${\cal N}(\mu,\Sigma)$. %
Then each observed entry $y_{i,j}$ is drawn independently according to $p(y_{i,j}|f_{i,j}=u_i^T v_j)$ where 
$u_i$ and $v_j$ are columns of the corresponding matrices and $p(\cdot)$
is any individual likelihood function for the observations.

Variational solutions for PMF have been extensively studied for the conjugate case \citep{LimTeh2007,RaikoIK07} as well as for the logistic, Poisson and ordinal likelihood functions \citep[e.g.,][]{SeegerB12,KoenigsteinP13,KoK14,KhanKS14,Hernandez-LobatoHG14,PaquetTW12}. But, most of these solutions are specific to a single likelihood, often making local variational bounds or using a (fully factorized) mean field variational distribution for $q()$. 

Note that since PMF is in \lvm{}d we can apply (\ref{eq:vlb}) and the methodology above directly
where $w$ captures the concatenation of all columns of $U$ and $V$.
However, for computational reasons we proceed with stronger assumptions on $q()$ that yield a more efficient algorithms. In particular we assume
 a variational distribution $q()$ which factorizes over the columns $q(U,V)=(\prod_{i} q(u_{i})) (\prod_{j} q(v_{j}))$ where $q(u_i)={\cal N}(m_{u_i},S_{u_i})$ and $q(v_j)={\cal N}(m_{v_j},S_{v_j})$ are Gaussian.\footnote{
The work of \cite{LimTeh2007} has shown that it is sufficient to assume $q(U,V)=q(U)q(V)$ and that the decomposition over columns arises from the form of the optimal solution.
 This assumption is much weaker than the complete mean field factorization over all entries of $U,V$ as used, 
for example, in \cite{Hernandez-LobatoHG14}.
}
Now, 
using $q(U,V,f)=q(U,V)p(f|U,V)$
one arrives at a VLB where the contributions of the different columns are separated out in the KL component. Using $O$ to  denote the set of observed entries, this gives:
\begin{eqnarray}
\label{eq:pmfVLB}
\log p(\mbox{data}) & \geq & 
\sum_{(i,j)\in O} 
E_{q(f_{i,j})} [\log p(y_{i,j}|f_{i,j})]  \ \ \ \ \ \\
& - & \sum_i d_{KL}(q(u_i) \| {p}(u_i))  %
 -\sum_j d_{KL}(q(v_j) \| {p}(v_j)).  \nonumber
\end{eqnarray}
It is clear that the general fixed-point update of \lvm{} from (\ref{eq:expfamilyUpdate}) is applicable for the VLB of PMF given in (\ref{eq:pmfVLB}) for each column of $U,V$ separately
so that the same algorithm is applicable with the factored variational distribution as well.

We next analyze the derivatives and show that this update can be done efficiently.
More specifically,
in the following we develop the terms $D_i$ in (\ref{eq:FP})  to show that a simple fixed-point update can be obtained leading to an efficient algorithm for non-conjugate PMF.

To achieve this 
we need to analyze the form of $E_{q(f_{i,j})} [\log p(y_{i,j}|f_{i,j})]$. Denoting this term by $A_{i,j}$ we have that
$A_{i,j}= \int_{u_i} {\cal N}(u_i|m_{u_i},S_{u_i}) B_{u_i} d u_i$ where
$B_{u_i} = \int_{v_j} {\cal N}(v_j|m_{v_j},S_{v_j}) \log p(y_{i,j}|f_{i,j}=u_i^T  v_j) d v_j.$
Note that $A_{i,j}$ is a function of the observation index, but $B_{u_i}$ is conditioned on a specific value of $u_i$.
With this conditioning, we can integrate out $v_j$ to get the marginal distribution over $f_{i,j}$ as
$f_{i,j}|u_i\sim {\cal N}(\alpha_{i,j},\beta_{i,j})$ where $\alpha_{i,j}=u_i^T m_{v_j}$ and $\beta_{i,j}= u_i^T S_{v_j} u_i$. Therefore, $B_{u_i}$ can be re-expressed as
$B_{u_i} = \int_{f_{i,j}} {\cal N}(f_{i,j}|\alpha_{i,j},\beta_{i,j}) \log p(y_{i,j}|f_{i,j}) d f_{i,j}.$
Using this notation it is clear that $B_{u_i}$ is identical to the expectation term in GLM 
where $u_i$ serves as the example descriptor $h_i$. 
We therefore have the following:
\begin{eqnarray}
\label{eq:pmf_expec_start}
\frac{\partial A_{i,j}}{\partial S_{v_j}} & = & 
\int_{u_i} {\cal N}(u_i|m_{u_i},S_{u_i}) 
\frac{\partial B_{u_i}}{\partial S_{v_j}} 
d u_i
\\
\frac{\partial B_{u_i}}{\partial S_{v_j}} & = & 
\frac{\partial B_{u_i}}{\partial \beta_{i,j}}  \frac{\partial \beta_{i,j}}{\partial S_{v_j}} %
=
\frac{\partial B_{u_i}}{\partial \beta_{i,j}}  u_i u_i^T %
=
\gamma_{i,j}  u_i u_i^T 
\end{eqnarray}
where $\gamma_{i,j} = \frac{\partial B_{u_i}}{\partial \beta_{i,j}} $ is identical to the corresponding $\gamma$ term from  equation (\ref{eq:rhoi-first}) of the previous section. %
Let
\begin{equation}
\label{eq:cij}
D_{i,j}=\frac{\partial A_{i,j}}{\partial S_{v_j}} = 
\int_{u_i} {\cal N}(u_i|m_{u_i},S_{u_i}) 
\gamma_{i,j}
u_i u_i^T
d u_i
\end{equation}
from which we can observe that $D_{i,j}$ is positive semi-definite for log-concave likelihoods (since $\gamma_{i,j}\leq 0$ is implied by (\ref{eq:rhoi-first})).
Using the derivatives for the KL terms in (\ref{eq:pmfVLB}), we get the generic fixed-point update for the covariance
\begin{equation}
S_{v_j} = (\Sigma^{-1} - 2\sum_{i\in O_j} D_{i,j})^{-1}
\end{equation}
where $O_j=\{i\mid (i,j)\in O\}$.
The derivation and update for $S_{u_i}$ are symmetric and can be obtained in the same manner. 

The equations above hold for any likelihood function. 
It is interesting to recall 
the conjugate case with Gaussian likelihood $p(y|f)={\cal N}(y|\mu,\sigma^2)$ where $\mu=f$.
In this case 
$\frac{\partial^2}{\partial f_i^2}  \log p(y_i|f_i) = -\frac{1}{\sigma^2}$ does not depend on $f$, implying that
$\gamma_{i,j}=-\frac{1}{2\sigma^2}$ is independent of the variational parameters and 
$D_{i,j}=\frac{\partial A_{i,j}}{\partial S_{v_j}} = -\frac{1}{2\sigma^2} (m_{u_i} m_{u_i}^T + S_{u_i})$.
We therefore get the fixed-point update $S_{v_j} = (\Sigma^{-1} + \sum_i \frac{1}{\sigma^2} (m_{u_i} m_{u_i}^T + S_{u_i}))^{-1}$ which is identical to the closed-form update of \cite{LimTeh2007}. 
For other local likelihoods we can use quadrature and possibly reduce run time through tabulation as in \cite{Challis2013}. However, as we show next, we can obtain a generic algorithm by using Monte Carlo sampling.

In particular, as in previous work,
we can approximate $D_{i,j}$ as follows:
sample $u_i\sim \mathcal{N}(m_{u_i},S_{u_i})~k_1$ times; then, for each sample $u_i^a$, sample $f_{i,j}^a\sim \mathcal{N}(m_{v_j}^T u_i^a,(u_i^a)^T S_{v_j} u_i^a)~k_2$ times; finally, calculate the corresponding average. This yields our estimate 
\begin{eqnarray}
\hat{D}_{i,j} & = &
\frac{1}{2}\frac{1}{k_1 k_2}\sum_{a=1}^{k_1} 
u_i^a (u_i^a)^T
\sum_{b=1}^{k_2}
\left[ \frac{\partial^2}{\partial f_{i,j}^2}  \log p(y_{i,j}|f^{a,b}_{i,j})\right].
\label{eq:pmf_Cij}
\end{eqnarray}
Note, that since we work with the expectations and derivatives in (\ref{eq:pmf_expec_start}-\ref{eq:cij}) directly, the parameters of the sampling distribution used to calculate (\ref{eq:pmf_Cij}) do not interact with the derivatives.
Therefore, the derivative estimates are stable and we do not need to resort to re-parameterization or variance reduction techniques as developed in \cite{Kingma2014, Ranganath2014}.

We next discuss the use of mini-batches for updates. Previous work \cite{Hernandez-LobatoHG14} has used sampling from the full dataset and updating corresponding columns of the matrices that are affected by the sampled data. 
However, the decomposition in $q()$ and its updates suggests a more effective structure.
More specifically, our algorithm uses an improved scheme iterating over columns of $U$,$V$ in round robin manner and updating the hyperparameters after each column. This balances the cost of global updates and parameters without the need for additional noise from sampling over the entire dataset. Preliminary experiments (omitted in the paper) showed that this scheme performs better than the standard sampling approach. Considering single columns, in most PMF problems, the data associated with each column is already sparse so that no sampling is needed. In case of full or large matrices we sub-sample  directly the data associated with the specific item that is being updated and not from the entire dataset. 

Therefore, to update $S_{v_j}$, we sample a mini-batch of observation pairs $\minibatch\subseteq O_j$ to obtain a stochastic gradient, and combine this 
 with a sample estimate of the gradients $\hat{D}_{i,j}$ as described above. 
 This yields the following \alg{} update for the covariance, which is applicable to any likelihood function, 
\begin{equation}
\label{eq:PMF-SVI-update-V}
S_{v_j}^{-1} = (1-\rho) S_{v_j}^{-1} + \rho (\Sigma^{-1} - 2\frac{|O_j|}{|\minibatch|}\sum_{i\in \minibatch} \hat{D}_{i,j}).
\end{equation}

\paragraph{Concrete Algorithms for PMF:}
As in the case of previous models we can define multiple algorithms.
All these algorithms iterate in a round robin manner over columns, subsampling the data associated with the current column if needed, and updating the corresponding parameters of the current column. Hyperparameters are updated after each column.

The \alg{} algorithm uses 
 (\ref{eq:PMF-SVI-update-V}) and (\ref{eq:PMF-mcssvi-update-m}) to update the variational parameters. 
The \hsvi{} algorithm uses 
 (\ref{eq:PMF-SVI-update-V}) and (\ref{eq:PMF-dsvi-update-mean}) to update the variational parameters. 
The \dsvi{} algorithm uses 
(\ref{eq:PMF-dsvi-update-C}) and (\ref{eq:PMF-dsvi-update-mean}) to update the Cholesky factor and mean parameters respectively.
The update for hyperparameters $\sigma_V^2$, $\sigma_U^2$ is given by 
(\ref{eq:PMF-hyp-update-sigma}).

\subsection{\alg{} for CTM using  the VLB (\ref{eq:vlbi})}
The correlated topic model (CTM) of \cite{Blei2006} is an extension of LDA that models correlations between document-level topic proportions.
For consistency with previous work,
in this section we follow the notation from \cite{Blei2006}  where $\theta,\eta$ are used to denote different quantities from the ones above. In particular,
 $\theta_d$ denotes the document-level topic proportions. 
For a document $d$, the generative model for CTM first draws $\eta \sim \mathcal{N}(\mu,\Sigma)$, $\eta\in\mathbb{R}^{K-1}$  where $\{\mu,\Sigma\}$ are model parameters, and then maps this vector to the $K$-simplex with the logistic transformation, $\theta_d=h(\eta_d)$ where
\begin{equation}
h_k(\eta) = 
\begin{cases}
\frac{\exp(\eta_k)}{1+\sum_{l=1}^{K-1} \exp(\eta_l)}, & \text{if}\ k<K \\
\frac{1}{1+\sum_{l=1}^{K-1} \exp(\eta_l)}, & \text{otherwise.}
\end{cases}
\label{eq:log_transform}
\end{equation}
For each position $n$ in the document, the latent topic variable, $z_{dn}$, is drawn from Discrete$(\theta_d)$, and the word $w_{dn}$ is drawn from a Discrete$(\beta_{\cdot, z_{dn}})$ where $\beta$ denotes the topics
and is treated as a parameter of the model. 

CTM fits within the \lvm{} where $\mu,\Sigma,\beta$ are parameters, the concatenation of $\eta_d$ corresponds to $w$,
examples are documents indexed by $d$, $z_d$ corresponds to $f_i$, and $w_d$ corresponds to $y_i$. In this formulation we have that $p(f|w)p(y|f)= \prod_i p(f_i|w)p(y_i|f_i)$; that is, CTM is in \lvm{}i and we can use the optimal bound of (\ref{eq:vlbi}).

Previous work \cite{Blei2006} used a mean field factorization where all components of $\eta$ and $z$ are independent. 
As in the case of PMF we can utilize the generic algorithm directly but use additional factorization for efficiency. In particular, 
we use the structured variational distribution, but factor the document-level topic vectors over documents
$q(\eta,z)=\prod_d q(\eta_d, \{z_d\}) = \prod_d q(\eta_d) q(z_{d}\vert \eta_d)$. 
Following the same steps as above we obtain the VLB
\begin{equation}
    \log p(\text{data}) \ge 
    - \sum_{d=1}^D d_{\text{KL}}(q(\eta_d)\Vert p(\eta_d))
+
    \sum_{d=1}^D  
    E_{q(\eta_{d})} [ \log  p(w_{d}\vert \eta_{d}) ]    
    \label{eq:ctm_vlba}
\end{equation}
which factors over documents.
Now, owing to the structure in CTM we obtain a further simplification:
\begin{eqnarray*}
p(w_d|\eta_d)
& = & \sum_{z_d} p(z_d|\eta_d) p(w_d|z_d,\beta) \nonumber \\
& = & \sum_{z_d} \prod_n p(z_{dn} |\eta_d) p(w_{dn} |z_{dn},\beta) \nonumber \\
& = & \prod_n \sum_{z_{dn}} p(z_{dn} |\eta_d) p(w_{dn} |z_{dn},\beta) \nonumber \\
& = & \prod_n \sum_k h_k(\eta_d) \beta_{k,w_{dn}} \nonumber 
\end{eqnarray*}
where $k$ is a topic index.
We therefore have the bound
\begin{equation}
    \log p(\text{data}) \ge 
    - \sum_{d=1}^D d_{\text{KL}}(q(\eta_d)\Vert p(\eta_d))
+
    \sum_{d=1}^D  \sum_{n=1}^{N_d}
    E_{q(\eta_{d})} [ \log  (\sum_k h_k(\eta_d) \beta_{k,w_{dn}} ) ].    
    \label{eq:ctm_vlb}
\end{equation}
We note this bound appears in \cite{Khanthesis}, but, to our knowledge, the optimal VLB has not been tested before.

To apply our approach we need to calculate Monte Carlo estimates of the gradients of the terms in these sums. 
The derivative  with respect to the variational parameters $\{m_d,\cov_d\}$ can be taken directly by differentiation of $q(\eta_d)$ inside the integral. 
It has been noted that using Monte Carlo sampling to directly estimate derivatives of this kind can result in high variance estimates and several schemes have been proposed to reduce the variance \citep{Kingma2014, Titsias2014,Rezende2014,Ranganath2014,Kingma2015}.
However, as in the case of PMF, for CTM we can use direct evaluation of the expectations to obtain a simple sampling scheme.
In particular,
since $\eta_d$ is
Gaussian, we can once again use
the multivariate identities (\ref{eq:mvn_iden_m}, \ref{eq:mvn_iden_V}) 
with $\xi_n(\eta_d)=\log  (\sum_k h_k(\eta_d) \beta_{k,w_{dn}} )$. 
For any $x\in\mathbb{R}^K$, let $\tilde{x}\in\mathbb{R}^{K-1}$ denote the first $K-1$ elements of $x$.
Then, we have that
\begin{eqnarray*}
\nabla_{\eta_{d}} \left[h_k(\eta_d)\right] & = & h_k(\eta_d)(\tilde{e}^{(k)}-\tilde{h}(\eta_d)) \\
\nabla^2_{\eta_{d}} \left[h_k(\eta_d)\right] & = & h_k(\eta_d) \left[ (\tilde{e}^{(k)}-\tilde{h}(\eta_d)) (\tilde{e}^{(k)}-\tilde{h}(\eta_d))^T - \text{diag}(\tilde{h}(\eta_d)) + \tilde{h}(\eta_d) \tilde{h}(\eta_d)^T \right] \\
\nabla_{\eta_{d}} \left[\xi_n(\eta_d)\right] & = & \exp(-\xi_n(\eta_d)) \sum_k \beta_{k,w_{dn}} \nabla_{\eta_d} h_k(\eta_d) \\
\nabla^2_{\eta_{d}} \left[\xi_n(\eta_d)\right] & = & \exp(-\xi_n(\eta_d)) \left( \sum_k \beta_{k,w_{dn}} \nabla^2_{\eta_d} h_k(\eta_d) \right) - \left(\nabla_{\eta_d} \xi_n(\eta_d)\right)\left(\nabla_{\eta_d} \xi_n(\eta_d)\right)^T \\
\end{eqnarray*}
where $e^{(k)}\in\mathbb{R}^K$ denotes the standard Euclidean unit vector in the $k$-th coordinate.

With these, we can use (\ref{eq:mvn_iden_m}-\ref{eq:mvn_iden_V}) to obtain stable Monte Carlo estimates of the gradients.
Letting $\hat{D}_n(\cov_d)$ stand for the approximation of 
$\frac{\partial}{\partial \cov_d} E_{q(\eta_{d})} [\xi_n(\eta_d)]$, that is,
\begin{equation}
\hat{D}_n(\cov_d) = \frac{1}{2}\frac{1}{N_{MC}} \sum_{\ell=1}^{N_{MC}} \nabla^2_{\eta_d} [\xi_n(\eta_d^{(\ell)})], \quad \eta_d^{(\ell)}\sim \mathcal{N}(\eta_d\vert m_d, \cov_d).
\label{eq:Gn_approx}
\end{equation}
The fixed-point update rule for the variational covariance is given by
\begin{equation}
\label{eq:ctmUpdate}
    \cov_d^{-1} \leftarrow (1-\rho) \cov_d^{-1} + \rho \left(\Sigma^{-1} - 2\sum_{n=1}^{N_d} \hat{D}_n(\cov_d) \right).
\end{equation}
To prevent an update to an indefinite or negative definite covariance, $-2\sum_{n=1}^{N_d} \frac{\partial}{\partial \cov_d} \hat{D}_n(\cov_d)$ is projected to the positive semi-definite cone, $S^{++}$, prior to the update. 
\begin{equation}
\label{eq:CTM-SVI-update-V}
    \cov_d^{-1} \leftarrow (1-\rho) \cov_d^{-1} + \rho \left(\Sigma^{-1} + \Pi_{S^{++}}(- 2\sum_{n=1}^{N_d} \hat{D}_n(\cov_d)) \right).
\end{equation}

Unlike the previous models, updates for hyperparameters require some further details. 
The updates for the global prior are easily derived in closed-form and are given by
\begin{equation}
\label{eq:CTM-hyp-update-mu}
\hat{\mu}  = \frac{1}{D} \sum_{d=1}^D m_d,
\end{equation} and 
\begin{equation}
\label{eq:CTM-hyp-update-Sigma}
\hat{\Sigma}  = \frac{1}{D} \sum_{d=1}^D \cov_d + (\hat{\mu}-m_d)(\hat{\mu}-m_d)^T.
\end{equation}
The update for $\beta$ requires numerical optimization with Monte Carlo sampling of the derivative.
To allow for unconstrained optimization, we convert each topic, $\beta_{k,\cdot}$ to its minimum representation, $\alpha_{k,\cdot}$, of $V-1$ elements (where the logistic transformation, (\ref{eq:log_transform}), recovers the topics).
Letting $\gamma_{dkw}=\int q(\eta_d) \frac{\exp(\eta_{dk})}{\sum_{l=1}^K \beta_{lw} \exp(\eta_{dl})} d\eta_d$, the derivative of the VLB w.r.t.\ $\alpha_{ku}$ for $1 \le k \le K$ and $1 \le u < V$ is given by
\[ \frac{\partial\text{VLB}}{\partial \alpha_{ku}} = -\beta_{ku} \sum_d \left( -c_{du} \gamma_{dku} + \sum_{v=1}^V c_{dv} \gamma_{dkv} \beta_{kv} \right) \]
where $c_{dw}$ denotes the count of the $w$-th vocabulary word in the $d$-th document.
The Monte Carlo approximation of this derivative is given by
\[ \frac{\partial\text{VLB}}{\partial \alpha_{ku}}\approx -\beta_{ku} \sum_d \frac{1}{N_{MC}} \sum_{\ell=1}^{N_{MC}} \left( -c_{du} \gamma_{dku}^{(\ell)} + \sum_{v=1}^V c_{dv} \gamma_{dkv}^{(\ell)} \beta_{kv} \right), \quad \eta_d^{(\ell)}\sim \mathcal{N}(\eta_d\vert m_d, \cov_d) \]
where $\gamma_{dkw}^{(\ell)} = \frac{\exp(\eta_{dk}^{(\ell)})}{\sum_{l=1}^K \beta_{lw} \exp(\eta_{dl}^{(\ell)})}$.
The computation of this derivative is expensive since it requires a sum over all documents. 
To avoid this, and avoid high storage requirements say for storing the values of $\gamma_{dkv}$, we propose to use stochastic gradients for updates of $\beta$ as well.
In particular, we pick one document $d$ (the one just updated) and approximate the gradient by
\begin{equation}
\label{eq:CTM-hyp-update-beta}
\frac{\partial\text{VLB}}{\partial \alpha_{ku}}\approx -\beta_{ku} \frac{D}{N_{MC}} \sum_{\ell=1}^{N_{MC}} \left( -c_{du} \gamma_{dku}^{(\ell)} + \sum_{v=1}^V c_{dv} \gamma_{dkv}^{(\ell)} \beta_{kv} \right), \quad \eta_d^{(\ell)}\sim \mathcal{N}(\eta_d\vert m_d, \cov_d) 
\end{equation}
where the dataset has $D$ documents. 

\paragraph{Concrete Algorithms for CTM:}
As in the previous models we can define multiple algorithms.
All the algorithms iterate in a round robin manner over documents (we do not subsample words in a document although that can be added if documents are very long) and update the corresponding variational parameters of the current document.
Model parameters are updated after each document.

The \hsvi{} algorithm uses 
 (\ref{eq:CTM-SVI-update-V}) and (\ref{eq:CTM-dsvi-update-mean}) to update the variational parameters. 
The \dsvi{} algorithm uses 
(\ref{eq:CTM-dsvi-update-C}) and (\ref{eq:CTM-dsvi-update-mean}) to update the Cholesky factor and mean parameters respectively.
The updates for hyperparameters  $\mu, \Sigma$ are given by 
(\ref{eq:CTM-hyp-update-mu}),
(\ref{eq:CTM-hyp-update-Sigma}), and for $\beta$ we use gradient ascent using 
(\ref{eq:CTM-hyp-update-beta}).

\section{Experiments}
\label{sec:experiments}

In this section, we demonstrate the applicability of the \alg{} approach in the LGM sub-family, presenting results for \alg, \hsvi{} and \dsvi{}.
The experiments illustrate
(i)
the advantage of the optimal structured approximation over the mean field approximation and over the simple structured approximation,
(ii) the advantage of natural gradients over standard gradients,
and
(iii) the improved performance of  \hsvi{} over ``pure" variants in latent Gaussian models.

To demonstrate general applicability we use the same 
optimization-related settings 
for all the experiments as follows.
Mean and covariance natural gradient updates use a $\frac{1}{t}$ schedule. Other updates use ADAGRAD \cite{Duchi2011short} and 1.0 global learning rate. 10 Monte Carlo samples are used in each expectation. In GME, 100 samples are used for the inner expectations. Data in supervised experiments are z-score normalized on the training set.
Unless otherwise noted, datasets were split 80\%/20\% for train and test. All NLL values are averages across the test set.
Finally, all variational- and hyper- parameters were initialized to the same values across algorithms
($m=0,S=10I,\sigma=1,\mu=0,\Sigma=I,\tau=5$). 
The GLM, GME, sGP, and CTM algorithms were implemented in Matlab (2014a) and the PMF algorithm was implemented in Matlab and C.
The implementation of mean field variational inference for CTM \cite{Blei2006} is written in C.\footnote{\url{http://www.cs.princeton.edu/~blei/ctm-c/ctm-dist.tgz}}
In all experimental results, lower values indicate better performance.
Additional experimental details and results are given in the appendix.

\subsection{GLM with logistic likelihood}
The first experiment provides a simple demonstration of \alg{} in a Bayesian GLM with logistic likelihood. 
We use the {\em epsilon} dataset
which has $100,000$ samples and $2000$ features.\footnote{\url{http://www.csie.ntu.edu.tw/~cjlin/libsvmtools/datasets/binary/epsilon_normalized.t.bz2}}
We compare \alg, \hsvi, and \dsvi{} all using mini-batch sampling with 2000 samples. In addition we evaluate a batch version of \hsvi{} without data subsampling (this algorithm still uses stochastic gradients with a $\frac{1}{t}$ step schedule due to Monte Carlo sampling). 
The prior over the global latent variables was fixed to $\mathcal{N}(0,I)$ in this experiment.

The results are presented in Figure \ref{fig:glmplot}.
We can observe that both \alg{} and \hsvi{} converge substantially faster than \dsvi{}  showing the advantage of natural gradients.
The \alg{} algorithm suffers from an initial instability that delays convergence, which is a pattern that also occurs  in other models and agrees with observations in \cite{ShethKh2016} on the fixed-point variant of this algorithm.
Comparing \hsvi{} to its batch version we see that data subsampling provides significant improvement in performance. 
The batch version also exhibits non-monotonic behavior similar to the one reported for fixed points \cite{ShethKh2016}. 
These results support points (ii-iii) above.

\putGLMPlot{fig:glmplot}{GLM with logistic likelihood experiment comparing performance of proposed algorithm, \dsvi{}, the batch version of the proposed algorithm, and full \alg{}.}

\subsection{GME with Poisson likelihood}
\putGMEPlot
We use the {\em epsilon} dataset again %
to compare \alg, \hsvi, and \dsvi{} trained with the optimal VLB (\ref{eq:vlb}) all using mini-batch sampling with 2000 samples. 
To generate labels, true parameters $(w_1,w_2)$ were sampled from the priors and count data were generated from a Poisson with logistic link function. 
Figure \ref{fig:gme} a shows test negative log likelihood (NLL) as function of training time.
H-MC-SSVI has the best performance followed by MC-SSVI with S-DSVI taking long to converge.
For reference, the test NLL of a trained GLM model is also shown demonstrating that the GME model can produce better fits.
Also, we plot the best test NLLs obtained from training with the sub-optimal VLB (\ref{eq:vlb}) and that obtained from using a mean field factorization.
The use of the optimal VLB yields the best performance.
These results support points (i-iii) above.

\subsection{Sparse GP}

\putSGPPlotRef
Here we explore the use of the optimal bound (\ref{eq:vlbT}) and the variants \emph{V1, V2}.
for the
Gaussian likelihood\footnote{zero-mean, RBF kernel w/ hyperparameters found by grid search on train subset} with the \emph{cahousing} dataset\footnote{\url{http://www.csie.ntu.edu.tw/~cjlin/libsvmtools/datasets/regression/cadata}} (20,640 samples).
We split the data 67/33, fix a randomly selected inducing set of 206 points from train, and focus on performance as a function of training set size.
For each training set size, a random subset is selected to compute the suboptimal (\ref{eq:vlb}), \emph{V1}, and \emph{V2} solutions which are then evaluated by test NLL.
Additionally, the optimal bound (\ref{eq:vlbT}) and full GP solutions are computed up to 1000 samples.
This is repeated 50 times and we report averages of the difference from each method to the sub-optimal solution (where negative implies better performance).
Figure \ref{fig:sgpplotref} shows the results. 
We see that (1) all methods are better than the sub-optimal solution, (2) the performance gap between the methods and the sub-optimal solution shrinks with increasing train set size, (3) the optimal solution indeed improves upon the sub-optimal one, and (4) the heuristic variants have either competitive or better performance than the optimal solution.
These results support point (i) above.

\subsection{Generalized PMF}

\putPMFPlotsVLB{fig:pmfplotsvlb}{Monte Carlo-sampled negative VLB with respect to training time across several likelihood functions (from top to bottom: binary, count, ordinal-5) and artificial (left) and real (right) data sets. S-D denotes S-DSVI and H-MC denotes H-MC-SSVI.} 

\putPMFPlotsNLL{fig:pmfplotsnll}{Average NLL per observation on test with respect to training time across several likelihood functions (from top to bottom: binary, count, ordinal-5) and artificial (left) and real (right) data sets. S-D denotes S-DSVI and H-MC denotes H-MC-SSVI.} 

\putPMFPlotsErr{fig:pmfplotserr}{Average of various error metrics per observation on test with respect to training time across several likelihood functions (from top to bottom: binary, count, ordinal-5) and artificial (left) and real (right) data sets. S-D denotes S-DSVI and H-MC denotes H-MC-SSVI.}

In this section we evaluate the performance of \hsvi{} and \dsvi{} across multiple conditions.
To explore the generality of the algorithm we perform this evaluation with several likelihood functions. More specifically, we evaluate PMF with the following observation models: binary (implemented with logistic likelihood), count (implemented with Poisson likelihood and logistic link function), and 5 category ordinal (see \cite{Sheth2015} for definition of likelihood).
We used 6 datasets overall, where
for each likelihood we used one real dataset where the matrix is sparsely observed and one artificial dataset where the matrix is fully observed. 
The artificial datasets are fully observed $1000 \times 1000$ matrices and were created from the generative model.
The real datasets (\emph{movielens 1M} for binary and ordinal and \emph{lastfm} for count) are sparsely observed matrices of between $\approx100,000$ entries and $\approx1,000,000$ entries.\footnote{\url{http://grouplens.org/datasets/.} The binary dataset was formed from the ordinal by taking ratings 1 and 5.}
The latent dimensionality was fixed to $D=100$ for the runs, and an 80/20 split was used for train/test set size.
The ordinal likelihood parameters remain fixed during these experiments.

We emphasize that the same code is used for all these runs where the only difference in implementation across the models is the definition of the likelihood function $\log p(y_{i,j}|f_{i,j})$ and its local derivatives
$\frac{\partial}{\partial f_{i,j}}  \log p(y_{i,j}|f_{i,j})$
and
$\frac{\partial^2}{\partial f_{i,j}^2}  \log p(y_{i,j}|f_{i,j})$.

Secondly, to explore the effect of data subsampling we ran the algorithms with varying levels of data sub-sampling per column as well as no data sub-sampling (batch mode), and we show 
how their performance varies across these runs. 

We report multiple evaluation criteria including the training set objective of optimizing the VLB, and test set objectives represented by negative log likelihood and several error measures. 

The full set of results with respect to the VLB are shown in Figure \ref{fig:pmfplotsvlb}.
As can be seen, data sub-sampling improves performance in several cases, but the optimal batch size is dataset dependent.
This suggests that automatic size selection \cite{Hernandez-LobatoHG14} may be a useful addition to the algorithm.
It is also clear from Figure \ref{fig:pmfplotsvlb} that
\hsvi{} converges significantly faster than \dsvi{} across all conditions. In some cases the best batch size for \dsvi{} performs better than the worst batch size for \hsvi{} but otherwise we see a significant gap between the algorithms.

Evaluation of predictive negative log-likelihood (NLL) on a held-out test set are shown in Figure \ref{fig:pmfplotsnll}.
In the case of the binary likelihood (top row) and count likelihood (middle row) the results agree with those observed with respect to the VLB where sampling size is dataset dependent and \hsvi{} provides better performance. 
For the ordinal likelihood (bottom row) we observe some instability for both datasets with \dsvi{} generally outperforming \hsvi.

One hypothesis to explain the difference of NLL results from the VLB is that the posterior covariances produced by \hsvi{} are too narrow, i.e., over-confident, and that this adversely affects the negative log likelihood in the ordinal case.
To explore this further, we evaluated common error metrics for all likelihoods and datasets, again on the held-out test sets: zero-one error for binary, relative error\footnote{
Relative error excludes true zero counts for which relative error would be infinite. Results for zero counts are shown in the appendix.}
 for count, and absolute error for ordinal.
In all cases, predictive estimates are used.
The results are given in Figure \ref{fig:pmfplotserr} and they show that \hsvi{} is superior to \dsvi{} when focusing on predictive performance. They also show that small fluctuations in VLB can lead to large difference in predictions which we attribute to instability in working with the ordinal likelihood.
These results also support the hypothesis regarding narrow posterior in the ordinal case since the predictive means of \hsvi{} appear to be closer to the observed values than those of \dsvi{} whereas the relationship is reversed for NLL.

In summary, we observe that both \hsvi{} and \dsvi{} are generally applicable, that \hsvi{} converges faster and provides better predictive performance, and that mini-batch sampling is useful but that the optimal mini-batch size is dataset dependent.
These results support points (ii-iii) above.

\subsection{CTM}

\putCTMStructurePlot{fig:mcsvi-ctm-structure}{Comparison of different batch variational approximations for CTM (with diagonal covariances). On the left plot, we show training performance and on the right plot we show evaluation on test. Note, the mean field implementation was written in C, and the others in Matlab.}

\putCTMKPlot{fig:mcsvi-ctm-K}{Performance in CTM on \emph{nips} dataset as a function of latent dimensionality. H-MC-SSVI uses diagonal covariances and updates one document prior to updating hyperparameters. Both algorithms were run for 24 hours.} 

For CTM we can compare the optimal structured approximation to the mean field approximation which was used by previous work and to the simple structured form given in (\ref{eq:SimplestructuredQ}).
For mean field we use the implementation of \cite{Blei2006} which is a batch algorithm and which uses diagonal matrices for the variational covariances. 
Therefore, our first experiment runs batch versions of \hsvi{} and \dsvi{} using diagonal covariances. In addition, we run a version of \hsvi{} derived from the simple structured approximation (\ref{eq:SimplestructuredQ}) with the same setting. This setup allows differences due to the use of different variational approximations  to be isolated.

For the experiment we use  the \emph{nips} dataset \citep{UCI} 
with latent dimensionality (number of topics) of 50.
We use an 80/20 split of the data for training and test performance and report VLB and NLL on the corresponding portions. 
The NLL is computed on the test set, where the normalization is a document's NLL divided by the number of words in the document, and the average is the sum of this metric applied across the test set divided by the number of test documents.
It is well known that efficient calculation of NLL for topic models is challenging \cite{Wallach2009} and the problem is compounded for CTM because of non-conjugacy.\footnote{More specifically,  in order to use the solutions in \cite{Wallach2009} one would have to sample both variational parameters and individual topic assignments which makes the evaluation more costly and would require more time to converge.
}
We therefore resorted to using a simple but expensive evaluation, applying importance sampling with a very large number of samples. 
Previous work has used the posterior as a sampling distribution. To avoid the problem of narrow posteriors we inflate its covariance with an additive diagonal term. We have experimented with several such scheme and the results are consistent across these evaluations, with details given in the appendix. 
Here we report on using the posterior over $\eta_d$ as the sampling distribution with $10^{-1}$ diagonal covariance inflation, where we use $10^7$ samples to estimate each NLL value.

The results are shown in Figure \ref{fig:mcsvi-ctm-structure}. 
We note that the results are biased in favor of mean field because its implementation in C is faster.
The left plot shows results for the VLB.
The simple structured bound is significantly worse than all other variants. 
The mean field algorithm is faster at first but  
the optimal structured approximation with \hsvi{} catches up and provides a better VLB.
Similar results for the {\em enron} dataset are given in the appendix where \hsvi{} catches up and provides a better VLB.
The \dsvi{} algorithm is slower to converge and has not crossed the bound of mean field within the time of the experiment. 
The right plot of Figure \ref{fig:mcsvi-ctm-structure} shows results for NLL
with \hsvi{} clearly performing best 
with respect to evaluation on the test set.

As shown by \cite{Blei2006}, 
one of the advantages of CTM over LDA is that it can support a larger number of topics without overfitting. 
It is therefore important to verify that this advantage is maintained or improved with the structured approximation.
To explore this point we compare the \hsvi{} and mean field algorithms 
as a function of latent dimensionality. 
In this experiment, \hsvi{} updates just one document prior to updating hyperparameters, and uses diagonal covariances.
Both algorithms were run for 24 hours and the final parameters are used for the evaluation.
Average NLL values on the test set are shown in Figure \ref{fig:mcsvi-ctm-K}. 
We see that across the range of $K$ tested, our algorithm provides better performance than the mean field algorithm.
These results support points (i-iii) above.

\section{Conclusion}
\label{sec:conclusion}

The paper makes several contributions in the context of variational analysis for latent variable models.
Our overview of variational bounds showed that many previous models can be analyzed in the same framework and that the seemingly stronger variational approximation in LGMs is in fact optimal in some cases but leaves room for improvement in others such as sparse GP. Using this analysis and the connection to fixed point updates 
we have shown that that the SVI algorithm (using natural gradients) is applicable in \lvm{} whenever $p(w)$ and $q(w)$ have the same exponentially family form and the gradient terms can be efficiently computed or approximated. Our algorithm \alg{} is applicable in the entire family of \lvm{} and has a decomposable structure allowing data sub-sampling in two subfamilies identified in the paper.
This significantly weakens the condition for conjugate complete conditionals for SVI required in prior work. 
We have also applied \alg{} to develop effective algorithms for GME, PMF and CTM and proposed a novel variant \hsvi{} which combines standard and natural gradients and provides additional performance improvements.
In the cases of PMF and CTM, we used additional factorization conditions to yield an efficient algorithm.
In sparse GP, where the optimal solution was computationally expensive to calculate, we proposed two efficient variants which provide significantly improved performance over the current, widely-used variational solution.

A few interesting directions arise from the work in this paper.
One concerns the general applicability of \alg{}.
All the models in our experiments are in the sub-family of LGM, that is, they have Gaussian global variables.
Therefore, there is empirical evidence for the success of \alg{} in several conjugate models and in non-conjugate LGM.
It would be interesting to develop and investigate the empirical performance in non-conjugate latent variable models that do not fall within the LGM sub-family.

Another direction would be to generalize the results of this paper along two dimensions. As mentioned above,
the approach is generic and applies across models, and even instances for a family of models can be generic. For example, our final algorithm and implementation for PMF is applicable for any local likelihood function $p(y_{ij}|f_{ij})$. 
However, 
significant work is still required when applying \alg{} to analyze the gradients, to identify the computational structure for calculating the gradients, and to integrate on-line posterior inference with on-line parameter optimization. 
Some of this can be alleviated by ``black-box'' inference algorithms and the recently developed sampling schemes discussed above \citep{Kingma2014, Titsias2014,Rezende2014,Ranganath2014,Kingma2015}.
For example \cite{Ranganath2014} develops a black-box sampling based method for the two level model and \cite{KucukelbirTRGB16} develops an extension combining sampling with automatic differentiation for a large class of models, but both use the mean field approximation and standard gradients. Further work is required to handle the marginalization used in the structured approximation, and to 
balance on-line posterior inference with on-line parameter optimization automatically to obtain efficient implementations. 
In the same context, it would be interesting to generalize the algorithm to work on general graphical models, beyond the \lvm{} of this paper.
An elegant  approach is given by the 
variational message passing algorithm of \cite{GhahramaniB00,WinnB05} that provides a generic mean field approximation for graphical models within the exponential family having conjugate conditional node models, and some extensions to non-conjugate models in the same framework have been developed \cite{KnowlesM11,Wand14}.
Generalizing the \alg{} approach to be directly applicable in general non-conjugate graphical models, while maintaining a structured approximation and natural gradients, is an important challenge for future work.

\appendix
\section{Appendix}
\label{sec:appendix}

This appendix contains additional update equations for the LGM models considered in the paper, 
information on the application of the weaker structured approximation (\ref{eq:SimplestructuredQ}) to GME and CTM and their update rules, and additional experimental details and results.

\subsection{Additional update equations}

\paragraph{Equations for LGM:}
In LGM, the \dsvi{} update for the variational mean is given by
\begin{equation}
\label{eq:LGM-dsvi-update-mean}
    m \leftarrow  m + \rho \left(\Sigma^{-1} (\mu-m) + \frac{N}{\vert \minibatch\vert}\sum_{i\in \minibatch} \hat{\alpha}_i h_i\right)
\end{equation}
where $\hat{\alpha}_i$ is an estimate of 
$\frac{\partial}{\partial m_i}E_{\mathcal{N}(f_i\vert m_i,v_i)}[\log p(y_i\vert f_i)]
=
E_{{\cal N}(f_i|m_{i},v_{i})}[ \frac{\partial}{\partial f_i}  \log p(y_i|f_i)]
$.

The \dsvi{} update for the Cholesky factor of the variational covariance, $C$ where $\cov=C^T C$, is given by
\begin{equation}
\label{eq:LGM-dsvi-update-C}
    C \leftarrow  C + \rho~\text{triu}\left( (C\circ I)^{-1} - C\Sigma^{-1}  + 2C\frac{N}{\vert \minibatch\vert}\sum_{i\in \minibatch} \hat{\lambda}_i h_i h_i^T\right).
\end{equation}
Here, triu$(\cdot)$ is a mask that zeros the lower-left portion of the input matrix (below the diagonal), $\circ$ denotes element-wise product, and $\hat{\gamma}_i$ is an estimate of (\ref{eq:rhoi-first}).

\paragraph{Equations for PMF:}
In PMF, the \alg{} update for the variational mean of column $v_j$ is given by
\begin{equation}
    S_{v_j}^{-1} m_{v_j} 
    \leftarrow 
    \del{1-\rho} S_{v_j}^{-1} m_{v_j} 
    + 
    \rho \del{\frac{|O_j|}{\vert \minibatch\vert}\sum_{i\in \minibatch} \hat{d}_{i,j}-2\hat{D}_{i,j} m_{v_j}},
    \label{eq:PMF-mcssvi-update-m}
\end{equation}
where $\hat{d}_{i,j}$ is %
\begin{eqnarray*}
\hat{d}_{i,j} & = &
\frac{1}{k_1 k_2}\sum_{a=1}^{k_1} 
u_i^a
\sum_{b=1}^{k_2}
\left[ \frac{\partial}{\partial f_{i,j}}  \log p(y_{i,j}|f^{a,b}_{i,j})\right].
\end{eqnarray*}
The $\{u_i^\cdot\}$ are $k_1$ samples from $\mathcal{N}(m_{u_i},S_{u_i})$ and $\{f_{i,j}^{a,\cdot}\}$ are $k_2$ samples from $\mathcal{N}(m_{v_j}^T u_i^a,(u_i^a)^T S_{v_j} u_i^a)$.
The \dsvi{} update for the variational mean of column $v_j$ is given by
\begin{equation}
\label{eq:PMF-dsvi-update-mean}
    m_{v_j} \leftarrow m_{v_j} + \rho \left(\Sigma^{-1} (\mu-m_{v_j}) + \frac{|O_j|}{|\minibatch|}\sum_{i\in \minibatch} \hat{d}_{i,j}\right)
    .
\end{equation}
The \dsvi{} update for the Cholesky factor of the variational covariance of column $v_j$ is given by
\begin{equation}
\label{eq:PMF-dsvi-update-C}
    C_{v_j} \leftarrow C_{v_j} + \rho~\text{triu}\left( (C_{v_j}\circ I)^{-1} - C_{v_j}\Sigma^{-1}  + 2C_{v_j}\frac{|O_j|}{\vert \minibatch\vert}\sum_{i\in \minibatch} \hat{D}_{i,j} \right).
\end{equation}
The update equations for the variational mean and Cholesky factor of the variational covariance of column $u_i$ are obtained symmetrically.
The update for hyperparameter $\sigma_V^2$ is given by 
\begin{equation}
\label{eq:PMF-hyp-update-sigma}
    \sigma_V^2 \leftarrow \frac{1}{M D}\sum_{j=1}^M \left( \text{trace}(S_{v_j}) + m_{v_j}^T m_{v_j}\right)
\end{equation}
and the update for $\sigma_U^2$ is similar.

\paragraph{Equations for CTM:}
In CTM, the \dsvi{} update for a document's variational mean $m_d$ is given by
\begin{equation}
\label{eq:CTM-dsvi-update-mean}
    m_d \leftarrow m_d + \rho \left( \Sigma^{-1}(\mu - m_d) + \sum_n \hat{d}_n(m_d)\right)
\end{equation}
where 
\begin{equation}
    \hat{d}_n(m_d) = \frac{1}{N_{MC}} \sum_{\ell=1}^{N_{MC}} \nabla_{\eta_d} [\xi_n(\eta_d^{(\ell)})], \quad \eta_d^{(\ell)}\sim \mathcal{N}(\eta_d\vert m_d, \cov_d).
\end{equation}
The \dsvi{} update for the Cholesky factor $C_d$ (where $\cov_d=C_d^T C_d$) is given by
\begin{equation}
\label{eq:CTM-dsvi-update-C}
    C_d \leftarrow C_d + \rho~\text{triu}\left( (C_d\circ I)^{-1} - C_d\Sigma^{-1}  + C_d\sum_n \hat{D}_n(C_d) \right).
\end{equation}
where
\begin{equation}
\hat{D}_n(C_d) = \frac{1}{N_{MC}} \sum_{\ell=1}^{N_{MC}} \epsilon^{(\ell)} \left( \nabla_{\eta_d} [\xi_n(m_d + C_d^T \epsilon^{(\ell)})] \right)^T, \quad \epsilon^{(\ell)}\sim \mathcal{N}(\epsilon\vert 0, I).
\end{equation}

\subsection{\hsvi{} for CTM using the VLB (\ref{eq:vlb})}
In this section
we provide details of the VLB, variational optimization, and hyperparameter optimization 
for the application of the simple structured approximation (\ref{eq:SimplestructuredQ}) to CTM, 
The evaluation of (\ref{eq:vlb}) for CTM yields
\begin{equation}
    \log p(\text{data}) \ge 
    - \sum_{d=1}^D d_{\text{KL}}(q(\eta_d)\Vert p(\eta_d))
+
    \sum_{d=1}^D  \sum_{n=1}^{N_d} \sum_k \log \beta_{k,w_{dn}}
    E_{q(\eta_{d})} [ h_k(\eta_d)  ].    
    \label{eq:ctm_vlb_subopt}
\end{equation}
The resulting fixed-point update rule for the variational covariance of a document is given by 
\begin{equation}
\label{eq:CTM-SVI-update-V-simple}
    \cov_d^{-1} \leftarrow (1-\rho) \cov_d^{-1} + \rho \left(\Sigma^{-1} + \Pi_{S^{++}}(- 2\sum_{n=1}^{N_d} \hat{D}_n^s(\cov_d)) \right)
\end{equation}
where
\begin{equation}
    \hat{D}_n^s(\cov_d) = \frac{1}{2}\frac{1}{N_{MC}}\sum_k \log \beta_{k,w_{dn}} \sum_{l=1}^{N_{MC}} \nabla^2_{\eta_d}[h_k(\eta_d^{(\ell)})], \quad \eta_d^{(\ell)}\sim \mathcal{N}(\eta_d\vert m_d, \cov_d).
\end{equation}
The S-DSVI update for a document's variational mean is given by 
\begin{equation}
    m_d \leftarrow m_d + \rho \left( \Sigma^{-1}(\mu - m_d) + \sum_n \hat{d}^s_n(m_d)\right)
\label{eq:CTM-dsvi-update-mean-simple}
\end{equation}
where
\begin{equation}
    \hat{d}_n^s(m_d) = \frac{1}{N_{MC}} \sum_k \log \beta_{k,w_{dn}} \sum_{\ell=1}^{N_{MC}} \nabla_{\eta_d} [h_k(\eta_d^{(\ell)})], \quad \eta_d^{(\ell)}\sim \mathcal{N}(\eta_d\vert m_d, \cov_d).
\end{equation}

Hyperparameter optimization for the global prior parameters is exactly the same as in the case of the approximation (\ref{eq:truestructuredQ}).
Optimization of the topics occurs by rows and uses the constraint $\sum_{i=1}^V \beta_{k,i}=1$ for a row $k$.
The Lagrangian is given by
\begin{align*}
    \mc{L}_k & = \sum_{d=1}^D \sum_{n=1}^{N_d} \log \beta_{k,w_{dn}} \int q(\eta_d)f_k(\eta_d)d\eta_d+ \gamma \left(\sum_{i=1}^V \beta_{k,i}-1\right)\\
    & = \sum_{d=1}^D \sum_{n=1}^{N_d} \sum_{i=1}^V w_{dni} \log \beta_{k,i} \int q(\eta_d)f_k(\eta_d)d\eta_d+ \gamma \left(\sum_{i=1}^V \beta_{k,i}-1\right)\\
    & = \sum_{i=1}^V \alpha_{ki} \log \beta_{k,i} + \gamma \left(\sum_{i=1}^V \beta_{k,i}-1\right)
\end{align*}
where $\alpha_{ki} = \sum_{d=1}^D \sum_{n=1}^{N_d} w_{dni} \int q(\eta_d)f_k(\eta_d)d\eta_d$, and $w_{dni} = 1$ if $w_{dni} = i$ and $0$ otherwise.
Maximizing with respect to $\beta_{k,i}$ yields:
\begin{equation*}
    \hat{\beta}_{k,i} = \frac{\alpha_{ki}}{\sum_{j=1}^V \alpha_{kj}}.
\end{equation*}

The algorithm  \hsvi{} iterates in a round robin manner over documents and updates the corresponding parameters of the current document 
using (\ref{eq:CTM-SVI-update-V-simple}) and (\ref{eq:CTM-dsvi-update-mean-simple}). 
Parameters are updated after each document.

\subsection{Update equations for GME using the VLB (\ref{eq:vlb})}
\label{app:subsec-gme-subopt}
For optimization of the sub-optimal VLB, derivatives of $\expec{q(f)}{\log p(y|f)}$ w.r.t. expectation parameters $h, H$, and $g$ are needed.
The expectation $\expec{q(f)}{\log p(y|f)}$ can be written as
\begin{align}
    \expec{q(f)}{\log p(y|f)}
    & = 
    \expec{q(w_1)q(w_2)}{
        \phi(w_1, w_2)
    }
    ,
\end{align}
with 
\begin{equation}
    \phi(w_1,w_2) = \expec{\pdfnorm{f}{w_1^T x}{w_2}}{\log p(y|f)}
    \label{eq:subopt-phi}
    .
\end{equation}
Similar to (\ref{eq:opth}) and (\ref{eq:optH}),
\begin{align}
    \pd{}{h}\sbr{
        \expec{q(w_1)q(w_2)}{
            \phi(w_1, w_2)
        }
    }
    & = 
    \expec{q(w_1)q(w_2)}{
        \nabla_{w_1}
        \phi(w_1, w_2)
        -
        \nabla_{w_1}^2
        \phi(w_1, w_2)
        m
    }
    \label{eq:subopth}
    \\
    \pd{}{H}\sbr{
        \expec{q(w_1)q(w_2)}{
            \phi(w_1, w_2)
        }
    }
    & = 
    -\frac{1}{2}
    \expec{q(w_1)q(w_2)}{
        \nabla_{w_1}^2
        \phi(w_1, w_2)
    }
    \label{eq:suboptH}
    ,
\end{align}
where
\begin{align}
    \nabla_{w_1}
    \phi
    & = 
    x
    \expec{\pdfnorm{f}{w_1^T x}{w_2}}{\pd{}{f} \log p(y|f)}
    \label{eq:subopt-phi-first-deriv-wone}
    ,
    \\
    \nabla_{w_1}^2
    \phi
    & = 
    x x^T
    \expec{\pdfnorm{f}{w_1^T x}{w_2}}{\pd[2]{}{f} \log p(y|f)}
    \label{eq:subopt-phi-second-deriv-wone}
    .
\end{align}
We have from (\ref{eq:derivexpqrayl})
\begin{align}
    \pd{}{g}\sbr{
        \expec{q(w_1)q(w_2)}{
            \phi(w_1, w_2)
        }
    }
    & = 
    \expec{q(w_1)q(w_2)}{
        \frac{w_2^2}{4\sigma^2}
        \pd{}{w_2} \phi(w_1,w_2)
    }
    ,
\end{align}
where
\begin{align}
    \pd{}{w_2}
    \phi
    & = 
    \frac{1}{2}
    \expec{\pdfnorm{f}{w_1^T x}{w_2}}{\pd[2]{}{f} \log p(y|f)}
    \label{eq:subopt-phi-first-deriv-wtwo}
    .
\end{align}

Given $\phi_i(w_1,w_2)$ (\ref{eq:subopt-phi}) and its derivatives (\ref{eq:subopt-phi-first-deriv-wone}, \ref{eq:subopt-phi-second-deriv-wone}, \ref{eq:subopt-phi-first-deriv-wtwo}), the suboptimal approximation update equations for standard and natural parameters are given by
\begin{align}
    m 
    & \leftarrow 
    m 
    + 
    \rho \del{
        -\Sigma^{-1}\del{m-\mu}
        +
        \expec{q(w_1)q(w_2)}{
            \frac{N}{|\minibatch|}  
            \sum_{i\in \minibatch}
            \nabla_{w_1}
            \phi_i(w_1, w_2)
        }
    },
    \label{eq:subopt-m}
    \\
    C 
    & \leftarrow 
    C 
    + 
    \rho \del{
        -\text{triu}\del{C \Sigma^{-1}}
        +\del{C \circ I}^{-1}
        +
        \text{triu}\del{
            C 
            \expec{q(w_1)q(w_2)}{
                \frac{N}{|\minibatch|}  
                \sum_{i\in \minibatch}
                \nabla_{w_1}^2 \phi_i(w_1, w_2)
            }
        }
    },
    \label{eq:subopt-C}
    \\
    \sigma 
    & \leftarrow 
    \sigma 
    + 
    \rho \del{
        -\del{
            \frac{2\sigma}{\tau^2}
            -\frac{2}{\sigma} 
        }
        +\frac{1}{\sigma}
        \expec{q(w_1)q(w_2)}{
            \frac{N}{|\minibatch|}  
            \sum_{i\in \minibatch}
            w_2^2 \pd{}{w_2} \phi_i(w_1, w_2)
        }
    },
    \label{eq:subopt-sigma}
    \\
    \cov^{-1}m 
    & \leftarrow 
    \del{1-\rho} \cov^{-1}m 
    + 
    \rho \del{
        \Sigma^{-1}\mu
        +
        \expec{q(w_1)q(w_2)}{
            \frac{N}{|\minibatch|}  
            \sum_{i\in \minibatch}
            \nabla_{w_1}
            \phi_i(w_1, w_2)
            -
            \nabla_{w_1}^2
            \phi_i(w_1, w_2)
            m
        }
    },
    \label{eq:subopt-nat-m}
    \\
    \cov^{-1}
    & \leftarrow 
    (1-\rho) \cov^{-1} 
    + 
    \rho \del{
        \Sigma^{-1}
        -\expec{q(w_1)q(w_2)}{
            \frac{N}{|\minibatch|}  
            \sum_{i\in \minibatch}
            \nabla_{w_1}^2 \phi_i(w_1, w_2)
        }
    },
    \label{eq:subopt-nat-V}
    \\
    -\frac{1}{2\sigma^2}
    & \leftarrow 
    \del{1-\rho} \del{-\frac{1}{2\sigma^2}}
    + 
    \rho \del{
        -\frac{1}{2\tau^2}
        +\frac{1}{4\sigma^2}
        \expec{q(w_1)q(w_2)}{
            \frac{N}{|\minibatch|}  
            \sum_{i\in \minibatch}
            w_2^2 \pd{}{w_2} \phi_i(w_1, w_2)
        }
    }
    \label{eq:subopt-nat-sigma}
    .
\end{align}
S-DSVI utilizes (\ref{eq:subopt-m}),(\ref{eq:subopt-C}),and (\ref{eq:subopt-sigma}).
MC-SSVI utilizes (\ref{eq:subopt-nat-m}),(\ref{eq:subopt-nat-V}),and (\ref{eq:subopt-nat-sigma}).
H-MC-SSVI utilizes (\ref{eq:subopt-m}),(\ref{eq:subopt-nat-V}),and (\ref{eq:subopt-nat-sigma}).

\subsection{Update equations for GME using a mean field approximation}
\label{app:subsec-gme-mf}
The factorizing distribution $q(w,f)=q(w)\prod_i q(f_i)$ leads to the mean field VLB
\begin{equation}
    -\text{KL}\del{q(w)\Vert p(w)}
    +
    \sum_i 
        \expec{q(w)}{
            \expec{q(f_i)}{
                \log
                \frac{p(y_i|f_i)p(f_i|w)}{q(f_i)}
            }
        }
    \label{eq:mfvlb}
    .
\end{equation}
Using $q(f_i)=\pdfnorm{f_i}{\beta_i}{\gamma_i^2}$, and letting $\phi_i(w_2)=\expec{q(w_1)q(f_i)}{\log p(f_i|w_1,w_2)}$,
\begin{align}
    \phi_i(w_2)
    & = 
    -\frac{1}{2}
    \expec{q(w_1)q(f_i)}{
        \log w_2 
        + 
        \log \del{2\pi} 
        + 
        \frac{1}{w_2}\del{
            f_i^2 - 2 f_i \del{w_1^T x_i} + \del{w_1^T x_i}^2
        }
    }
    \nonumber
    \\
    & = 
    -\frac{1}{2}
    \expec{q(w_1)}{
        \log w_2 
        + 
        \log \del{2\pi} 
        + 
        \frac{1}{w_2}
        \del{
            \beta_i^2 + \gamma_i^2 
            - 2\beta_i \del{w_1^T x_i} 
            + \del{w_1^T x_i}^2
        }
    }
    \nonumber
    \\
    & = 
    -\frac{1}{2}
    \del{
        \log w_2 
        + 
        \log \del{2\pi} 
        + 
        \frac{1}{w_2}
        \del{
            \beta_i^2 + \gamma_i^2 
            - 2\beta_i \del{m^T x_i} 
            + x_i^T \del{m m^T + \cov} x_i
        }
    }
    .
\end{align}
The mean field VLB (\ref{eq:mfvlb}) becomes
\begin{equation}
    -\text{KL}\del{q(w)\Vert p(w)}
    +
    \sum_i
    \expec{q(f_i|\beta_i,\gamma_i^2)}{\log p(y_i|f_i)}
    + \expec{q(w_2)}{\phi_i(w_2)}
    +\frac{1}{2}\log\del{2\pi e \gamma_i^2}
    .
\end{equation}

Noting $\expec{q(w_2)}{\frac{1}{w_2}}=\frac{\sqrt{2\pi}}{\sigma}$, the derivatives of $\expec{q(w_2)}{\phi_i(w_2)}$ w.r.t. the standard parameters $m$, $\cov$, $\beta_i$, and $\gamma_i$ are given by
\begin{align}
    \pd{}{m} \expec{q(w_2)}{\phi_i(w_2)}
    & = 
    \frac{\sqrt{2\pi}}{\sigma}\del{\beta_i-x_i^T m} x_i
    ,
    \\
    \pd{}{\cov} \expec{q(w_2)}{\phi_i(w_2)}
    & = 
    -\frac{\sqrt{2\pi}}{2 \sigma} x_i x_i^T
    ,
    \\
    \pd{}{\beta_i} \expec{q(w_2)}{\phi_i(w_2)}
    & = 
    \frac{\sqrt{2\pi}}{\sigma}\del{x_i^T m-\beta_i}
    ,
    \\
    \pd{}{\gamma_i} \expec{q(w_2)}{\phi_i(w_2)}
    & = 
    -\frac{\sqrt{2\pi}}{\sigma}\gamma_i
    ,
    \\
    \pd{}{\sigma} \expec{q(w_2)}{\phi_i(w_2)}
    & = 
    \frac{1}{\sigma} \expec{q(w_2)}{w_2\pd{}{w_2}\phi_i(w_2)}
    \nonumber
    \\
    & = 
    -\frac{1}{2\sigma}
    +\frac{\sqrt{2\pi}}{2\sigma^2}\del{
        \beta_i^2 + \gamma_i^2 
        - 2\beta_i \del{m^T x_i} 
        + x_i^T \del{m m^T + \cov} x_i
    }
    .
\end{align}

The updates for the standard parameters are given by
\begin{align}
    m 
    & \leftarrow 
    \del{\Sigma^{-1} + \frac{\sqrt{2\pi}}{\sigma}\del{\sum_i x_i x_i^T}}^{-1}
    \del{\Sigma^{-1}\mu + \frac{\sqrt{2\pi}}{\sigma}\del{\sum_i \beta_i x_i}}
    \label{eq:mf-m}
    ,
    \\
    \cov
    & \leftarrow 
    \del{\Sigma^{-1} + \frac{\sqrt{2\pi}}{\sigma}\del{\sum_i x_i x_i^T}}^{-1}
    \label{eq:mf-V}
    ,
    \\
    \beta_i 
    & \leftarrow
    \beta_i 
    +
    \rho
    \del{
        \frac{\sqrt{2\pi}}{\sigma} \del{ m^T x_i -\beta_i}
        +
        \expec{q(f_i|\beta_i,\gamma_i^2)}{\pd{}{f_i}\log p(y_i|f_i)}
    }
    ,
    \\
    \gamma_i 
    & \leftarrow
    \gamma_i 
    +
    \rho
    \del{
        -\frac{\sqrt{2\pi}}{\sigma} \gamma_i
        +
        \gamma_i\expec{q(f_i|\beta_i,\gamma_i^2)}{\pd[2]{}{f_i}\log p(y_i|f_i)}
        +
        \frac{1}{\gamma_i}
    }
    ,
    \\
    \sigma 
    & \leftarrow 
    \text{RealRoot}\del{
        -\frac{2}{\tau^2}\sigma^3
        +(2-\frac{N}{2}) \sigma
        +\frac{\sqrt{2\pi}}{2} \sum_i z_i
    }
    \label{eq:mf-sigma}
    ,
\end{align}
where $z_i = \beta_i^2 + \gamma_i^2 - 2\beta_i \del{m^T x_i} + x_i^T \del{m m^T + \cov} x_i$.
The updates for $m$, and $\cov$ are available in closed-form.
The update for $\sigma$ requires finding the root of a 3rd order polynomial equation in $\sigma$.
The updates for $\beta_i$ and $\gamma_i^2$ require optimization.

\subsection{Additional experimental details}
\label{sec:appendix_exp_details}

\paragraph{General Details:}

All training runs were conducted using 4 physical cores of either an Intel Xeon E5-2660 v2 CPU (2.20 GHz) or an AMD Opteron 6380 CPU (2.5 GHz).

The value of the VLB is not available in closed form and was estimated for the plots.  
In the case of GME with Poisson likelihood, 10 samples for the expectation over $w$ and 100 samples for the expectation over $f_i$ were used to estimate the negative VLB.
For PMF, $10\times 10$ samples were used.
In the case of CTM, 10 samples were used to estimate the negative VLB.

Where required, the GPML toolbox\footnote{\url{http://www.gaussianprocess.org/gpml/code/matlab/doc/}} is used to calculate expectations of the log liklihood and its derivatives.

\paragraph{GME:}

Let the learned posteriors over $w_1$ and $w_2$ be denoted by $q(w_1)=\pdfnorm{w_1}{m}{\cov}$ and $q(w_2)=\text{Rayl}(w_2\vert\sigma^2)$.
Then, the predictive distribution for a new example $\xtest$ is given by
\begin{align*}
    p(\ytest | \D) & = \int p(\ytest | \ftest) p(\ftest | \D) \dif \ftest \\
    & = \int p(\ytest | \ftest) \pdfnorm{\ftest}{w_1^T \xtest}{w_2} q(w_1) q(w_2) \dif \ftest \dif w_1 \dif w_2 \\
    & = \int p(\ytest | \ftest) \pdfnorm{\ftest}{m^T \xtest}{\xtest^T \cov \xtest + w_2}  q(w_2) \dif \ftest \dif w_2 \\
\end{align*}
This integral is calculated with 200-point Gauss-Hermite quadrature:
\begin{equation*}
    p(\ytest\vert\D) \approx \sum_i g(u_i) v_i,
\end{equation*}
where $g(u_i) = \int p(\ytest\vert\ftest) \pdfnorm{\ftest}{m^T \xtest}{\xtest^T \cov \xtest + u_i} \dif \ftest$.

\paragraph{sGP:}
In the conjugate sparse GP experiment, labels were normalized using $Z$-score normalization (normalization performed on train and applied to test).

A zero mean function and RBF kernel were used.
Hyperparameter selection was performed with brute force grid search over length scale $[0.1,50]$, kernel variance $[0.1^2,10^2]$, and likelihood variance $[0.01,1]$.
20 points per parameter interval were used and the final parameters were selected as those that resulted in the best log marginal likelihood on a randomly selected 100 sample subset of training data.

\paragraph{PMF:}
The PMF priors for the columns of $U$ and $V$ were zero-mean, spherical Gaussian distributions with component variances given by $\sigma_U^2$ and $\sigma_V^2$, respectively.
The settings for both the generative model and training initialization were $\sigma_U^2=\sigma_V^2=1$.
The ordinal likelihood slope was set to 100 and delta set to 15.

In PMF, the predictive log likelihood and predictive estimate $\hat{y}_{ij}$ of a test set observation $y_{ij}$ are approximated as expectations with respect to a univariate Gaussian distribution as in \cite{Hernandez-LobatoHG14}.
Specifically, the predictive log likelihood is approximated as
\begin{equation*}
    \log p(y_{ij}) \approx \log \int p(y_{ij}\vert f_{ij}) \mc{N}(f_{ij}\vert m_{ij}, S_{ij}) df_{ij}
\end{equation*}
where $m_{ij} = E(u_i^T v_j) = m_{u_i}^T m_{v_j}$ and $S_{ij} = \text{Var}(u_i^T v_j) = \text{tr}(\cov_{u_i}\cov_{v_j}) + m_{u_i}^T \cov_{v_j} m_{u_i} + m_{v_j}^T \cov_{u_i} m_{v_j}$.
This 1-D integral is estimated by Monte Carlo with 1000 samples.
The predictive ordinal and count estimates use the same normal approximation to the latent variable and are calculated with 100 and 20-point Gauss-Hermite quadrature, respectively.
Predictive binary estimates are determined by comparing $m_{ij}$ to $\frac{1}{2}$.

\paragraph{CTM:}

The 1500-document \emph{nips} dataset \citep{UCI} was pre-processed to remove vocabulary words that did not occur more than 10 times in the corpus and documents that did not contain more than 10 words.
After an 80/20 split into training and test sets, the test set was further processed by removing vocabulary words that only appeared in the test set.
The final vocabulary size was 10,916 words; the training corpus size was 1193 documents (1,535,973 words); and the test corpus size was 298 documents (383,108 words).

The hyperparameters for the CTM experiments were initialized as $\mu=0$ and $\Sigma=I$ and the rows of $\beta$ (topics) were initialized proportional to the vocabulary counts of randomly selected documents plus random numbers between 1 and 2.

Default training settings of the mean field variational inference implementation \cite{Blei2006} were used with two exceptions: 1) the maximum number of variational EM iterations was increased from 1000 to 2000, and 2) the variational EM tolerance for stopping was lowered from $10^{-3}$ to $10^{-9}$. 
Both of these changes were made to support comparison of variational bound values between all the methods.

The predictive log likelihood of a document $w$ is given by
\begin{equation}
    \log p(w\vert \mu, \Sigma, \beta) = \log \int p(w\vert \eta, \beta) p(\eta \vert \mu, \Sigma) d\eta
	\label{eq:ctmnll}
\end{equation}
This quantity is estimated using multiple types of sampling schemes.
In the first scheme, Monte Carlo sampling from the prior $p(\eta\vert \mu, \Sigma)$ is utilized.
For the remaining schemes, importance sampling from a posterior is utilized. 
The posterior is calculated for a test document using 100 iterations of batch \hsvi{} with diagonal covariances.
To provide some protection against narrow posteriors, we utilize two additional variants of the importance sampling scheme that additively inflate the posterior covariance by $10^{-1}$ and $1$.
In each scheme, 10 batches of $10^6$ samples are used to estimate the previous integral.

\subsection{Additional experimental results}

\paragraph{sGP}
\putSGPPlotNonRef
In Figure \ref{fig:sgpplotnonref}, we show the absolute, i.e., non-referenced, performance of the methods.
As the figure shows, there is variability in performance based on the sub-sample used to train.
Although simpler to interpret, this plot hides the fact that almost every trial of the full GP, optimal (cubic) approximation, and variant 1 performed better than the sub-optimal approximation.
For this reason, we provide the referenced plot in the main paper.

\paragraph{PMF}
Figure \ref{fig:pmfplotserrb} shows an additional error metric for the artificial dataset in PMF with count likelihood. 
The error measure, non-zero error, refers to the fraction of true zero-counts that were predicted to have non-zero counts. 
Again, we observe the best performing algorithm to be H-\alg{}.
Note, a similar plot does not exist for the real dataset because true zero counts cannot be distinguished from unobserved entries.

Figure \ref{fig:pmfplotsvlbzoom} is a zoom-in of the bottom row of Figure \ref{fig:pmfplotsvlb}.
We further observe that optimal sub-sampling size, in this case for H-\alg{}, is problem dependent even with a fixed likelihood function.

\alg{} was unstable in PMF and required tuning of the learning rate for stable performance.
However, even with tuning, it did not achieve competitive performance. Of the six datasets its best performance was on the artificial binary dataset.
Figure \ref{fig:pmfplotvlbmcssvi} is the same as Figure \ref{fig:pmfplotsvlb} top, left column, but with a run of MC-SSVI at 250 sub-samples. As can be seen H-\alg{} performs significantly better. Overall we can conclude that H-\alg{} performs best in generalized PMF.

\putPMFPlotsErrB
\putPMFPlotsVLBZoom
\putPMFPlotVLBMCSSVI
\putStaircasePlots
\putCTMPlotsNLL
\putWangBleiEval
\putCTMKPlots

\paragraph{CTM}
An evaluation of the normalized predictive log likelihood calculation is shown in Figure \ref{fig:staircaseplots}.
Each variant of importance sampling is utilized to approximate (\ref{eq:ctmnll}) on 10 test documents of \emph{nips}.
As the number of Monte Carlo samples increases, we see the estimates output by the variants converging to roughly the same values.

Figure \ref{fig:ctmplotsnll} shows CTM NLL calculated on the \emph{nips} dataset as a function of training time per algorithm.
As mentioned previously, four schemes were used to estimate the NLL. 
The plots show some variability in the ordering 
of the mean field algorithm with respect to \dsvi{}. However, all 4 evaluations show that \hsvi{} with the optimal approximation performs best and 
\hsvi{} with the simple structured approximation performing worst.

As an alternative to importance sampling, \cite{Wang2013} approximate (\ref{eq:ctmnll}) with 
\begin{equation}
    \log p(w_\text{2nd half}\vert \eta=E_\eta q(\eta \vert w_\text{1st half}), \beta)
    \label{eq:wangblei}
\end{equation}
where $w_\text{2nd half}$ represents half the words of a test document, and $q(\eta\vert w_\text{1st half})$ represents the learned variational posterior using the training algorithm on the other half of the test document words.
Note that (\ref{eq:ctmnll}) can equivalently be expressed as
\begin{equation}
    \log p(w\vert \mu, \Sigma, \beta) = \log p(w\vert \eta, \beta) + \log p(\eta \vert \mu, \Sigma) - \log p(\eta \vert w, \mu, \Sigma, \beta)
    \label{eq:ctmnllpoint}
\end{equation}
for any point $\eta$.
Hence, the approximation of \cite{Wang2013} can be understood as the evaluation of the first term of (\ref{eq:ctmnllpoint}) at the mean value of $\eta$ w.r.t. the variational posterior.

The left plot of Figure \ref{fig:wangbleieval} shows the results of this approximation on \emph{nips} with $K=50$ as a function of training time.
Variational posteriors over test were learned by the individual algorithms, and each algorithm was run for 1000 iterations.
H-MC-SSVI on the optimal approximation achieves the best performance followed by mean field and then S-DSVI. 
The performance of the suboptimal approximation curiously decreases over time.
However, the right plot provides some caution against utilizing just the first term of (\ref{eq:ctmnllpoint}).
Here, the log prior term is added and results in a larger performance gap between mean field and H-MC-SSVI on the optimal approximation similar to the posterior sampling results in Figure \ref{fig:ctmplotsnll}.

We note that calculation of the third term of (\ref{eq:ctmnllpoint}) is non-trivial.
One approach would be to use a ``Chib''-style estimation (e.g,. see \cite{Wallach2009}), but the dimensionality involved here makes the task extremely computationally demanding.

Figure \ref{fig:ctmkplots} shows the same 4 types of evaluation for the experiment comparing performance as a function of 
the latent dimensionality. We see that all 4 evaluations agree with the one in the main paper and \hsvi{} performs better than the mean field approximation.

Figure \ref{fig:mcsvi-ctm-vlb-enron} shows the output of an additional CTM experiment conducted on the \emph{enron} dataset \cite{UCI}. 
The full dataset was sub-sampled to 4000 documents resulting in a vocabulary size of 22,505 words and 617,666 total words in the training corpus. 
The latent dimensionality of this experiment was 10. 
The \hsvi{} algorithm used diagonal covariances, and processed all documents for each update of the hyperparameters.
As in the {\em nips} datset, we see that with increased training time \hsvi{}  can locate a better optimum than the mean field algorithm due to the use of a structured approximation.

\putCTMVLBPlotEnron{fig:mcsvi-ctm-vlb-enron}{Additional experiment in CTM on \emph{enron} dataset showing Monte Carlo-sampled negative VLB as a function of training time for batch H-\alg{} and the mean field algorithm (both with diagonal covariances).}

\clearpage
\newpage
\small
\bibliography{../rsheth80}

\begin{thebibliography}{10}

\bibitem{Amari2000}
Shun-Ichi Amari and Hiroshi Nagaoka.
\newblock {\em Methods of Information Geometry}, volume 191 of {\em
  Translations of Mathematical monographs}.
\newblock Oxford University Press, 2000.

\bibitem{Blei2006}
David~M. Blei and John~D. Lafferty.
\newblock Correlated topic models.
\newblock In {\em Advances in Neural Information Processing Systems 18}, pages
  147--154, 2006.

\bibitem{Blei2003}
David~M. Blei, Andrew~Y. Ng, and Michael~I. Jordan.
\newblock Latent {D}irichlet allocation.
\newblock {\em Journal of Machine Learning Research}, pages 993--1022, 2003.

\bibitem{Challis2013}
Edward Challis and David Barber.
\newblock {G}aussian {K}ullback-{L}eibler approximate inference.
\newblock {\em Journal of Machine Learning Research}, 14:2239--2286, 2013.

\bibitem{Duchi2011short}
John Duchi, Elad Hazan, and Yoram Singer.
\newblock Adaptive subgradient methods for online learning and stochastic
  optimization.
\newblock {\em {JMLR}}, pages 2121--2159, 2011.

\bibitem{GhahramaniB00}
Zoubin Ghahramani and Matthew~J. Beal.
\newblock Propagation algorithms for variational bayesian learning.
\newblock In {\em Advances in Neural Information Processing Systems 13, Papers
  from Neural Information Processing Systems {(NIPS)} 2000, Denver, CO, {USA}},
  pages 507--513, 2000.

\bibitem{Hensman2013}
James Hensman, Nicolo Fusi, and Neil~D. Lawrence.
\newblock {G}aussian processes for big data.
\newblock In {\em Proceedings of the 29th UAI Conference}, pages 282--290,
  2013.

\bibitem{HensmanRL12}
James Hensman, Magnus Rattray, and Neil~D. Lawrence.
\newblock Fast variational inference in the conjugate exponential family.
\newblock In {\em NIPS}, pages 2897--2905, 2012.

\bibitem{Hernandez-LobatoHG14}
Jos{\'{e}}~Miguel Hern{\'{a}}ndez{-}Lobato, Neil Houlsby, and Zoubin
  Ghahramani.
\newblock Stochastic inference for scalable probabilistic modeling of binary
  matrices.
\newblock In {\em {ICML}}, pages 379--387, 2014.

\bibitem{Hoffman2015}
Matthew~D. Hoffman and David~M. Blei.
\newblock Structured stochastic variational inference.
\newblock In {\em AISTATS}, pages 361--369, 2015.

\bibitem{Hoffman2013}
Matthew~D. Hoffman, David~M. Blei, Chong Wang, and John Paisley.
\newblock Stochastic variational inference.
\newblock {\em Journal of Machine Learning Research}, pages 1303--1347, 2013.

\bibitem{Khanthesis}
Mohammad~Emtiyaz Khan.
\newblock {\em {Variational Learning for Latent Gaussian Models of Discrete
  Data}}.
\newblock PhD thesis, The University of British Columbia, 2012.

\bibitem{KhanKS14}
Mohammad~Emtiyaz Khan, Young~Jun Ko, and Matthias~W. Seeger.
\newblock Scalable collaborative {B}ayesian preference learning.
\newblock In {\em AISTATS}, pages 475--483, 2014.

\bibitem{Khan2017}
Mohammad~Emtiyaz Khan and Wu~Lin.
\newblock Conjugate-computation variational inference: Converting variational
  inference in non-conjugate models to inferences in conjugate models.
\newblock In {\em {AISTATS}}, volume~54, pages 878--887, 2017.

\bibitem{Kingma2015}
Diederik~P. Kingma, Tim Salimans, and Max Welling.
\newblock Variational dropout and the local reparameterization trick.
\newblock In {\em NIPS}, pages 2575--2583, 2015.

\bibitem{Kingma2014}
Diederik~P. Kingma and Max Welling.
\newblock Auto-encoding variational {B}ayes.
\newblock arXiv:1311.6371, 2014.

\bibitem{KnowlesM11}
David~A. Knowles and Tom Minka.
\newblock Non-conjugate variational message passing for multinomial and binary
  regression.
\newblock In {\em Advances in Neural Information Processing Systems}, pages
  1701--1709, 2011.

\bibitem{KoK14}
Young-Jun Ko and Mohammad Khan.
\newblock Variational {G}aussian inference for bilinear models of count data.
\newblock In {\em Proceedings of the Sixth Asian Conference on Machine
  Learning}, pages 330--343, 2014.

\bibitem{KoenigsteinP13}
Noam Koenigstein and Ulrich Paquet.
\newblock Xbox movies recommendations: variational {B}ayes matrix factorization
  with embedded feature selection.
\newblock In {\em 7th {ACM} Conf. on Rec. Systems}, pages 129--136, 2013.

\bibitem{KucukelbirTRGB16}
Alp Kucukelbir, Dustin Tran, Rajesh Ranganath, Andrew Gelman, and David~M.
  Blei.
\newblock Automatic differentiation variational inference.
\newblock {\em ArXiv 1603.00788}, 2016.

\bibitem{UCI}
M.~Lichman.
\newblock {UCI} machine learning repository, 2013.
\newblock \url{http://archive.ics.uci.edu/ml}.

\bibitem{LimTeh2007}
Yew~J. Lim and Yee~W. Teh.
\newblock Variational {B}ayesian approach to movie rating prediction.
\newblock In {\em Proceedings of KDD Cup and Workshop}, 2007.

\bibitem{PaquetTW12}
Ulrich Paquet, Blaise Thomson, and Ole Winther.
\newblock A hierarchical model for ordinal matrix factorization.
\newblock {\em Statistics and Computing}, pages 945--957, 2012.

\bibitem{RaikoIK07}
Tapani Raiko, Alexander Ilin, and Juha Karhunen.
\newblock Principal component analysis for large scale problems with lots of
  missing values.
\newblock In {\em ECML}, pages 691--698, 2007.

\bibitem{Ranganath2014}
Rajesh Ranganath, Sean Gerrish, and David~M. Blei.
\newblock Black box variational inference.
\newblock In {\em Proceedings of the Seventeenth International Conference on
  Artificial Intelligence and Statistics, {AISTATS}}, pages 814--822, 2014.

\bibitem{Rezende2014}
Danilo~Jimenez Rezende, Shakir Mohamed, and Daan Wierstra.
\newblock Stochastic backpropagation and approximate inference in deep
  generative models.
\newblock In {\em ICML}, pages 1278--1286, 2014.

\bibitem{robbins-monro}
H.~Robbins and S.~Monro.
\newblock A stochastic approximation method.
\newblock {\em Annals of Mathematical Statistics}, 22:400--407, 1951.

\bibitem{SalakhutdinovM07}
Ruslan Salakhutdinov and Andriy Mnih.
\newblock Probabilistic matrix factorization.
\newblock In {\em NIPS}, pages 1257--1264, 2007.

\bibitem{SalakhutdinovM08a}
Ruslan Salakhutdinov and Andriy Mnih.
\newblock Bayesian probabilistic matrix factorization using {M}arkov chain
  {M}onte {C}arlo.
\newblock In {\em ICML}, pages 880--887, 2008.

\bibitem{salimans2013fixed}
Tim Salimans, David~A Knowles, et~al.
\newblock Fixed-form variational posterior approximation through stochastic
  linear regression.
\newblock {\em Bayesian Analysis}, 8(4):837--882, 2013.

\bibitem{Sato2001}
Masa{-}aki Sato.
\newblock Online model selection based on the variational {B}ayes.
\newblock {\em Neural Computation}, pages 1649--1681, 2001.

\bibitem{SeegerB12}
Matthias~W. Seeger and Guillaume Bouchard.
\newblock Fast variational {B}ayesian inference for non-conjugate matrix
  factorization models.
\newblock In {\em AISTATS}, pages 1012--1018, 2012.

\bibitem{ShethKh2016}
Rishit Sheth and Roni Khardon.
\newblock A fixed-point operator for inference in variational {B}ayesian latent
  {G}aussian models.
\newblock In {\em AISTATS}, pages 761--769, 2016.

\bibitem{ShethKh2017}
Rishit Sheth and Roni Khardon.
\newblock Excess risk bounds for the {B}ayes risk using variational inference
  in latent {G}aussian models.
\newblock In {\em NIPS}, 2017.

\bibitem{Sheth2015}
Rishit Sheth, Yuyang Wang, and Roni Khardon.
\newblock {Sparse Variational Inference for Generalized Gaussian Process
  Models}.
\newblock In {\em Proceedings of the International Conference on Machine
  Learning}, 2015.

\bibitem{TehNW06}
Yee~Whye Teh, David Newman, and Max Welling.
\newblock A collapsed variational bayesian inference algorithm for latent
  dirichlet allocation.
\newblock In {\em NIPS}, pages 1353--1360, 2006.

\bibitem{Titsias2009}
Michalis Titsias.
\newblock Variational learning of inducing variables in sparse {G}aussian
  processes.
\newblock In {\em AISTATS}, pages 567--574, 2009.

\bibitem{Titsias2014}
Michalis Titsias and Miguel L\'{a}zaro-Gredilla.
\newblock Doubly stochastic variational {B}ayes for non-conjugate inference.
\newblock In {\em ICML}, pages 1971--1979, 2014.

\bibitem{Titsias2010}
Michalis~K. Titsias and Neil~D. Lawrence.
\newblock {Bayesian Gaussian process latent variable model}.
\newblock In {\em AISTATS}, pages 844--851, 2010.

\bibitem{Wallach2009}
Hanna~M. Wallach, Iain Murray, Ruslan Salakhutdinov, and David Mimno.
\newblock Evaluation methods for topic models.
\newblock In {\em Proceedings of the 26th Annual International Conference on
  Machine Learning}, ICML '09, pages 1105--1112, 2009.

\bibitem{Wand14}
Matt~P. Wand.
\newblock Fully simplified multivariate normal updates in non-conjugate
  variational message passing.
\newblock {\em Journal of Machine Learning Research}, 15(1):1351--1369, 2014.

\bibitem{Wang2013}
Chong Wang and David~M Blei.
\newblock Variational inference in nonconjugate models.
\newblock {\em Journal of Machine Learning Research}, 14:1005--1031, 2013.

\bibitem{WinnB05}
John~M. Winn and Christopher~M. Bishop.
\newblock Variational message passing.
\newblock {\em Journal of Machine Learning Research}, 6:661--694, 2005.

\end{thebibliography}
\bibliographystyle{plain}

\end{document}